\pdfoutput=1
\PassOptionsToPackage{hyperfootnotes=false}{hyperref}
\documentclass[11pt]{article}
\usepackage{booktabs}
\usepackage{longtable}
\usepackage{amsmath}
\usepackage{booktabs}
\usepackage[preprint]{acl}

\usepackage{times}
\usepackage{latexsym}
\usepackage[utf8]{inputenc}
\usepackage[T2A]{fontenc}
\usepackage[russian,english]{babel}
\usepackage{multirow}
\usepackage{adjustbox, booktabs, multirow, multicol, amssymb, url}

\usepackage[T1]{fontenc}

\usepackage[utf8]{inputenc}

\usepackage{microtype}

\usepackage{inconsolata}

\usepackage{graphicx}
\usepackage{pgfplots}
\usepgfplotslibrary{groupplots}
\pgfplotsset{compat=1.18}
\usepackage{booktabs}
\usepackage{multirow}
\usepackage[table]{xcolor}

\usepackage[many]{tcolorbox}
\usepackage{xcolor}
\usepackage{caption}
\usepackage{float}
\usepackage[utf8]{inputenc} 
\usepackage{color}
\usepackage{pifont}        
\usepackage{xcolor}        
\usepackage{booktabs}      

\tcbuselibrary{listings}

\usepackage{pifont}
\newcommand{\circnum}[1]{\ding{\numexpr171+#1}}

\newcommand{\cmark}{\textcolor{green!60!black}{\ding{51}}}  
\newcommand{\xmark}{\textcolor{red!70!black}{\ding{55}}}    

\definecolor{darkgreen}{rgb}{0.0,0.5,0.0}
\definecolor{tidered}{HTML}{C0392B}
\definecolor{tideblue}{HTML}{2980B9}
\definecolor{tidegreen}{HTML}{27AE60}
\definecolor{tideorange}{HTML}{D68910}
\definecolor{headshade}{RGB}{230,238,248}
\definecolor{groupshade}{RGB}{244,246,249}
\definecolor{cutshade}{RGB}{252,244,227}
\usepackage{colortbl}
\definecolor{tidetint}{rgb}{0.90,0.945,0.98} 
\newcommand{\hmark}{$\circ$}
\usepackage{makecell}
\usepackage{tikz}
\definecolor{tideDrop}{HTML}{C0392B}

\definecolor{paraA}{HTML}{EEF0F2}   
\definecolor{paraB}{HTML}{DDE1E6}
\definecolor{ragA}{HTML}{E1EBF7}    
\definecolor{ragB}{HTML}{C4D8EF}
\definecolor{iclA}{HTML}{E7F1E2}    
\definecolor{iclB}{HTML}{CBE3C2}
\definecolor{inccol}{HTML}{0F7B6C}  
\definecolor{cuecol}{HTML}{51607A}  

\newcommand{\bs}[1]{#1}                                   
\newcommand{\gi}[1]{\textcolor{inccol}{\scriptsize #1}}  
\newcommand{\bb}[1]{\textbf{#1}}                         
\newcommand{\uu}[1]{\underline{#1}}                       
\newcommand{\stg}[1]{\textbf{\small #1}}                  
\newcommand{\cue}[1]{\textcolor{cuecol}{\scriptsize\itshape #1}}
\newcommand{\lockshut}{\raisebox{-1pt}{\begin{tikzpicture}[x=1pt,y=1pt,line width=0.5pt,line join=round]
  \draw (-1.5,1.5) -- (-1.5,3) arc (180:0:1.5) -- (1.5,1.5);   
  \fill[rounded corners=0.6pt] (-2.5,-3) rectangle (2.5,1.5);  
  \fill[white] (0,-0.3) circle (0.5);                          
  \fill[white] (-0.28,-1.7) rectangle (0.28,-0.3);
\end{tikzpicture}}}
\newcommand{\lockopen}{\raisebox{-1pt}{\begin{tikzpicture}[x=1pt,y=1pt,line width=0.5pt,line join=round]
  \draw (-1.5,1.5) -- (-1.5,3.4) arc (180:30:1.5);             
  \fill[rounded corners=0.6pt] (-2.5,-3) rectangle (2.5,1.5);
  \fill[white] (0,-0.3) circle (0.5);
  \fill[white] (-0.28,-1.7) rectangle (0.28,-0.3);
\end{tikzpicture}}}

\definecolor{tideDrop}{HTML}{C0392B}
\definecolor{tideInc}{HTML}{2E7D32}
\newcommand{\inc}[2]{{\setlength{\fboxsep}{1.3pt}%
  \colorbox{tideInc!#1}{\scriptsize\textcolor{black!75}{+#2}}}}

\providecommand{\exhead}[1]{\par\medskip\noindent\textbf{#1}\par\smallskip}
\providecommand{\exfield}[2]{\par\noindent\textbf{#1:}\ #2\par\smallskip}
\providecommand{\exlabel}[1]{\par\noindent\textbf{#1:}\par}

\tcbset{expanel/.style 2 args={enhanced, breakable,
  colback=#1!6!white, colframe=#1!50!black,
  colbacktitle=#1!50!black, coltitle=white,
  fonttitle=\bfseries, title={#2},
  boxrule=0.5pt, arc=3pt, left=5pt, right=5pt, top=4pt, bottom=4pt}}

\makeatletter
\def\@fnsymbol#1{\ensuremath{\ifcase#1\or \spadesuit\or \clubsuit\or \diamondsuit\else\@ctrerr\fi}}
\makeatother

\title{\textit{Time Present and Time Past:} Benchmarking Large Language Models on Temporally Evolving Document Understanding}

\author{
 \textbf{Mahbub E Sobhani\textsuperscript{1}\textsuperscript{$\spadesuit$}},
 \textbf{Md. Faiyaz Abdullah Sayeedi\textsuperscript{1}\textsuperscript{$\spadesuit$}},
 \textbf{Fahmid Hasan Chowdhury\textsuperscript{1}},
\\
 \textbf{Md Adnan Arefeen\textsuperscript{2}},
 \textbf{Farig Sadeque\textsuperscript{1}},
 \textbf{Md. Faizul Bari\textsuperscript{3}},
 \textbf{Swakkhar Shatabda\textsuperscript{1}\textsuperscript{$\clubsuit$}}
\\
\\
 \textsuperscript{1}BRAC University \quad
 \textsuperscript{2}North South University \quad
 \textsuperscript{3}Spectrum Software \& Consulting Ltd.
\\[0.4em]
\small{
   \href{https://huggingface.co/datasets/mahbubhimel/TIDE}{%
   \includegraphics[height=1.5em]{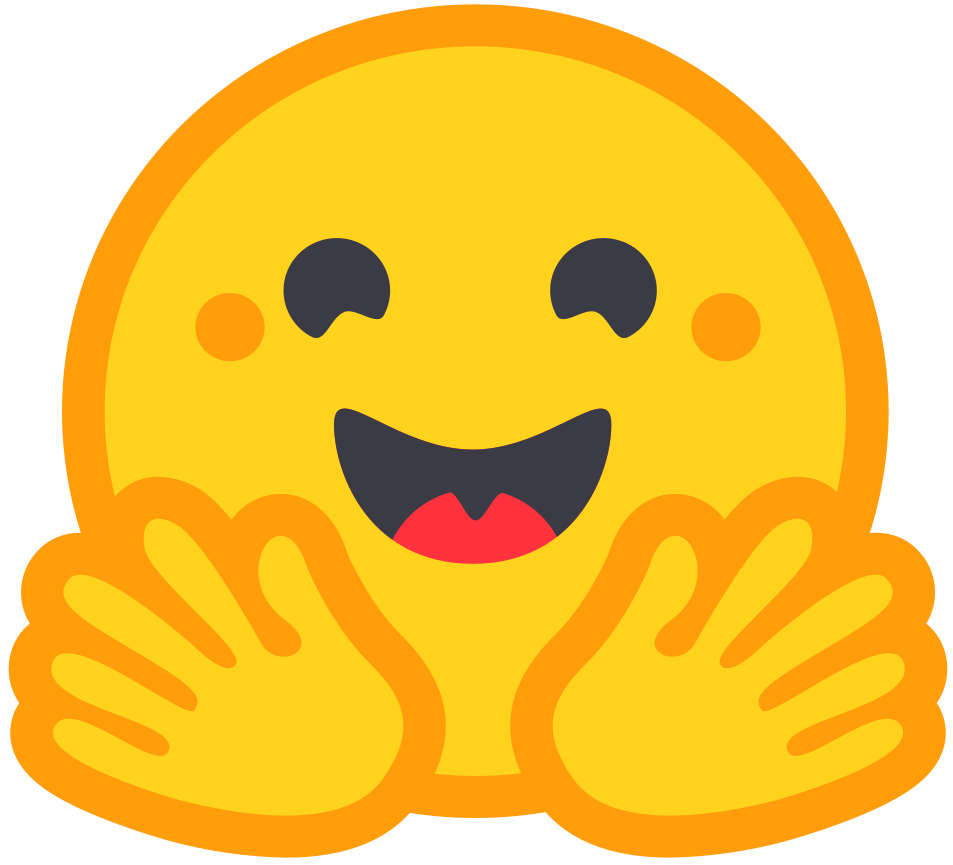}\enspace HuggingFace/Datasets/TIDE}
}
}

\newcommand{\dataname}{\textcolor{tidered}{\textbf{T}}\textcolor{tideblue}{\textbf{I}}\textcolor{tidegreen}{\textbf{D}}\textcolor{tideorange}{\textbf{E}}}

\begin{document}
\maketitle
\renewcommand{\thefootnote}{\fnsymbol{footnote}}
\setcounter{footnote}{1}\footnotetext{Equal contribution.}
\setcounter{footnote}{2}\footnotetext{Corresponding author: \href{mailto:swakkhar.shatabda@bracu.ac.bd}{swakkhar.shatabda@bracu.ac.bd}}
\renewcommand{\thefootnote}{\arabic{footnote}}
\setcounter{footnote}{0}
\begin{abstract}

\textit{Evolving documents}, such as laws, tax codes, and software documentation, are amended, replaced, and sometimes reverted over time, so a question has different correct answers at different dates. In contrast to encyclopedic knowledge, where an old fact is simply overwritten, an amendment is itself an official text that states what it replaces and when it takes effect, and the earlier version stays correct for its validity period. The central challenge is therefore \emph{version resolution}, that is, identifying the version in force on the queried date. Existing temporal QA datasets treat time only as an annotation, so version resolution stays untested. We present \texttt{\dataname{}}, an expert-verified benchmark of 3,050 QA pairs over 644 official customs instruments issued between 1969 and 2025 by the Government of Bangladesh, covering eight task types over deeply code-mixed documents that are heterogeneous in layout and dated in two calendars. In addition, we evaluate nine recent LLMs under a single protocol across parametric, gold-context, and retrieval access, scored by a three-judge LLM council with a hard date gate separating correct meaning from correct time. The best macro-averaged accuracy is only 68.5\%. Resolving a version from an implicit date reaches 59.7\%, and detecting that the supplied version does not govern the query reaches only 26.7\%. Models are more likely to find correct versions than to reject incorrect ones, and they tend to follow a confident parametric answer over the supplied authoritative text. All code and data are available at  \url{https://github.com/icsetepa44/TIDE}
\end{abstract}

\section{Introduction}\label{intro}

Wikipedia is the traditional form of evolving knowledge, where a fact changes by simply overwriting the old text \citep{nakshatri2025factschangeprobingllms}. In contrast, an official rule changes through a formal amendment or revision, which is itself an official text that states what it replaces or reverts and when it takes effect \citep{levinson2001designing}. The earlier version therefore stays correct for the period it covers. We term such texts \textit{evolving documents}. Laws, tax codes, drug regulations, and medical guidelines all work in this way. For example, a customs notification supersedes an earlier duty rate from a stated date, and a new software release deprecates an interface while older versions keep the previous behavior. Several versions of the same rule can all be correct, each for its own period. An LLM answering a question about such a document must therefore identify the version that governs the queried date. This challenge is \textit{version resolution}. When it fails, an answer taken from a superseded version can cause legal, financial, or clinical harm.

\begin{figure}[t]
    \centering
    \includegraphics[width=1\linewidth]{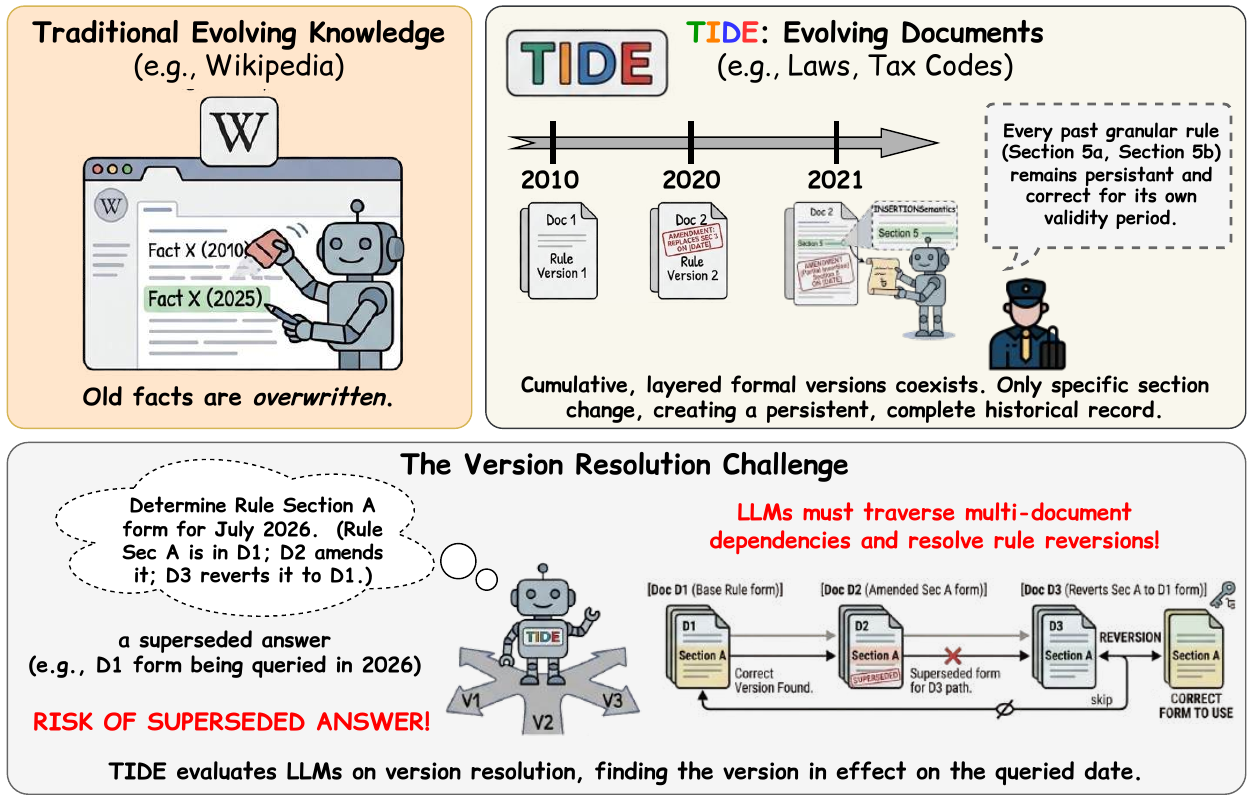}
    \caption{Comparison of traditional evolving knowledge and evolving documents.}
    \label{fig:motivational_figure}
    \vspace{-3mm}
\end{figure}

Version resolution is hard for three reasons, which correspond to the three ways a model can obtain knowledge. First, parametric knowledge freezes at or before the training cutoff \citep{cheng2024dateddatatracingknowledge}. Second, the provided context may come from the wrong period \citep{zhu-etal-2025-evolvebench}. Third, retrieval matches meaning rather than time \citep{zhang-etal-2025-mrag}. In addition, our corpus adds a second axis of difficulty. The documents are deeply code-mixed Bangla--English, they often embed tariff tables, and 97\% of them state dates in both the Bangla and the Gregorian calendar. A model must therefore normalise script and calendar before it can locate the version (see Figure \ref{fig:motivational_figure}).

Time-sensitive QA benchmarks like TempReason \citep{tan-etal-2023-towards} and MenatQA \citep{wei-etal-2023-menatqa}, as well as studies on knowledge that evolves post-training \citep{kim-etal-2026-large}, treat time merely as an annotation. Closest to our setting, evolveQA \citep{nakshatri2025factschangeprobingllms} builds questions from time-stamped AWS, Azure, and WHO updates, but it probes closed-book knowledge and treats superseded values as outdated. Similarly, EvolveBench \citep{zhu-etal-2025-evolvebench} tests misaligned context and implicit dates, but only over Wikidata facts with model-generated contexts. In contrast to an \textit{evolving document}, no source text in either benchmark states what it replaces or when it takes effect. Furthermore, version-aware retrieval ignores amendment semantics \citep{huwiler2025versionragversionawareretrievalaugmentedgeneration}, automatically generated corpora lack expert verification \citep{gruber-etal-2025-complextempqa}, and cross-lingual benchmarks cross languages but not time \citep{asai2021xorqacrosslingualopenretrieval, longpre-etal-2021-mkqa}. Therefore, no existing benchmark tests version resolution over \textit{evolving documents} in a mixed-language setting.

We address these gaps with \texttt{\dataname{}}, an expert-verified benchmark built on authentic \textit{evolving documents}. The benchmark spans eight task categories. Moreover, we evaluate every model under a single protocol with three knowledge-access settings. In summary, our contributions are as follows.

\begin{itemize}
\item We formalize \textit{evolving documents} and propose \texttt{\dataname{}}, an expert-verified benchmark of 3,050 QA pairs over 644 official, deeply code-mixed Bangladesh customs instruments, covering eight task types.
\item We designed a unified evaluation protocol across the parametric, gold-context, and retrieval settings, scored by an LLM council with a hard date gate that separates correct meaning from correct time.
\item We benchmark nine LLMs and show that reading the correct text is necessary but not sufficient, because models locate correct versions far more readily than they reject wrong ones, and they often follow a confident parametric answer over the authoritative text.
\end{itemize}

\section{The \texttt{\dataname{}} Benchmark}\label{sec:tide}

\begin{figure*}[t!] 
    \centering
    \includegraphics[width=\textwidth]{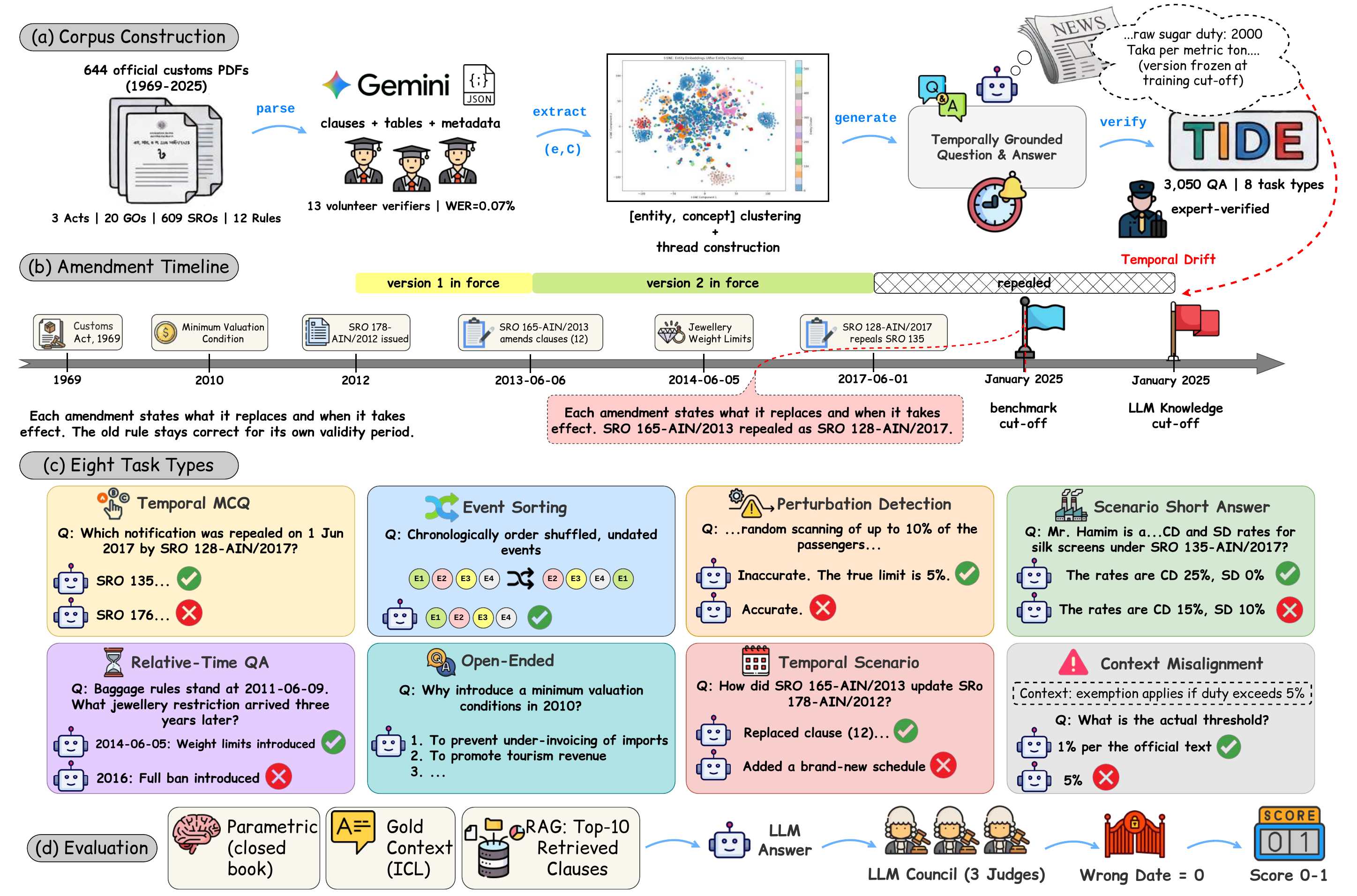}
    \caption{Overview of \texttt{\dataname{}} construction. From 644 official customs instruments, we build verified clauses and threads and generate 3,050 QA pairs across eight task types, evaluated under parametric, gold-context, and retrieval settings with a three-judge LLM council.}
    \label{fig:tide_data_pipeline}
\end{figure*}

We introduce \texttt{\dataname{}} (\textcolor{tidered}{\textbf{T}}emporal \textcolor{tideblue}{\textbf{I}}nformation \textcolor{tidegreen}{\textbf{D}}rift in \textcolor{tideorange}{\textbf{E}}volving Documents), a benchmark for evaluating the evolving knowledge of LLMs. Figure~\ref{fig:tide_data_pipeline} shows how we construct \texttt{\dataname{}}, and Table~\ref{tab:benchmark_comparison} compares it with existing benchmarks. For more details, see Appendix \ref{app:related_works}. 

\begin{table}[!ht]
\centering
\begin{adjustbox}{width=\columnwidth}
\footnotesize
\setlength{\tabcolsep}{1.5pt}
\newcommand{\grouphead}[1]{\multicolumn{8}{l}{\textit{\textcolor{gray}{#1}}}}
\begin{tabular}{llcccccc}
\toprule
 & & \multicolumn{6}{c}{\textbf{Benchmark properties}} \\
\cmidrule(lr){3-8}
\textbf{Benchmark} & \textbf{Source} & \textbf{Temp.} & \textbf{Evol.} & \textbf{Amend.} & \textbf{Pert.} & \textbf{Misal.} & \textbf{P/C/R} \\
\midrule
\grouphead{Temporal and time-sensitive QA} \\
TRAM & Suite & \cmark & \xmark & \xmark & \xmark & \xmark & \xmark \\
TempReason & Wikidata & \cmark & \cmark & \xmark & \xmark & \xmark & \cmark \\
MenatQA & Wikipedia & \cmark & \cmark & \xmark & \cmark & \xmark & \xmark \\
\addlinespace
\grouphead{Evolving knowledge} \\
EvolveBench & Wikidata & \cmark & \cmark & \xmark & \xmark & \cmark & \hmark \\
evolveQA & Web docs & \cmark & \cmark & \xmark & \xmark & \xmark & \xmark \\
OAKS & Synthetic & \cmark & \cmark & \xmark & \xmark & \hmark & \xmark \\
\addlinespace
\grouphead{Version-aware retrieval} \\
TempRAGEval & QA sets & \cmark & \cmark & \xmark & \cmark & \xmark & \xmark \\
VersionQA & Tech docs & \cmark & \cmark & \xmark & \xmark & \xmark & \xmark \\
\addlinespace
\grouphead{Cross-lingual QA} \\
XOR-TyDi QA & Wikipedia & \xmark & \xmark & \xmark & \xmark & \xmark & \xmark \\
MKQA & NQ & \xmark & \xmark & \xmark & \xmark & \xmark & \xmark \\
\midrule
\rowcolor{tidetint}
\texttt{\texttt{\dataname{}} (ours)} & \textbf{Customs law} & \cmark & \cmark & \cmark & \cmark & \cmark & \cmark \\
\bottomrule
\end{tabular}
\end{adjustbox}
\caption{\texttt{\dataname{}} offers a more thorough evaluation of LLMs on time-evolving
knowledge. \textbf{Amend.}: the amending text names what it replaces and its effective date.
\textbf{Pert.}: a value in the statement is altered. \textbf{Misal.}: the context conflicts
with the correct answer. \textbf{P/C/R}: the parametric, gold-context, and retrieval settings,
with \hmark{} marking support for only two. NQ is Natural Questions.}
\label{tab:benchmark_comparison}
\end{table}


\subsection{Data Sourcing}
\textit{Evolving documents}, in which later versions amend earlier ones, appear in many domains, including finance, customs, income tax, VAT, law, software requirements, and medicine. Among these, we selected Bangladesh Customs because entity-level information, such as VAT, tax, and rules, changes frequently over time. We collected 644 PDFs from official the website\footnote{\url{https://nbr.gov.bd/regulations/acts/customs-acts/eng}}: 3 \texttt{Acts of Parliament}, 20 \texttt{General Orders} (GOs), 609 \texttt{Statutory Regulatory Orders} (SROs), and 12 \texttt{Rules}, spanning 1969 to 2025. 


\subsection{Document Parsing}\label{sec:parsing}
One key challenge was extracting every document without text loss. The PDFs are heterogeneous, written in code-mixed Bangla and English, and range from 1 to 169 pages. Therefore, we defined a general extraction schema that preserves clauses, clause hierarchy, tables, and metadata, and used \texttt{Gemini 3.1 Pro} for information extraction. Thirteen third-year Computer Science undergraduate volunteers verified OCR fidelity. \texttt{Gemini} omitted no words but occasionally substituted garbage tokens due to the code-mixed script and tokenization errors, resulting in a word error rate of 0.07–0.08\%. The annotators manually corrected all errors to minimize text loss.

\subsection{Entity and Concept Extraction}\label{subsec:ec-extraction}
We start from our corpus of OCR-parsed Bangladesh customs documents, organized in a four-level hierarchy $H = \{\texttt{Act}, \texttt{SRO}, \texttt{Rule}, \texttt{GO}\}$. Each document is segmented into clauses and metadata elements, with embedded tables restored inline so that no content is lost. For each element $u$ with level $\ell_u \in H$, we use \texttt{Gemini 3.1 Pro} as an extractor $G_{\mathrm{ec}}(\cdot)$ to obtain $m$ entity--concept pairs $(e_i, C_i)$, where $e_i$ is the canonical subject in its official English name and $C_i$ is a set of short plain-English descriptions of what $u$ states about $e_i$. We condition the prompt on $\ell_u$, so that concepts capture foundational rules for Acts, legal changes for SROs, procedures for Rules, and operational instructions for GOs. Moreover, we re-inject previously extracted names $N_u$ as preferred canonical names, which keeps naming consistent across documents. Finally, a second pass $G_{\mathrm{sub}}(\cdot)$ decomposes each entity into sub-entities $S_i$ grounded in $C_i$.

\subsection{Temporally Grounded QA Generation}
\label{subsec:qa-generation}
\textbf{Entity Clustering.}
A question about an \textit{evolving document} is meaningful only once we know which provisions
speak about the same referent. We therefore treat the entities $e_i$ of
Section~\ref{subsec:ec-extraction} and their sub-entities $S_i$ as one universe
$\mathcal{E}$ of surface forms, and we embed each surface form $e$ with its concept text
as $\phi(e)$. Embeddings only propose candidates, whereas the model decides. For every
$e \in \mathcal{E}$, we keep its nearest neighbours $\mathcal{N}(e)$ above a high cosine
similarity ($\geq 0.90$). Moreover, a greedy set cover packs these neighbourhoods into
bounded judge units $\{U_j\}$, so that every candidate pair is examined at least once.

\noindent \textbf{Alias Judgement.}
Threshold-based merging alone is unsafe, because similarity chains unrelated entities into
one giant component. A judge $G_{\mathrm{ali}}$ therefore reads one unit $U_j$ at a time and
confirms only true aliases, abbreviations, and numeral variants. In contrast, it rejects
merely related and part-whole terms, and it leaves unclear cases separate. A union-find
closure over the confirmed pairs then yields the entity clusters $\{K_c\}$, only a minority
of which merge two or more variants. Finally, every element $u$ is tagged with the clusters
of the entities it~mentions.

\noindent \textbf{Evolving Concept Threads.}
Since evolution needs two time points, a cluster $K_c$ qualifies only when it spans enough
provisions across at least two distinct dates. For each qualifying cluster, the model
$G_{\mathrm{thr}}$ reads its date-sorted provisions and splits the cluster into evolving
threads. Each thread is a regulatory topic $R$, that is, a time-ordered list of dated
instruments and the values they set. Beyond this timeline, a thread also records a summary,
its member provisions, and the dimensions along which the rule moved, for example legal
basis, duty structure, eligibility, and scope. Restatements are separated out, because
official texts repeat boilerplate across years. This step yields the thread set $\{R_i\}$
that drives generation.

\noindent \textbf{Question \& Answer Pair Generation.}
The generator $G_{\mathrm{gen}}$ takes a thread $R$ as its only input, so every question is
written against a history rather than a snapshot. Each thread context provides its summary,
evolution dimensions, and full timeline, plus a temporally stratified sample of provisions
that always keeps the earliest and latest. We issue one $G_{\mathrm{gen}}$ call per task type
$y \in \mathcal{Y}$, which generates each type in depth under its own prompt. A call returns
a pair $(q, a)$ whose answer $a$ must state the version in force and its effective date, and
this fixes temporal correctness at generation time. Furthermore, each item must be grounded
in the context with no invented date, rate, instrument, or entity, and must be
self-contained, non-trivial, and distinct within the thread. When a thread cannot honestly
support a type, $G_{\mathrm{gen}}$ returns nothing, so weak questions are never forced. Every
pair also cites its supporting provisions, which resolve to exact clause identifiers and
source files, making both the asked fact and the gold answer traceable to official text.
Finally, two subject-matter experts (SMEs) independently verify each pair on a three-point
Likert scale \citep{amidei-etal-2019-use}, rating $94.89\%$ as factually correct with strong
agreement at Cohen's $\kappa = 0.91$. Appendix~\ref{app:expert_verification} gives the
per-task human-validation breakdown, and Appendix~\ref{app:all_prompts} lists all generation
prompts.

\subsection{Task Types}
\label{sec:task_types}
\texttt{\dataname{}} comprises eight temporally grounded QA task types, totaling 3,050
instances. Appendix~\ref{app:qualitative_examples} gives two verbatim examples per type. We model each regulatory topic $R$ as a time-ordered list of
$n$ dated instruments. The $k$-th instrument $I_k$ takes effect on date $d_k$, where
$d_1 < \dots < d_n$, and puts a value $v_k$ in force, such as a duty rate or item code. We
write $V = [v_1, \dots, v_n]$ for the value timeline of $R$. For a query date $t$, $v_t$ is
the value in force and $I_t$ the instrument that set it. The temporal tasks differ mainly in
how $t$ is provided.

\smallskip
\noindent\textbf{Temporal MCQ} ($y_{\mathrm{mcq}}$). The model must identify the correct
instrument, code, or date among four close options. Because instrument numbers restart every
year, the distractors are lexically close to the in-force instrument $I_t$. Therefore, a
plausible option is not always correct, and the task isolates version resolution from the
surrounding reasoning.

\noindent\textbf{Event Sorting} ($y_{\mathrm{sort}}$). The model reconstructs the
chronological order of a shuffled set of instruments $\{I_k\}$ whose dates have been removed.
Because no instrument shows its date, the order must follow from what each change did rather
than from the surface. Moreover, the order is never stated in a single clause. Therefore, the
task also tests whether the model can combine evidence across the timeline~$V$.

\noindent\textbf{Perturbation Detection} ($y_{\mathrm{pert}}$). The model receives a fluent
statement $s$ about a value $v_k$, its instrument $I_k$, or its date $d_k$, in which one field
may have been altered. It must decide whether $s$ is accurate and, when $s$ is wrong, name the
altered field and correct it rather than flag it. Because $s$ reads as confident, the
task measures verification against the source rather than agreement with the~prompt.

\noindent\textbf{Scenario Short Answer} ($y_{\mathrm{ssa}}$). We place a single named
instrument $I_k$ inside a realistic scenario, and the model extracts its value $v_k$, such as
a duty rate. The scenario never quotes the clause, and the value usually sits inside a
code-mixed table. Therefore, the task measures grounding across the language and layout
boundary. Furthermore, it serves as the single-instrument baseline against which the temporal
tasks are read.

\noindent\textbf{Temporal Scenario Short Answer} ($y_{\mathrm{tssa}}$). Given a query date
$t$ for a topic with $n \ge 2$ instruments, the model returns the value in force $v_t$. This
requires reasoning over several instruments at once, because later rules amend earlier ones.
In contrast to the previous task, a single provision is never sufficient, since the answer
depends on which instrument $t$ selects. Therefore, this is the most direct test of amendment
semantics.

\noindent\textbf{Relative-Time QA} ($y_{\mathrm{rel}}$). This task gives an anchor date $t_0$
and a relative offset $\Delta$, such as ``three years later''. The model must first compute
the target date $t_0 + \Delta$ and then return the value in force there, $v_{t_0 + \Delta}$.
Because this date is never stated, it cannot be matched by words or retrieved directly.
Therefore, the task separates temporal arithmetic from content lookup.

\noindent\textbf{Open-Ended} ($y_{\mathrm{oe}}$). This task asks about the reason behind a
policy change rather than a single value. It checks whether the model understands the
direction and purpose of a change from $v_k$ to $v_{k+1}$, not only the value that changed.
Similarly, it shows whether success on the short-answer tasks reflects understanding or
copying from the text.

\noindent\textbf{Context Misalignment} ($y_{\mathrm{cm}}$). A misleading preamble $\tilde{c}$
states a wrong version, and the model must ignore it and answer from the verified value $v_t$.
This mirrors deployment, where a superseded version is pasted into the prompt. Moreover, in
the retrieval setting the correct clauses are placed above $\tilde{c}$. Therefore, the task
measures whether authoritative evidence can override a confident false~claim.

\subsection{Dataset Statistics}

\begin{table}[t]
\centering
\resizebox{\linewidth}{!}{%
\begin{tabular}{lrlrrr}
\toprule
\rowcolor{headshade} \textbf{Task Type} & \textbf{Count ($n$)} & \multicolumn{2}{c}{\cellcolor{headshade}\textbf{Stimulus}} & \textbf{Question} & \textbf{Answer} \\
\midrule
Temporal MCQ                   & 354 & 4 options      & 136 & 128 & 41  \\
Event Sorting                  & 207 & 3.5 events     & --  & 69  & --  \\
Perturbation Detection         & 498 & Statement      & 167 & 67  & 181 \\
Scenario Short Answer          & 562 & Scenario       & 493 & 104 & 120 \\
Temporal Scenario Short Answer & 363 & Scenario       & 285 & 156 & 253 \\
Relative-Time QA               & 463 & Seed \& offset & --  & 167 & 199 \\
Open-Ended                     & 487 & --             & --  & 122 & \textbf{381} \\
Context Misalignment           & 116 & Context        & 216 & 127 & 271 \\
\midrule
\textbf{Total} & \textbf{3050} & & & & \\
\bottomrule
\end{tabular}
}
\caption{\texttt{\dataname{}} instances per task (mean characters).}
\label{tab:dataset_stats}
\vspace{-3mm}
\end{table}

\texttt{\dataname{}} consists of 3,050 QA instances drawn from 265 source threads across 157 topic clusters, and every instance retains full provenance metadata, including source clause IDs, thread identifiers, and originating files. Table \ref{tab:dataset_stats} reports the distribution and average character lengths of inputs and answers across the eight task types, where counts include code-mixed Bengali text. Notably, \textit{Scenario Short Answer} has the longest average input (493 characters), while \textit{Open-Ended} requires the longest answers (381 characters). For a detailed understanding of the complexity of the dataset, please see Appendix \ref{app:dataset_analysis}.

\section{Evaluation Setup}

Every \texttt{Gemini} call, as subject model and as council judge, was served by the
Google AI Studio Gemini Developer API,\footnote{\url{https://aistudio.google.com/api-keys}}
while all other models were served through
OpenRouter.\footnote{\url{https://openrouter.ai/}}

\subsection{Models} 
We evaluate nine recent LLMs (six open-weight, three proprietary) from seven model families on \texttt{\dataname{}}. The open-weight group includes \texttt{Qwen3.6-35B} \citep{yang2025qwen3}, \texttt{GPT-OSS-120B} \citep{openai2025gptoss120bgptoss20bmodel}, \texttt{Gemma-4-31B} \citep{gemmateam2026gemma4technicalreport}, \texttt{Kimi-K2.5} \citep{kimiteam2026kimik25visualagentic}, \texttt{GLM-4.6} \citep{5team2025glm45agenticreasoningcoding}, and \texttt{DeepSeek-V4-Flash} \citep{deepseekai2026deepseekv4highlyefficientmilliontoken}. In addition, we evaluate the proprietary \texttt{GPT-5.2} \citep{openai2025gpt52}, \texttt{Gemini-3.5-Flash} \citep{googledeepmind2026gemini35flash}, and \texttt{Claude-Sonnet-4.5} \citep{anthropic2025claudesonnet45} to benchmark against enterprise-grade models. An LLM's knowledge is limited by its training cut-off date. This is crucial for \texttt{\dataname{}}, so we select models whose cut-off dates cover the date of our most recent document. To evaluate all models consistently, we set the temperature to \(T = 0.0\) and a maximum generation length of 32k tokens, with thinking mode enabled for all models.

\subsection{Evaluation Metrics}
\label{sec:evaluation_metrics}
\paragraph{Accuracy.}
For the deterministic tasks, we use accuracy as the main metric. For temporal
MCQ, an answer is correct if the predicted option matches the gold option. For
perturbation detection, we score the binary verdict against the gold label over
perturbed and unperturbed statements together. For event ordering, we report
exact-match accuracy.
\paragraph{LLM-as-a-Council.}
\label{eval:llm_council}
For the other six free-form tasks, answers are validated for temporal, factual, and
conclusive equivalence with the ground truth by an \textbf{LLM-as-a-Council} of three
heterogeneous judges: \texttt{Claude Opus 4.6}, \texttt{Gemini 3.5 Flash}, and
\texttt{GPT-5.1}. Rather than assigning numeric scores, each judge answers a fixed set of
yes/no questions in JSON, which a deterministic program turns into scores. The metric is
therefore reproducible and testable. The judges reach an inter-rater reliability of
Fleiss' $\kappa \in [0.8297, 0.861]$. Equation~\ref{eq:council_metric} formalizes this
process.
\begin{equation}
\footnotesize
J_k:(A_{llm},A_{gt}) \mapsto (R_k, \mathbf{b}_k), \qquad
\mathcal{C} = \operatorname{maj}\big(\mathbf{b}_1, \mathbf{b}_2, \mathbf{b}_3\big)
\label{eq:council_metric}
\end{equation}
Each judge $J_k$ maps the model answer $A_{llm}$ and the ground truth $A_{gt}$ to a
reasoning trace $R_k$ and a boolean vector $\mathbf{b}_k$. Its first part holds one
boolean per canonical ground-truth date, recording whether the answer satisfies that date
and whether it contradicts it. Its second part holds one item-level boolean for whether
the answer agrees with the ground truth in meaning. The verdict $\mathcal{C}$ is an
element-wise majority vote over the three vectors, which the deterministic scorer of
\S\ref{label:scoring} reads. Reference dates are extracted deterministically and given to
the council as a canonical list, so judges never search for them. Our council extends the
LLM-as-a-judge setup of \citet{kim2026harnessingtemporaldatabasessystematic}. To check
reliability, two independent subject-matter experts (SMEs) rescore the same answers with
the rubric of \citet{e-sobhani-etal-2026-mathmist}. Their average agreement with the
council is 94.21\%.
\paragraph{Deterministic scoring.} \label{label:scoring}
Let $n$ be the number of canonical ground-truth dates, and let $m$ be the number of those
dates that $\mathcal{C}$ marks as satisfied at the required granularity. Let $\mu$ be the
item-level meaning boolean of $\mathcal{C}$ after contradiction forcing, so $\mu$ is false
whenever the answer contradicts a canonical date, even when the judges accept its meaning.
The final score $S \in [0,1]$ follows Equation~\ref{eq:base_rubric}.
{\footnotesize
\begin{equation}
S(m,n,\mu)=
\begin{cases}
[\mu] & n = 0\\[2pt]
0.5 + 0.5\,[\mu] & n = 1,\; m = 1\\[2pt]
0 & n = 1,\; m = 0\\[2pt]
0.5\frac{m}{n} + 0.5\,[m = n \wedge \mu] & n \geq 2
\end{cases}
\label{eq:base_rubric}
\end{equation}}
Here, $[\cdot]$ is the Iverson bracket (1 if the condition holds, 0 otherwise). The date
component is a hard gate. A wrong date on a single-date item scores zero, and multi-date
items earn the meaning credit only when every reference date is satisfied. For
relative-time QA, the required granularity depends on the offset: the full date up to
three months, the year and month for longer month-level offsets, and the year alone for
year-level offsets.

\subsection{Knowledge Access Settings}\label{sec:settings}
We evaluate every model under three settings that differ only in the model's access to the corpus.

\noindent \textbf{Parametric Knowledge.}\label{sec:closed_book}
Here we probe the knowledge stored in a model's parameters over our corpus of \textit{evolving documents}. We select models whose pre-training data covers our full corpus, spanning $[1969, 2025]$. Since documents are available online and models are pre-trained on extensive web data, we provide only the question and evaluate the closed-book prediction $a = f(q)$, where the model $f$ answers the question $q$ from its internal knowledge alone.

\noindent \textbf{In-Context Learning (ICL).}\label{sec:gold}
In this setting, we include the gold context documents in the input. For each question, the model predicts $a = f(q, D)$, where $D$ is the set of gold documents used to build the corresponding question--answer pair. In contrast to the parametric setting, the model no longer relies on memory alone, so this setting measures how well it reads and uses the given evidence.

\noindent \textbf{Retrieval-Augmented Generation (RAG).}\label{sec:rag}
In this setting the model predicts $a = f(q, C)$, where the retrieved set
$C$ replaces the gold documents $D$ of the ICL setting
\S\ref{sec:gold}. We embed the parsed corpus into $\mathbb{R}^{3072}$ with
\texttt{gemini-embedding-2} \citep{lee2025geminiembeddinggeneralizable} and
index it in \texttt{ChromaDB}:\footnote{\url{https://www.trychroma.com/}}
4,858 elements whole, plus 2,081 overlapping 1,200-character chunks for the 82
clauses above the 2,048-token limit, for 6,939 vectors in total. For each
instance we embed the task-specific input fields in query mode
\citep{ram-etal-2023-context}, excluding the misleading context in context
misalignment, and keep the $k = 40$ nearest vectors by cosine similarity. We
then map chunks to parent clauses and deduplicate to at most 10 complete
clauses (on average $|C| = 9.9$), which we prepend as authoritative
sources. We use \texttt{gemini-embedding-2} with Top-10 retrieval throughout
the main experiments because this configuration provides the most stable
performance across task types and answer models. Appendix~\ref{app:rag_ablation}
presents the complete ablation over embedding models and retrieval depths.
Moreover, the retriever is a standard semantic one without temporal re-ranking
\citep{zhang-etal-2025-mrag}. All other components match the ICL setting.

\section{Results \& Analysis}

\begin{table*}[t]
  \centering
  \setlength{\tabcolsep}{5pt}
  \renewcommand{\arraystretch}{1.25}
  \begin{adjustbox}{max width=\textwidth}
  \footnotesize
  \begin{tabular}{@{}l ccc cccccc@{}}
  \toprule
  & \multicolumn{3}{c}{\textbf{\lockshut~Proprietary models}} & \multicolumn{6}{c}{\textbf{\lockopen~Open-Source models}} \\
  \cmidrule(lr){2-4}\cmidrule(lr){5-10}
  \makecell[l]{\textbf{Task type}\\[1pt]\cue{strict\;(\textcolor{inccol}{+partial})}} & \textbf{\texttt{GPT-5.2}} & \textbf{\texttt{Claude-Sonnet-4.5}} & \textbf{\texttt{Gemini-3.5-Flash}} & \textbf{\texttt{Kimi-K2.5}} & \textbf{\texttt{DeepSeek-V4-Flash}} & \textbf{\texttt{GPT-OSS-120B}} & \textbf{\texttt{Qwen3.6-35B}} & \textbf{\texttt{Gemma-4-31B}} & \textbf{\texttt{GLM-4.6}} \\
  \midrule
  \rowcolor{paraB} \multicolumn{10}{c}{\stg{Parametric Knowledge}} \\
  \rowcolor{paraA} Temporal MCQ & \bs{\uu{69.57}} & \bs{54.24} & \bs{64.72} & \bs{\uu{69.57}} & \bs{67.27} & \bs{58.21} & \bs{56.88} & \bs{\bb{70.17}} & \bs{48.97} \\
  \rowcolor{paraA} Perturbation Det. & \bs{58.63} & \bs{56.83} & \bs{\uu{59.84}} & \bs{56.43} & \bs{59.64} & \bs{56.63} & \bs{44.38} & \bs{\bb{61.85}} & \bs{46.39} \\
  \rowcolor{paraA} Event Sorting & \bs{\uu{25.67}} & \bs{17.45} & \bs{\bb{27.13}} & \bs{24.15} & \bs{21.39} & \bs{21.36} & \bs{19.88} & \bs{23.26} & \bs{18.45} \\
  \rowcolor{paraA} Scenario SA & \bs{\bb{41.81}}\,\gi{\bb{+0.25}} & \bs{30.12}\,\gi{+0.13} & \bs{\uu{41.35}} & \bs{40.92}\,\gi{+0.16} & \bs{38.64}\,\gi{+0.24} & \bs{35.99}\,\gi{+0.14} & \bs{33.16}\,\gi{+0.13} & \bs{35.68}\,\gi{+0.13} & \bs{30.68}\,\gi{+0.13} \\
  \rowcolor{paraA} Temporal Scenario SA & \bs{\bb{46.39}}\,\gi{\bb{+3.18}} & \bs{30.68}\,\gi{+3.16} & \bs{\uu{39.44}}\,\gi{+3.09} & \bs{38.88}\,\gi{+2.49} & \bs{39.43}\,\gi{\uu{+3.41}} & \bs{38.69}\,\gi{+2.63} & \bs{29.23}\,\gi{+3.49} & \bs{38.61}\,\gi{+2.53} & \bs{36.19}\,\gi{+3.22} \\
  \rowcolor{paraA} Relative-Time QA & \bs{\uu{11.08}}\,\gi{+20.66} & \bs{3.77}\,\gi{+10.07} & \bs{9.91}\,\gi{\uu{+23.36}} & \bs{10.25}\,\gi{+22.85} & \bs{8.05}\,\gi{+22.41} & \bs{\bb{12.37}}\,\gi{\bb{+24.06}} & \bs{4.58}\,\gi{+20.91} & \bs{4.88}\,\gi{+13.96} & \bs{5.22}\,\gi{+24.85} \\
  \rowcolor{paraA} Open-Ended & \bs{\bb{27.96}}\,\gi{\bb{+0.94}} & \bs{12.91}\,\gi{+1.16} & \bs{\uu{23.23}}\,\gi{\uu{+1.19}} & \bs{21.65}\,\gi{+1.09} & \bs{22.23}\,\gi{+1.02} & \bs{21.49}\,\gi{+0.98} & \bs{16.48}\,\gi{+1.15} & \bs{19.11}\,\gi{+0.92} & \bs{16.86}\,\gi{+1.03} \\
  \rowcolor{paraA} Context Misalign. & \bs{1.78}\,\gi{+10.74} & \bs{1.79}\,\gi{+11.28} & \bs{\bb{7.87}}\,\gi{\bb{+7.98}} & \bs{4.36}\,\gi{+9.65} & \bs{4.31}\,\gi{+9.84} & \bs{1.75}\,\gi{+11.66} & \bs{\uu{6.01}}\,\gi{\uu{+9.03}} & \bs{2.62}\,\gi{+9.48} & \bs{1.72}\,\gi{+10.53} \\
  \rowcolor{paraB} \textbf{Macro avg.} & \bs{\bb{35.39}}\,\gi{\bb{+4.45}} & \bs{25.91}\,\gi{+3.23} & \bs{\uu{34.15}}\,\gi{\uu{+4.42}} & \bs{33.28}\,\gi{+4.56} & \bs{32.64}\,\gi{+4.62} & \bs{30.76}\,\gi{+5.04} & \bs{26.39}\,\gi{+4.33} & \bs{32.04}\,\gi{+3.38} & \bs{25.54}\,\gi{+5.06} \\
  \rowcolor{ragB} \multicolumn{10}{c}{\stg{RAG}} \\
  \rowcolor{ragA} Temporal MCQ & \bs{\uu{84.53}} & \bs{82.01} & \bs{84.21} & \bs{\bb{85.01}} & \bs{83.93} & \bs{80.59} & \bs{83.97} & \bs{73.79} & \bs{69.59} \\
  \rowcolor{ragA} Perturbation Det. & \bs{83.73} & \bs{82.00} & \bs{\bb{85.14}} & \bs{79.72} & \bs{80.32} & \bs{80.32} & \bs{\uu{83.94}} & \bs{82.73} & \bs{80.52} \\
  \rowcolor{ragA} Event Sorting & \bs{\bb{31.42}} & \bs{26.05} & \bs{\uu{30.95}} & \bs{29.58} & \bs{29.51} & \bs{26.16} & \bs{28.58} & \bs{26.11} & \bs{18.41} \\
  \rowcolor{ragA} Scenario SA & \bs{\uu{87.74}}\,\gi{\uu{+0.12}} & \bs{\bb{94.06}} & \bs{84.35}\,\gi{+0.16} & \bs{84.72}\,\gi{+0.44} & \bs{84.57} & \bs{81.54}\,\gi{+0.19} & \bs{83.69} & \bs{80.49}\,\gi{+0.29} & \bs{78.16} \\
  \rowcolor{ragA} Temporal Scenario SA & \bs{\bb{66.71}}\,\gi{\bb{+1.74}} & \bs{62.04}\,\gi{\uu{+4.52}} & \bs{56.53}\,\gi{+2.13} & \bs{60.61}\,\gi{+2.31} & \bs{52.65}\,\gi{+2.98} & \bs{\uu{63.93}}\,\gi{+2.02} & \bs{53.76}\,\gi{+2.32} & \bs{47.42}\,\gi{+2.03} & \bs{50.45}\,\gi{+2.58} \\
  \rowcolor{ragA} Relative-Time QA & \bs{\uu{32.41}}\,\gi{+11.84} & \bs{\bb{48.02}}\,\gi{\bb{+9.02}} & \bs{27.45}\,\gi{+12.92} & \bs{28.12}\,\gi{\uu{+18.27}} & \bs{27.75}\,\gi{+11.41} & \bs{28.52}\,\gi{+15.21} & \bs{26.65}\,\gi{+16.13} & \bs{21.86}\,\gi{+10.36} & \bs{18.89}\,\gi{+10.33} \\
  \rowcolor{ragA} Open-Ended & \bs{\bb{48.57}}\,\gi{\uu{+0.87}} & \bs{\uu{48.09}}\,\gi{\bb{+1.75}} & \bs{36.14}\,\gi{+0.81} & \bs{42.35}\,\gi{+1.32} & \bs{36.59}\,\gi{+1.06} & \bs{43.94}\,\gi{+1.14} & \bs{34.91}\,\gi{+1.06} & \bs{33.34}\,\gi{+1.06} & \bs{33.56}\,\gi{+0.94} \\
  \rowcolor{ragA} Context Misalign. & \bs{21.54}\,\gi{+5.65} & \bs{\uu{22.09}}\,\gi{\uu{+5.52}} & \bs{18.13}\,\gi{+6.83} & \bs{\bb{25.09}}\,\gi{\bb{+8.18}} & \bs{20.76}\,\gi{+6.03} & \bs{16.48}\,\gi{+7.75} & \bs{21.57}\,\gi{+5.45} & \bs{15.52}\,\gi{+5.55} & \bs{13.89}\,\gi{+7.08} \\
  \rowcolor{ragB} \textbf{Macro avg.} & \bs{\uu{57.06}}\,\gi{\uu{+2.54}} & \bs{\bb{58.05}}\,\gi{\bb{+2.65}} & \bs{52.87}\,\gi{+2.94} & \bs{54.46}\,\gi{+3.78} & \bs{52.08}\,\gi{+2.61} & \bs{52.69}\,\gi{+3.36} & \bs{52.13}\,\gi{+3.14} & \bs{47.69}\,\gi{+2.41} & \bs{45.43}\,\gi{+2.65} \\
  \rowcolor{iclB} \multicolumn{10}{c}{\stg{ICL}} \\
  \rowcolor{iclA} Temporal MCQ & \bs{\bb{91.88}} & \bs{86.08} & \bs{\uu{91.07}} & \bs{90.49} & \bs{90.75} & \bs{82.28} & \bs{86.76} & \bs{85.37} & \bs{77.26} \\
  \rowcolor{iclA} Perturbation Det. & \bs{\uu{80.32}} & \bs{80.00} & \bs{\bb{81.33}} & \bs{77.11} & \bs{80.12} & \bs{74.50} & \bs{77.31} & \bs{79.52} & \bs{73.90} \\
  \rowcolor{iclA} Event Sorting & \bs{\uu{66.23}} & \bs{60.05} & \bs{60.42} & \bs{\bb{66.73}} & \bs{65.25} & \bs{39.15} & \bs{58.07} & \bs{63.32} & \bs{33.25} \\
  \rowcolor{iclA} Scenario SA & \bs{\uu{86.58}}\,\gi{\uu{+0.15}} & \bs{\bb{88.04}} & \bs{84.79}\,\gi{+0.11} & \bs{83.56} & \bs{85.97} & \bs{75.46} & \bs{84.92} & \bs{79.42} & \bs{74.02} \\
  \rowcolor{iclA} Temporal Scenario SA & \bs{\bb{75.29}}\,\gi{\bb{+1.25}} & \bs{64.06}\,\gi{+1.52} & \bs{63.45}\,\gi{+1.33} & \bs{62.02}\,\gi{+2.11} & \bs{\uu{66.43}}\,\gi{\uu{+1.59}} & \bs{65.64}\,\gi{+2.05} & \bs{62.35}\,\gi{+1.73} & \bs{59.83}\,\gi{+1.95} & \bs{52.17}\,\gi{+1.69} \\
  \rowcolor{iclA} Relative-Time QA & \bs{\uu{59.66}}\,\gi{\uu{+8.78}} & \bs{\bb{68.03}}\,\gi{\bb{+8.01}} & \bs{49.58}\,\gi{+13.57} & \bs{49.56}\,\gi{+12.34} & \bs{44.96}\,\gi{+14.66} & \bs{46.73}\,\gi{+11.84} & \bs{48.88}\,\gi{+13.34} & \bs{49.28}\,\gi{+10.23} & \bs{36.43}\,\gi{+15.56} \\
  \rowcolor{iclA} Open-Ended & \bs{\bb{61.04}}\,\gi{\bb{+0.54}} & \bs{\uu{54.06}}\,\gi{\uu{+2.49}} & \bs{49.16}\,\gi{+0.76} & \bs{48.94}\,\gi{+1.08} & \bs{53.61}\,\gi{+0.63} & \bs{53.44}\,\gi{+0.61} & \bs{48.73}\,\gi{+0.79} & \bs{43.93}\,\gi{+1.19} & \bs{38.05}\,\gi{+0.94} \\
  \rowcolor{iclA} Context Misalign. & \bs{\uu{26.73}}\,\gi{\uu{+4.31}} & \bs{14.71}\,\gi{+2.52} & \bs{19.88}\,\gi{+5.69} & \bs{\bb{27.62}}\,\gi{\bb{+3.49}} & \bs{18.18}\,\gi{+5.36} & \bs{19.89}\,\gi{+6.04} & \bs{22.42}\,\gi{+5.39} & \bs{19.85}\,\gi{+4.78} & \bs{17.23}\,\gi{+7.36} \\
  \rowcolor{iclB} \textbf{Macro avg.} & \bs{\bb{68.49}}\,\gi{\bb{+4.62}} & \bs{\uu{64.39}}\,\gi{+4.23} & \bs{62.45}\,\gi{+5.91} & \bs{63.27}\,\gi{+4.83} & \bs{63.11}\,\gi{\uu{+5.66}} & \bs{57.11}\,\gi{+6.65} & \bs{61.13}\,\gi{+4.44} & \bs{60.05}\,\gi{+4.63} & \bs{50.38}\,\gi{+8.03} \\
  \bottomrule
  \end{tabular}
  \end{adjustbox}
  \caption{Temporal reasoning performance on \texttt{\dataname{}} across the Parametric Knowledge, RAG, and ICL settings. Scores are percentages (\%). The base number is strict accuracy, and the small teal value is the extra credit from the partial rubric in Equation~\ref{eq:base_rubric}, so base plus teal equals partial accuracy. A single number means the task is deterministic and admits no partial credit, and for event sorting we report exact-match accuracy. \textbf{Bold} and \underline{underline} mark the best and second-best model per row, on the base for strict accuracy and the teal value for partial.}
  \label{tab:main_results}
\end{table*}
 
Table \ref{tab:main_results} reports strict and partial accuracy for all nine models under the parametric, RAG, and ICL settings. Two patterns hold for every model. First, performance rises steadily from the parametric setting to RAG and then to ICL, so access to the source text matters more than model identity. Second, even the best configuration stays far from solved. \texttt{GPT-5} obtains the best macro average in the parametric and ICL settings, while \texttt{Claude} leads under RAG, yet the strongest single number is only $68.49\%$ (\texttt{GPT-5}, ICL), and the version-sensitive tasks collapse well below this level. Therefore, version resolution over \textit{evolving documents} remains an open problem, and the gap is largest exactly where a query depends on which amendment is in force. Further analysis of these results is provided in Appendix \ref{app:more_results}.

\noindent \textbf{\textit{Injecting the source text helps, yet a gold context is still not enough.}}
Every model improves substantially once it can read the documents. For \texttt{GPT-5} the macro average grows from $35.39\%$ in the parametric setting to $57.06\%$ under RAG and $68.49\%$ under ICL, and the same ordering holds for all other models. However, for the ICL setting, even with the relevant documents that contain the relevant clause, together with unrelated clauses, metadata, and tables, performance still tops out at $68.49\%$. Reading the correct text is therefore necessary but not sufficient, because the model must locate the right provision, select the version in force, and attach the correct date.
 
\noindent \textbf{ICL vs RAG.}
RAG trails ICL for every model, and the gap ranges from $4.42$ points (\texttt{GPT-OSS}) to $12.36$ points (\texttt{Gemma}). The two settings differ in what they provide. ICL includes the full gold documents, so every provision in those documents is available even when it is surrounded by unrelated text. RAG instead provides only the top retrieved clauses, so a needed provision can be missing. The loss therefore concentrates in tasks that need the full amendment history. Event Sorting is the clearest case, as \texttt{GPT-5} falls from $66.23\%$ under ICL to $31.42\%$ under RAG, and every model shows a comparable drop. Because the timeline is never stated in a single clause, a semantic retriever without temporal re-ranking misses some of the required provisions, so the order cannot be reconstructed. In contrast, single-instrument Scenario SA stays high under RAG, reaching $94.06\%$ for \texttt{Claude}, because the one relevant clause is easy to retrieve. Retrieval completeness is therefore the limiting factor for amendment-wide reasoning.

\noindent \textbf{\textit{The date gate separates correct meaning from correct time.}} The partial score retains credit for correct dates even when the meaning is wrong, whereas meaning earns credit only when every reference date is satisfied. The gap between partial and strict accuracy is therefore largest for Relative-Time QA, indicating that models often identify the correct date but associate it with the wrong regulatory change, which receives no strict credit because strict scoring requires both the correct date and the correct meaning. In the parametric setting, strict accuracy stays between 3.77\% and 12.37\%, while the partial bonus adds between 10.07 and 24.85 points. The gap shrinks once the documents are available, and \texttt{Claude} becomes the strongest temporal resolver, reaching 68.03\% on Relative-Time QA under ICL and 48.02\% under RAG, well above every other model. These results suggest that retrieving the relevant regulation alone is insufficient; models must also correctly resolve temporal offsets relative to the retrieved evidence.

\noindent \textbf{\textit{Models trust a confident false premise over the authoritative text.}}
Context Misalignment is the hardest task in the benchmark. Strict accuracy never exceeds $7.87\%$ in the parametric setting and $27.62\%$ under ICL, even though the correct answer is fixed by the verified ground truth. In the RAG setting we stack the authoritative clauses above the misleading premise, yet accuracy still stays at or below $25.09\%$. This shows that a fluent but wrong preamble can override correct evidence that sits in the same prompt. Because a superseded or wrong version is often pasted into a real query, this failure mode is the most concerning for deployment.


\begin{table}[t]
\centering
\begin{adjustbox}{width=0.48\textwidth}
\footnotesize
\setlength{\tabcolsep}{4.5pt}
\renewcommand{\arraystretch}{1.12}
\begin{tabular}{@{}l rlr rlr rlr@{}}
\toprule
& \multicolumn{3}{c}{\textbf{Parametric}} & \multicolumn{3}{c}{\textbf{RAG}} & \multicolumn{3}{c}{\textbf{In-Context (ICL)}}\\
\cmidrule(lr){2-4}\cmidrule(lr){5-7}\cmidrule(lr){8-10}
\textbf{Model} & Det. & Ident. & $\Delta$ & Det. & Ident. & $\Delta$ & Det. & Ident. & $\Delta$\\
\midrule
\rowcolor{black!6}\multicolumn{10}{l}{\textit{\lockshut~Proprietary}}\\
\texttt{GPT-5.2} & \underline{91.83} & \underline{21.47}\,\inc{6}{1.91} & \cellcolor{tideDrop!27}76.6 & 96.11 & 57.56\,\inc{7}{2.33} & \cellcolor{tideDrop!12}40.1 & \underline{95.33} & \underline{60.70}\,\inc{9}{2.97} & \cellcolor{tideDrop!11}\underline{36.3} \\
\texttt{Claude-Sonnet-4.5} & 65.37 & 8.33\,\inc{12}{4.42} & \cellcolor{tideDrop!32}87.3 & 93.10 & \textbf{80.17}\,\inc{16}{5.94} & \cellcolor{tideDrop!2}\textbf{13.9} & 89.66 & \textbf{79.31}\,\inc{24}{9.15} & \cellcolor{tideDrop!1}\textbf{11.5} \\
\texttt{Gemini-3.5-Flash} & \textbf{92.22} & 19.55\,\inc{6}{1.65} & \cellcolor{tideDrop!28}78.8 & \textbf{97.67} & 53.96\,\inc{5}{1.29} & \cellcolor{tideDrop!14}44.8 & 94.94 & 41.41\,\inc{7}{2.20} & \cellcolor{tideDrop!19}56.4 \\
\addlinespace[1pt]\rowcolor{black!6}\multicolumn{10}{l}{\textit{\lockopen~Open-weight}}\\
\texttt{Kimi-K2.5} & 87.16 & 21.08\,\inc{9}{3.10} & \cellcolor{tideDrop!27}75.8 & \textbf{97.67} & \underline{65.53}\,\inc{5}{1.57} & \cellcolor{tideDrop!9}\underline{32.9} & 94.16 & 48.90\,\inc{9}{3.03} & \cellcolor{tideDrop!16}48.1 \\
\texttt{DeepSeek-V4-Flash} & 80.54 & 21.17\,\inc{14}{5.12} & \cellcolor{tideDrop!26}73.7 & \underline{97.28} & 55.12\,\inc{5}{1.55} & \cellcolor{tideDrop!14}43.3 & \textbf{96.11} & 48.18\,\inc{6}{1.95} & \cellcolor{tideDrop!16}49.9 \\
\texttt{GPT-OSS-20B} & 78.21 & \textbf{27.50}\,\inc{20}{7.66} & \cellcolor{tideDrop!22}\textbf{64.8} & 96.89 & 62.61\,\inc{6}{2.02} & \cellcolor{tideDrop!10}35.4 & \underline{95.33} & 52.72\,\inc{8}{2.59} & \cellcolor{tideDrop!14}44.7 \\
\texttt{Qwen3.6-35B} & 66.54 & 20.33\,\inc{26}{10.23} & \cellcolor{tideDrop!24}\underline{69.4} & \textbf{97.67} & 59.40\,\inc{5}{1.42} & \cellcolor{tideDrop!12}39.2 & 93.39 & 48.90\,\inc{10}{3.46} & \cellcolor{tideDrop!15}47.6 \\
\texttt{Gemma-4-31B} & 80.93 & 12.65\,\inc{9}{2.97} & \cellcolor{tideDrop!31}84.4 & 94.16 & 48.80\,\inc{9}{3.03} & \cellcolor{tideDrop!16}48.2 & 94.94 & 33.82\,\inc{6}{1.80} & \cellcolor{tideDrop!22}64.4 \\
\texttt{GLM-4.6} & 67.70 & 15.53\,\inc{19}{7.41} & \cellcolor{tideDrop!28}77.1 & 75.10 & 37.42\,\inc{31}{12.41} & \cellcolor{tideDrop!16}50.2 & 68.87 & 29.83\,\inc{34}{13.48} & \cellcolor{tideDrop!19}56.7 \\
\midrule
\textit{Mean} & 78.94 & 18.62 & 76.4 & 93.96 & 57.84 & 38.7 & 91.41 & 49.31 & 46.2\\
\bottomrule
\end{tabular}
\end{adjustbox}
\caption{Perturbation detection vs.\ identification, over perturbed items only. \textbf{Ident.} is the mean explanation score across all perturbed items, with the increment over flagged items only tiled in green. $\Delta$ is the relative drop from Det.\ to Ident.; both tints scale with magnitude. Best per column in \textbf{bold}, second \underline{underlined}.}
\label{tab:perturbation}
\end{table}

\noindent\textbf{\textit{Detecting an altered provision is much easier than identifying what was altered.}}
Table~\ref{tab:perturbation} reports two scores per model. Detection is the share of perturbed provisions that a model flags as wrong. Strict identification is the share where the model also identifies the altered fact. Both use perturbed items only, so they differ from Table~\ref{tab:main_results}, which scores detection over all items, including the unaltered ones. The green tile beside each identification score adds the same rubric score of \S\ref{label:scoring} measured over flagged items only, so it is the identification credit each model loses to missed detections. Without grounding, detection stays high, but identification does not. \texttt{Gemini} reaches 92.22\% detection and only 19.55\% identification, and \texttt{Claude} falls from 65.37\% to 8.33\%. On average, models detect 78.9\% but identify only 18.6\%. Grounding closes most of the gap. \texttt{Claude} rises from 8.33\% to 80.17\% with RAG and 79.31\% with ICL, yielding mean identification values of 57.8\% and 49.3\%. It gains the most in both grounded settings, since it leverages contextual evidence more reliably than it recalls the same provision from memory. Its drastic parametric failure comes from naming the wrong element rather than from declining to answer. In contrast, open-weight models remain competitive. \texttt{GPT-OSS} has the highest parametric identification of all nine at 27.50\%, and \texttt{Kimi} reaches 65.53\% with RAG.

\noindent \textbf{Task Type-Wise Analysis.}
Ordered from easiest to hardest, the results show where the challenge lies. Recognition-style Temporal MCQ is easiest, reaching $91.88\%$ under ICL, and the single-clause tasks Perturbation Detection and Scenario SA follow, reaching $85.14\%$ and $94.06\%$ once grounded, because one provision is enough to answer. Difficulty then rises with the amount of temporal reasoning required, so the best ICL scores fall to $75.29\%$ on Temporal Scenario SA and $61.04\%$ on Open-Ended. The three hardest tasks each fail for a different reason. Event Sorting reaches $66.73\%$ under ICL but collapses to $31.42\%$ under RAG because the retriever misses parts of the timeline, Relative-Time QA is limited by the target date rather than the content, and Context Misalignment never exceeds $27.62\%$ because models follow a confident false premise. Task type therefore predicts performance more than model identity, although \texttt{Claude} leads the grounded Scenario SA and Relative-Time QA tasks while \texttt{GPT-5} leads Temporal Scenario SA and Open-Ended. Appendix~\ref{app:failure_mode_examples} presents qualitative analysis examples. 

\section{Conclusion}
This study introduced \texttt{\dataname{}}, a novel benchmark for evaluating large language models on \textit{evolving documents}, where the answer depends on the version in force on the queried date. Spanning 644 expert-verified customs instruments and eight task types that separate document reading from version resolution, our evaluation of nine recent LLMs under parametric, gold-context, and retrieval settings revealed several actionable insights. Persistent failures on version-sensitive tasks show that version resolution is not solved by retrieval or long-context access alone. Our contributions offer a foundation for reliable reasoning over \textit{evolving documents}.


\section*{Limitations}
While \texttt{\dataname{}} provides a controlled environment for studying version resolution in evolving documents, it has several limitations urging future research. First, we evaluate models in a frozen state using prompts, which does not clarify whether version resolution needs additional training or if frozen models can perform it with prompting alone. Second, supervised fine-tuning and continued pretraining would help identify what models cannot do versus what prompts fail to elicit. Subsequently, \texttt{\dataname{}} focuses solely on Bangladesh customs in a specific language pair and calendar system, leaving untested cross-domain generalization relevant to areas like income tax, VAT, and medical guidelines. Additionally, multilingual question answering could complicate temporal challenges further. Fifth, the parametric setting relies on unverifiable assumptions, as providers do not disclose training datasets. A low parametric score might indicate an absent document rather than a version resolution failure. Sixth, our retriever is semantic and lacks temporal re-ranking, so RAG scores are a lower bound with respect to retrieval quality. Incorporating temporally aware retrieval could narrow the gap with in-context learning \citep{zhang-etal-2025-mrag, huwiler2025versionragversionawareretrievalaugmentedgeneration}. Finally, since our corpus is set to 2025, the benchmark will become outdated with new instruments, so we plan periodic updates to include these changes.

\section*{Ethical Considerations}
All 644 instruments in \texttt{\dataname{}} are official Bangladesh customs documents published by the National Board of Revenue for public compliance. Therefore, this corpus contains no personally identifiable or sensitive information. The personas used in our scenario questions are fictional and do not represent real importers, agents, or taxpayers. \texttt{\dataname{}} is intended purely for research purposes, while the official gazette serves as the authoritative source for any duty rates or effective dates. Nothing in this benchmark should be considered legal advice. Additionally, our findings indicate that models can follow a confident but incorrect premise over authoritative texts. We advise caution against deploying LLMs in customs, tax, or compliance workflows without ensuring version-aware safeguards and expert review. We also disclose that an LLM assisted in parsing and generating questions, and every released pair underwent human verification by subject matter experts, who were compensated at standard rates for their time. Separately, the authors used AI writing assistance tools (e.g., ChatGPT, Claude, Gemini) solely for improving the fluency and clarity of the writing; all scientific content, ideas, experiments, and conclusions are entirely the work of the human authors, who take full responsibility for the integrity and accuracy of the content presented in this paper.

\section*{Acknowledgments}
This work was supported by the Research and Development Grant of the Improving Computer and Software Engineering Tertiary Education Project (ICSETEP), University Grants Commission of Bangladesh (No.~A-44), financed by the Asian
Development Bank and the Government of the People's Republic of Bangladesh.

\bibliography{custom}

\clearpage
\newpage

\appendix
\section{Appendix}

\subsection{Related Work}\label{app:related_works}
\paragraph{Temporal QA Benchmarks.}
Benchmarks such as TimeBench \citep{chu-etal-2024-timebench} and TRAM \citep{wang2024trambenchmarkingtemporalreasoning} test whether LLMs understand time itself, covering event order, duration, and temporal arithmetic, where even \texttt{GPT-4} stays clearly below human performance. Synthetic benchmarks remove memorization so that scores reflect reasoning rather than recall \citep{fatemi2024testtimebenchmarkevaluating, uddin-etal-2025-unseentimeqa}. Closer to our setting, \citet{tan-etal-2023-towards} evaluated TempReason in closed-book, open-book, and reasoning QA settings, an early ancestor of our unified three-setting protocol. \citet{wei-etal-2023-menatqa} built MenatQA with 2,853 samples around scope, order, and counterfactual factors, showing that LLMs are highly sensitive to perturbed temporal context. Likewise, \citet{liška2022streamingqabenchmarkadaptationnew} compared closed-book and open-book models on StreamingQA over fourteen years of time-stamped news. Recent benchmarks target knowledge that keeps evolving after training. \citet{zhu-etal-2025-evolvebench} tested whether models notice temporally misaligned contexts and refuse invalid timestamps; all 15 evaluated LLMs struggled. \citet{nakshatri2025factschangeprobingllms} built evolveQA from time-stamped AWS, Azure, and WHO reports and observed accuracy drops of 6\% to 31\% on evolving facts, with models often giving outdated open-ended answers despite choosing the current answer in multiple-choice format. \citet{lin-etal-2025-dynaquest} generated DynaQuest from Wikipedia infobox changes and balanced retrieved and parametric knowledge with reinforcement learning, while \citet{kim-etal-2026-large} showed on OAKS that models fail to track facts that supersede earlier ones in a stream. Further resources track evolving Wikipedia knowledge \citep{jang2023temporalwikilifelongbenchmarktraining, mousavi2024dyknowdynamicallyverifyingtimesensitive, tang2024evowikievaluatingllmsevolving}, and \citet{jiang2026benchmarksagetemporalmisalignment} showed that static factuality benchmarks age and unfairly penalize up-to-date models. Across these works, temporal change means drifting world facts or story states from English Wikipedia, news, or synthetic sources. None models formal amendment, where a dated instrument replaces an older rule from an explicit effective date while the old rule stays correct for its own validity period, and none verifies its questions with domain experts.
\paragraph{Cross-Lingual QA and Temporal Retrieval.}
\citet{asai2021xorqacrosslingualopenretrieval} introduced XOR QA, where questions in seven typologically diverse languages, including Bengali, are answered from English Wikipedia, showing that crossing the language boundary makes both retrieval and answering much harder. \citet{longpre-etal-2021-mkqa} aligned 10k QA pairs across 26 languages in MKQA and found low-resource languages hardest. Both target open-domain encyclopedic facts over monolingual evidence, without temporal or robustness settings. On the retrieval side, \citet{zhang-etal-2025-mrag} showed that off-the-shelf retrievers fail on temporally perturbed questions and proposed a semantic-temporal hybrid ranking to recover accuracy.
Across prior work, legal and regulatory documents are absent, temporal change is never a formal amendment, robustness settings such as perturbed or misaligned contexts appear only in isolation, unified parametric versus grounded comparisons are rare, and expert verification is uncommon.

\subsection{Additional Analysis of the Dataset}
\label{app:dataset_analysis}
 
This appendix reports two aspects of \texttt{\dataname{}}. Section~\ref{app:licensing} states where the source documents come from, under what terms we collected them, and how we license the release. Section~\ref{app:doc_complexity} then quantifies the structural and linguistic complexity of those documents.
 
\subsubsection{Licensing and Terms of Use}
\label{app:licensing}
 
\paragraph{Provenance.}
All 644 instruments in \texttt{\dataname{}} come from the public regulations area of the National Board of Revenue (NBR) portal, which is the official publisher of Bangladeshi customs, value added tax, and income tax instruments.\footnote{\url{https://nbr.gov.bd/regulations/acts/customs-acts/eng}} The portal lists Acts, Rules, SROs, and General Orders as downloadable PDF files. Access is free, and it requires no account, no payment, and no API key. We fetched only these public files, we followed the published download links rather than any internal endpoint, and we rate-limited our requests so that the collection placed no unusual load on the server. 
 
\paragraph{Terms of the source portal.}
The portal publishes a terms-of-use page, posted in May 2012, that applies to visitors of the site. Three points matter for a dataset release. First, the material is offered without warranty of any kind, and accuracy is explicitly among the excluded warranties. Second, the page states that the material on the portal is subject to copyright, and it treats the NBR name, images, and logos as proprietary marks that may not be copied without prior approval. Third, the page permits direct linking to hosted pages without permission. Moreover, the page frames the portal as a resource for individual use and viewing, and it grants no bulk redistribution licence.
 
\paragraph{Permission from the source authority.}
Because the site terms grant no redistribution licence, we did not rely on them. We approached the National Board of Revenue directly, and the Board granted us written permission to use the instruments for academic research and to release the resulting dataset. The permission is limited to research purposes, therefore our release carries the same limitation. This permission is the legal basis of the release. In addition, Bangladeshi copyright law treats official texts as Government works and exempts the reproduction of certain Government works from infringement unless such reproduction is prohibited, and the Right to Information Act 2009 obliges public authorities to publish this class of information.\footnote{Copyright Act 2023, which repealed the Copyright Act 2000. The instruments are additionally mirrored on the Ministry of Law portal, \url{http://bdlaws.minlaw.gov.bd/}.} 
 
\paragraph{How we license the release.}
The research-only permission governs the whole release, therefore we do not attach a licence that would allow commercial reuse. We release the data under CC BY-NC-SA 4.0, and we release the pipeline code under the MIT licence. 
The release has two layers of provenance. The annotation layer is our own contribution, and it covers clause identifiers, entity--concept pairs, canonical entity names, concept threads, timelines, and all question--answer pairs. In contrast, the excerpt layer holds verbatim clause text, which remains Government of Bangladesh material and is redistributed under the permission described above. Every record carries its source file and retrieval link, so a user can check any item against the authoritative document.

\subsubsection{Complexity of Source Documents}
\label{app:doc_complexity}
 
\begin{figure}[!ht]
    \centering
    \includegraphics[width=0.95\linewidth]{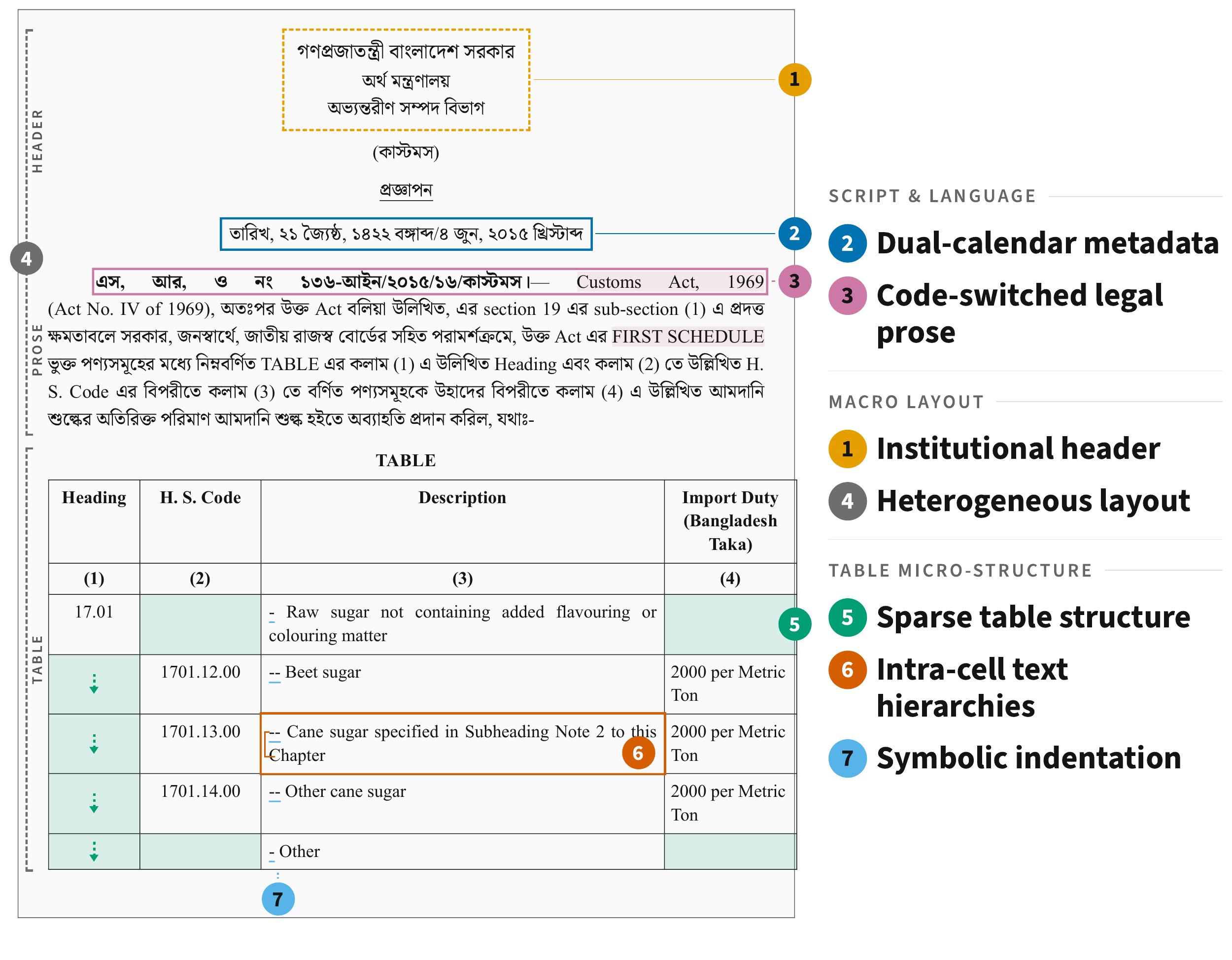}
    \caption{Beyond the challenges of temporal reasoning, the documents
    in the \texttt{\dataname{}} benchmark are inherently complex, both
    structurally and linguistically. Comprising mixed Bangla--English
    customs acts, notifications, and gazettes, they require models to
    parse multi-line institutional headers (\circnum{1}), dual-calendar
    dates in Bengali numerals (\circnum{2}), dense intra-sentential
    code-switching (\circnum{3}), heterogeneous
    layouts (\circnum{4}), sparse tables whose empty cells inherit
    parent values (\circnum{5}), nested intra-cell hierarchies
    (\circnum{6}), and symbolic indentation that encodes taxonomic
    depth rather than noise (\circnum{7}).}
    \label{fig:motivational_figure_v1}
\end{figure}

Beyond temporal reasoning, the \texttt{\dataname{}} source documents are structurally and linguistically complex, as Figure~\ref{fig:motivational_figure_v1} illustrates qualitatively. This section quantifies that complexity over the 644 official instruments in the corpus.

\paragraph{Scale and heterogeneity.}
The corpus spans 4,131 OCR pages, with a median of only 2 pages per document but a maximum of 169. Document length is likewise very uneven. The longest instrument holds 50,829 words against a median of 360, a spread of about 141 times, and the coefficient of variation of length reaches 2.6. This follows from the instrument mix, because SROs make up 95\% of the corpus and are short tariff notifications, whereas the three Acts are book-length statutes. Consequently, no single prompt template fits every document shape.
 
\paragraph{Code-mixing and mixed notation.}
The documents are written in Bengali but saturated with English legal vocabulary, so the two scripts interleave within the same sentence. English accounts for 37.0\% of alphabetic characters and 37.9\% of words, and 96.4\% of documents contain both scripts, of which 80.1\% are genuinely bilingual with the minority script above 5\% of words. This is bilingual text rather than Bengali with loanwords. Moreover, the notation is doubly mixed, since 91.7\% of documents use both Bengali and ASCII digits, and 97.0\% state each date in both the Bangla and the Gregorian era. Therefore, a model must normalize scripts, digits, and calendars before it can compare two dates.
 
\paragraph{Domain density.}

\begin{table}[t]
\centering
\small
\setlength{\tabcolsep}{2pt}
\begin{tabular}{lccccc}
\toprule
\rowcolor{headshade} \textbf{Type} & \textbf{Docs} & \textbf{Med.\ words} & \textbf{Max words} & \textbf{Eng.\ \%} & \textbf{Tariff \%} \\
\midrule
Acts  & 3   & 42,585 & 50,829 & 89.1 & 0.0 \\
Rules & 12  & 2,711  & 21,926 & 9.2  & 0.0 \\
GOs   & 20  & 641    & 26,428 & 4.2  & 10.0 \\
SROs  & 609 & 333    & 18,291 & 13.8 & 54.4 \\
\bottomrule
\end{tabular}
\caption{Per-instrument-type complexity in \texttt{\dataname{}}. \emph{Eng.\ \%} is the per-document median share of English words, and \emph{Tariff \%} is the share of documents with tabular tariff content, proxied by three or more H.S. codes. The three Acts are two parallel monolingual versions of the Customs Act 1969 plus one mixed amendment, so their median English share is not a typical value.}
\label{tab:doc_complexity}
\end{table}

The text is also dense with domain references, as 98.6\% of documents cite a section and 63.6\% contain H.S. tariff codes, for 27,829 H.S. code occurrences across the corpus. Furthermore, 51.7\% of documents carry tabular tariff content, which OCR flattens into linear text and thereby breaks the alignment between codes and rates. These dimensions vary sharply by instrument type, as Table~\ref{tab:doc_complexity} shows. Acts and Rules are long and prose-heavy, in contrast to SROs, which are short but numerically dense and highly tabular. A benchmark drawn from all four types therefore forces a model to handle both long statutory prose and terse tariff notifications.

\subsection{Expert Verification of QA Pairs}\label{app:expert_verification}
Two customs subject-matter experts (SMEs) independently review every generated pair for factual and temporal correctness. Each expert rates a pair on a three-point Likert scale, defined as follows. A score of $3$ (\emph{correct}) means the answer is factually accurate and, where the item requires a date, names the version in force and its effective date at the required granularity. A score of $2$ (\emph{partially correct}) means the core fact is right but a supporting detail is imprecise or incomplete, for example a missing month on a date-level item or one omitted value in a multi-value answer. A score of $1$ (\emph{incorrect}) means the answer asserts a wrong fact or a wrong or contradicting date. A pair counts as factually correct only when it is rated $3$. The two experts also repair the corpus as they review it. A pair rated $2$ is corrected in place, so the imprecise or missing detail is fixed and the pair is retained. A pair rated $1$ is discarded, and the experts write a new pair on the same thread theme to replace it. The two experts agree strongly, with Cohen's $\kappa = 0.91$ over the full set of pairs. Since the rubric turns on what each task must get right, Table~\ref{tab:verify_rubric} states the criterion for a score of $3$ per task type.
 
\begin{table}[!ht]
  \centering
  \begin{adjustbox}{width=0.47\textwidth}
    \begin{tabular}{p{0.28\textwidth} p{0.64\textwidth}}
      \toprule
      \rowcolor{headshade} \textbf{Task Type} & \textbf{A Score of 3 Requires} \\
      \midrule
      Temporal MCQ & The selected option matches the gold instrument, code, or date exactly. \\
      \addlinespace
      Event sorting & The full predicted ordering matches the gold chronology. \\
      \addlinespace
      Perturbation detection & The verdict is correct, and an inaccurate statement has its changed value named and corrected. \\
      \addlinespace
      Scenario short answer & The extracted value, such as a duty rate, matches the gold value. \\
      \addlinespace
      Temporal scenario short answer & The amendment is resolved correctly, with the right superseding instrument and effective date. \\
      \addlinespace
      Relative-time QA & The resolved target date matches at the required granularity, and the identified change is correct. \\
      \addlinespace
      Open-ended & The stated rationale matches the gold rationale in substance. \\
      \addlinespace
      Context misalignment & The answer follows the gold text and ignores the misleading premise. \\
      \bottomrule
    \end{tabular}
  \end{adjustbox}
  \caption{Verification rubric. For each task type, the criterion that a pair must meet to earn the top score of $3$ on the three-point scale. Scores of $2$ and $1$ follow the general definitions of \emph{partially correct} and \emph{incorrect}.}
  \label{tab:verify_rubric}
\end{table}
 
Table~\ref{tab:verify_results} reports the score distribution by task type, averaged over the two experts. Along the scale, the overwhelming majority of pairs are fully correct, only a small fraction are partially correct, and incorrect pairs are rare and never exceed $2.8\%$ of any task. Across task types, correctness is highest on the constrained tasks, where the target is a single value or a single ordering, and lowest on temporal scenario short answer and open-ended, which demand reasoning over several dated instruments or an open rationale. In aggregate, $94.89\%$ of pairs are rated correct on the first pass. After the repairs above, every retained or rewritten pair meets the score-$3$ criterion, which is what makes the benchmark gold data.
 
\begin{table}[!ht]
  \centering
  \begin{adjustbox}{width=0.47\textwidth}
    \begin{tabular}{p{0.30\textwidth} r r r r r}
      \toprule
      \rowcolor{headshade} \textbf{Task Type} & \textbf{n} & \textbf{3 (\%)} & \textbf{2 (\%)} & \textbf{1 (\%)} & \textbf{Mean} \\
      \midrule
      Temporal MCQ & 354 & 97.5 & 1.4 & 1.1 & 2.96 \\
      Event sorting & 207 & 96.3 & 2.3 & 1.4 & 2.95 \\
      Perturbation detection & 498 & 95.6 & 2.6 & 1.8 & 2.94 \\
      Scenario short answer & 562 & 96.4 & 2.4 & 1.2 & 2.95 \\
      Temporal scenario short answer & 363 & 92.3 & 4.9 & 2.8 & 2.90 \\
      Relative-time QA & 463 & 94.0 & 3.8 & 2.2 & 2.92 \\
      Open-ended & 487 & 92.9 & 5.1 & 2.0 & 2.91 \\
      Context misalignment & 116 & 94.1 & 3.8 & 2.1 & 2.92 \\
      \midrule
      \rowcolor{gray!15} \textbf{Overall} & \textbf{3050} & \textbf{94.89} & \textbf{3.31} & \textbf{1.80} & \textbf{2.93} \\
      \bottomrule
    \end{tabular}
  \end{adjustbox}
  \caption{Expert verification results by task type on the first pass, before repair, averaged over the two SMEs. Columns $3$, $2$, and $1$ give the percentage of pairs at each level of the scale in Table~\ref{tab:verify_rubric}, and \emph{Mean} is the average score in $[1,3]$. A pair is factually correct when rated $3$. Pairs rated $2$ are then corrected in place, and pairs rated $1$ are rewritten on the same theme.}
  \label{tab:verify_results}
\end{table}

\subsection{Qualitative Examples by Task Type}
\label{app:qualitative_examples}

This appendix presents two representative examples for each of the eight task types, drawn verbatim from the released benchmark, with one colored panel per type. The examples illustrate the properties that the generation pipeline of Section~\ref{subsec:qa-generation} enforces by construction and that the expert review of Appendix~\ref{app:expert_verification} confirms, where $94.89\%$ of pairs are rated factually correct with strong inter-expert agreement. Each panel opens with the property it illustrates, so the examples can be read as evidence for the corresponding design claim of Section~\ref{sec:task_types}.

\begin{tcolorbox}[expanel={blue}{Temporal MCQ ($y_{\mathrm{mcq}}$)}]
\small
\textit{Illustrates: the distractors are lexically close instruments that differ only in number or date, so surface matching fails, and version resolution is required.}

\exhead{Example 1.}
\exfield{Question}{Which instrument introduced the requirement for customs offices to use methods like electronic flagging or the AEO Accreditation Logo for priority dispute resolution?}
\exlabel{Options}
\begin{itemize}\setlength{\itemsep}{0pt}
  \item[(a)] AEO Rules 2024 (2024-06-13)
  \item[(b)] AEO Rules Amendment (2025-02-17)
  \item[(c)] AEO Operational Guidelines (2025-01-21)
  \item[(d)] AEO Rules Amendment (2025-02-16)
\end{itemize}
\exfield{Gold answer}{(c) AEO Operational Guidelines (2025-01-21)}
\exfield{Rationale}{The 2025-01-21 guidelines mandated that customs offices prioritize accredited entities in disputes and implement identification methods like the AEO Logo or electronic flagging.}

\exhead{Example 2.}
\exfield{Question}{In which year did the government reduce the duty rate for specific exempted categories from 3\% to 2\%?}
\exlabel{Options}
\begin{itemize}\setlength{\itemsep}{0pt}
  \item[(a)] 2011
  \item[(b)] 2012
  \item[(c)] 2013
  \item[(d)] 2014
\end{itemize}
\exfield{Gold answer}{(c) 2013}
\exfield{Rationale}{SRO 145-AIN/2013, dated 2013-06-06, explicitly amended the previous SRO 157-AIN/2012 to replace the 3\% duty rate with 2\%.}
\end{tcolorbox}

\begin{tcolorbox}[expanel={green}{Event Sorting ($y_{\mathrm{sort}}$)}]
\small
\textit{Illustrates: the events carry no dates or ordering cues, so the chronology must be recovered from what each change did.}

\exhead{Example 1.}
\exfield{Instruction}{Arrange these events in chronological order (earliest first).}
\exlabel{Events (shuffled, dates removed)}
\begin{itemize}\setlength{\itemsep}{0pt}
  \item[E1] The primary baggage allowance for passengers is reduced to 36kg.
  \item[E2] The baggage rules are updated to introduce a specific 40kg weight limit for passengers under twelve years of age.
  \item[E3] The baggage rules establish a 65kg primary allowance and a 35kg secondary allowance for clothing and personal items.
  \item[E4] The language regarding baggage composition is refined to specify that the 65kg allowance applies to total baggage brought by any means.
\end{itemize}
\exlabel{Gold chronological order}
\begin{enumerate}\setlength{\itemsep}{0pt}
  \item \textit{2011-06-09} --- The baggage rules establish a 65kg primary allowance and a 35kg secondary allowance for clothing and personal items.
  \item \textit{2016-04-08} --- The baggage rules are updated to introduce a specific 40kg weight limit for passengers under twelve years of age.
  \item \textit{2016-06-02} --- The language regarding baggage composition is refined to specify that the 65kg allowance applies to total baggage brought by any means.
  \item \textit{2025-06-03} --- The primary baggage allowance for passengers is reduced to 36kg.
\end{enumerate}

\exhead{Example 2.}
\exfield{Instruction}{Arrange these events in chronological order (earliest first).}
\exlabel{Events (shuffled, dates removed)}
\begin{itemize}\setlength{\itemsep}{0pt}
  \item[E1] The use of technical explanatory notes for machinery classification was mandated for duty purposes.
  \item[E2] Large-scale industries were granted flexibility to extend import timelines for machinery up to three years.
  \item[E3] The government established a concessional duty threshold of one percent for industrial machinery.
  \item[E4] The concessional duty threshold was adjusted to three percent, with an expanded scope including specific components.
  \item[E5] The legal basis for duty concessions was transitioned to the Customs Act.
\end{itemize}
\exlabel{Gold chronological order}
\begin{enumerate}\setlength{\itemsep}{0pt}
  \item \textit{2010-06-10} --- The government established a concessional duty threshold of one percent for industrial machinery.
  \item \textit{2011-06-09} --- The concessional duty threshold was adjusted to three percent, with an expanded scope including specific components.
  \item \textit{2016-06-02} --- Large-scale industries were granted flexibility to extend import timelines for machinery up to three years.
  \item \textit{2020-06-11} --- The use of technical explanatory notes for machinery classification was mandated for duty purposes.
  \item \textit{2025-06-02} --- The legal basis for duty concessions was transitioned to the Customs Act.
\end{enumerate}
\end{tcolorbox}

\begin{tcolorbox}[expanel={red}{Perturbation Detection ($y_{\mathrm{pert}}$)}]
\small
\textit{Illustrates: each statement differs from the regulation in exactly one value, and the gold answer names and corrects that value.}

\exhead{Example 1.}
\exfield{Statement}{Under SRO 175/2024, pharmaceutical raw materials listed in TABLE-2 are granted a customs duty exemption provided the existing duty exceeds a threshold of 5\%.}
\exfield{Question}{Is this statement accurate? If not, identify and correct the error.}
\exfield{Gold answer}{The statement is inaccurate. Per SRO 175/2024 (dated June 6, 2024), the duty exemption for pharmaceutical raw materials in TABLE-2 applies if the existing duty exceeds 10\%, whereas the 5\% threshold applies to TABLE-1.}

\exhead{Example 2.}
\exfield{Statement}{Under the AEO Rules 2024, the formalization of rejection criteria for tier promotion and cancellation due to non-compliance was introduced on 2025-02-17.}
\exfield{Question}{Is this statement accurate? If not, identify and correct the error.}
\exfield{Gold answer}{The statement is inaccurate. The formalization of rejection criteria for tier promotion and cancellation was introduced on 2025-02-16, whereas 2025-02-17 focused on the introduction of explicit tier-based promotion and demotion mechanisms.}
\end{tcolorbox}

\begin{tcolorbox}[expanel={orange}{Scenario Short Answer ($y_{\mathrm{ssa}}$)}]
\small
\textit{Illustrates: the queried value is embedded in a realistic persona and setting, and the scenario never quotes the clause that answers it.}

\exhead{Example 1.}
\exfield{Scenario}{A local manufacturer is expanding their facility in 2023 to begin domestic production of electric switches and sockets. To remain competitive, they are planning to import specialized components from abroad. Before placing their first order, the management team needs to understand the specific regulatory framework that allows for duty concessions on these industrial inputs. They are particularly concerned with identifying the legal basis for these exemptions and the conditions they must meet to qualify.}
\exfield{Question}{What is the legal basis and the specific condition for duty concessions on these industrial imports?}
\exfield{Gold answer}{The import is governed by Section 19(1) of the Customs Act 1969 and Section 126(1) of the VAT and Supplementary Duty Act 2012, allowing for duty exemptions where the import duty exceeds 5\% ad valorem, subject to conditions prescribed for local manufacturing.}

\exhead{Example 2.}
\exfield{Scenario}{A long-time manufacturing executive whose company recently achieved Class-1 Authorized Economic Operator (AEO) status is reviewing their supply chain efficiency for February 2025. They have invested heavily in compliance and secure logistics to earn this trusted partnership with customs. As they plan their next high-volume import, they want to understand the exact regulatory benefits they are entitled to regarding physical intervention. Specifically, they need to know the upper limit of the selectivity rate for physical examination applied to their shipments.}
\exfield{Question}{What is the maximum selectivity rate for physical examination for a Class-1 AEO as of February 2025?}
\exfield{Gold answer}{The selectivity rate for a Class-1 AEO is set at a maximum of 50\%.}
\end{tcolorbox}

\begin{tcolorbox}[expanel={teal}{Temporal Scenario Short Answer ($y_{\mathrm{tssa}}$)}]
\small
\textit{Illustrates: the answer requires composing several dated instruments, including repeals that nonetheless preserve an earlier threshold.}

\exhead{Example 1.}
\exfield{Scenario}{A Bangladeshi flight crew member has been working on international flights since 2012. They recall that when they started, they were allowed to bring in baggage worth 100 USD duty-free. They continued their service through the updates in 2016, 2023, and 2024, and are now planning a return flight in late 2025.}
\exfield{Question}{How has the duty-free baggage allowance for this flight crew member changed from their start date in 2012 to the current regulation in 2025?}
\exfield{Gold answer}{The duty-free allowance increased from 100 USD (2012) to 300 USD (2016, 2023, 2024), and was subsequently reduced to 150 USD in 2025.}

\exhead{Example 2.}
\exfield{Scenario}{An industrial manufacturer has been importing machinery since 2010 under the original 1\% excess duty threshold established by SRO 178/2010. In 2016, they observed that SRO 151/2016 repealed the 2010 order. The manufacturer wants to know if their eligibility for the 1\% threshold for machinery and spare parts remains consistent despite this change.}
\exfield{Question}{Does the repeal of SRO 178/2010 by SRO 151/2016 in 2016 remove the 1\% excess duty threshold for industrial machinery, or does the new instrument maintain it?}
\exfield{Gold answer}{The 1\% excess threshold is maintained; SRO 151/2016 explicitly repealed SRO 178/2010 while simultaneously codifying the 1\% excess duty threshold for the specified industrial goods.}
\end{tcolorbox}

\begin{tcolorbox}[expanel={violet}{Relative-Time QA ($y_{\mathrm{rel}}$)}]
\small
\textit{Illustrates: the target date is never stated, so the anchor and offset must be resolved before the regulatory change can be identified.}

\exhead{Example 1.}
\exfield{Question}{In 2016-06-02, SRO 160/2016 formalized VAT-7 form submissions. What major change regarding VAT registration and electronic verification was mandated six years later?}
\exfield{Gold answer}{Six years later, on 2022-06-09, SRO 118/2022 mandated that entities must hold a 13-digit BIN and undergo electronic verification of monthly VAT returns via online systems as a prerequisite for customs clearance.}

\exhead{Example 2.}
\exfield{Question}{In 2010-06-10, the Government relied on Section 19(1) of the Customs Act 1969 to issue notifications; five years later, what legal framework was additionally cited in conjunction with the Customs Act 1969 for issuing customs notifications?}
\exfield{Gold answer}{By 2015-06-04 (SRO 140-Ain/2015), the Government began reading Section 19(1) of the Customs Act 1969 in conjunction with Section 14(1) of the Value Added Tax Act, 1991 to issue or amend customs notifications.}
\end{tcolorbox}

\begin{tcolorbox}[expanel={brown}{Open-Ended ($y_{\mathrm{oe}}$)}]
\small
\textit{Illustrates: the gold answer traces a rationale and its evolution across instruments rather than a single value.}

\exhead{Example 1.}
\exfield{Question}{How did the minimum valuation framework and documentation requirements for importing motor vehicles under Section 25(3) of the Customs Act 1969 evolve between June 2010 and June 2012?}
\exfield{Gold answer}{In June 2010, S.R.O. 190-AIN/2010 established minimum valuation conditions specifically for new motor vehicles under Heading 87.03, requiring a 'Manufacturer's Price Certificate' prior to assessment. By June 2011, this was replaced by S.R.O. 168-AIN/2011, which expanded the scope to include double cabin pick-ups and other vehicles under Heading 87.04, and shifted the mandatory documentation requirement to a 'Manufacturer's invoice'. This structure (Heading 87.03 and 87.04 with a Manufacturer's invoice) was maintained in the subsequent June 2012 update under S.R.O. 163-AIN/2012, which repealed the intervening S.R.O. 298-AIN/2011.}

\exhead{Example 2.}
\exfield{Question}{How has the regulatory approach to handling repealed minimum value notifications evolved between 2020 and 2022?}
\exfield{Gold answer}{Starting with SRO 132/2020 and continuing through SRO 133/2021 and SRO 147/2022, the government formalized a requirement that even when a previous minimum value notification is repealed, that historical price must still be considered when determining the dutiable value under the Customs Valuation (Determination of Value of Imported Goods) Rules 2000.}
\end{tcolorbox}

\begin{tcolorbox}[expanel={magenta}{Context Misalignment ($y_{\mathrm{cm}}$)}]
\small
\textit{Illustrates: the premise is fluent but wrong, and the gold answer corrects it from the official timeline.}

\exhead{Example 1.}
\exfield{Misaligned context}{In 2017, the National Board of Revenue issued SRO 132-AIN/2017 to establish the legal framework for LED lamp and energy-saving lamp manufacturing concessions under the Customs Act 2023.}
\exfield{Question}{Does the 2017 notification (SRO 132-AIN/2017) provide the legal basis for LED lamp manufacturing concessions under the Customs Act 2023?}
\exfield{Gold answer}{No. According to the timeline, SRO 132-AIN/2017, issued on 2017-06-01, was exclusively for cellular phone manufacturing and assembly. LED and energy-saving lamp manufacturing concessions were only introduced later via SRO 201-AIN/2022, and the legal basis was updated to the Customs Act 2023 only in 2024 via SRO 173-AIN/2024.}

\exhead{Example 2.}
\exfield{Misaligned context}{On June 2, 2016, the government issued SRO 169-AIN/2016 to modify Table-1 of SRO 143-AIN/2009 regarding chemical product classifications.}
\exfield{Question}{Does the description of SRO 169-AIN/2016 align with the established regulatory history?}
\exfield{Gold answer}{No, this is incorrect. SRO 169-AIN/2016, issued on June 2, 2016, amended SRO 11-AIN/2005. Amendments to SRO 143-AIN/2009 on that same date were handled by SRO 168-AIN/2016.}
\end{tcolorbox}

\subsection{Model Details and Knowledge Cut-offs}
We evaluate a diverse set of large language models, spanning both open-weight and proprietary systems, to ensure our findings generalize across model families and access types. Table~\ref{tab:models} summarizes these models along with their parameter counts, release dates, and reported knowledge cut-offs. Knowledge cut-off dates are particularly important for our evaluation, as several of our benchmark categories directly test temporal reasoning and awareness of time-sensitive information. 

\begin{table}[!ht]
\centering
\small
\setlength{\tabcolsep}{2pt}
\renewcommand{\arraystretch}{1.15}
\begin{tabular}{@{}lcl>{\columncolor{cutshade}}c@{}}
\toprule
\rowcolor{headshade}
\textbf{Model} & \textbf{Params} & \textbf{Release} & \textbf{Cut-off} \\
\midrule
\rowcolor{groupshade}
\multicolumn{4}{@{}l}{\textsc{Open-weight}} \\
\texttt{Qwen 3.6}   & 35B (3B act.)    & Apr 2026 & \textcolor{gray}{n.d.} \\
\texttt{GPT-OSS}      & 117B (5.1B act.) & Aug 2025 & \textcolor{gray}{n.d.} \\
\texttt{Gemma-4}       & 31B (dense)      & Apr 2026 & \textcolor{gray}{n.d.} \\
\texttt{Kimi-K2.5}         & 1T (32B act.)    & Jan 2026 & \textcolor{gray}{n.d.} \\
\texttt{GLM-4.6}           & 355B (32B act.)  & Sep 2025 & \textcolor{gray}{n.d.} \\
\texttt{DeepSeek-V4-Flash} & 284B (13B act.)  & Apr 2026 & \textcolor{gray}{n.d.} \\
\addlinespace[2pt]
\rowcolor{groupshade}
\multicolumn{4}{@{}l}{\textsc{Proprietary}} \\
\texttt{GPT-5.2}           & \textcolor{gray}{n.d.} & Dec 2025 & Aug 2025 \\
\texttt{Gemini-3.5-Flash}  & \textcolor{gray}{n.d.} & May 2026 & Jan 2025 \\
\texttt{Claude-Sonnet-4.5} & \textcolor{gray}{n.d.} & Sep 2025 & Jan 2025$^{\ast}$ \\
\bottomrule
\end{tabular}
\caption{Evaluated models grouped by access type. Params reports
total parameters, and active parameters per token for
Mixture-of-Experts models. The shaded column lists the officially
reported knowledge cut-off. n.d. marks values not disclosed by the
provider. $^{\ast}$Anthropic reports a reliable knowledge cut-off of
January 2025 and a training data cut-off of July 2025.}
\label{tab:models}
\end{table}

However, not all providers disclose this information. We mark such cases as not disclosed (n.d.) in Table~\ref{tab:models}. Where a provider distinguishes between a reliable knowledge cut-off and a training data cut-off, as is the case with \texttt{Claude-Sonnet-4.5}, we report the reliable cut-off and note the distinction in the table caption. All experiments were run through two paid billing accounts. Specifically, every
\texttt{Gemini} call, both as a subject model and as a council judge, was served by the Google AI Studio Gemini Developer API, whereas all remaining subject models, judges, and embedding retrievers were served through OpenRouter under a single OpenAI-compatible endpoint, so that prompts, decoding parameters, and retry
logic stayed constant across model families. In total, building and evaluating
\texttt{\dataname{}} cost around \$1{,}100, or roughly \$0.0045 per item per model per run.

\subsection{More Details on Evaluation Metrics}\label{app:more_eval_metric}
Table \ref{tab:temporal_rubric} summarizes the scoring rubric used across the pipeline.
\begin{table}[!ht]
  \centering
  \begin{adjustbox}{width=0.47\textwidth}
    \begin{tabular}{p{0.26\textwidth} p{0.32\textwidth} p{0.36\textwidth}}
      \toprule
      \rowcolor{headshade} \textbf{Task Type} & \textbf{How It Is Scored} & \textbf{What Forces a Zero} \\
      \midrule
      \rowcolor{gray!10} \multicolumn{3}{l}{\textbf{Council-judged tasks} \quad \textit{scored by code from Eq.~\ref{eq:base_rubric}}} \\
      \addlinespace
      Scenario, temporal scenario, \& open-ended short answer &
      Base rubric. Items with no date are scored on meaning alone. &
      A wrong date on a one-date item. A contradicting date also cancels the meaning credit. \\
      \addlinespace
      Relative-time QA &
      Base rubric. The offset sets how exact the date must be: year, full date, or year and month. &
      Same as above, checked only at the required level of exactness. \\
      \addlinespace
      Context misalignment &
      Base rubric against the gold answer. &
      Same as above. \\
      \addlinespace
      Perturbation detection (explanation) &
      Base rubric, but only if the changed part is correctly pointed out. &
      Missing the changed part. Raising a false alarm. \\
      \midrule
      \rowcolor{gray!10} \multicolumn{3}{l}{\textbf{Deterministic tasks} \quad \textit{matched directly against the ground truth}} \\
      \addlinespace
      Perturbation detection (verdict) &
      Verdict vs.\ correct verdict. Accuracy, P, R, F1. &
      \textemdash \\
      \addlinespace
      Temporal MCQ &
      Chosen option vs.\ correct option. Accuracy, macro P, R, F1. &
      \textemdash \\
      \addlinespace
      Event sorting &
      Exact match. &
      \textemdash \\
      \bottomrule
    \end{tabular}
  \end{adjustbox}
  \caption{Scoring rubric of the evaluation pipeline. For council-judged tasks, a wrong date forces a zero no matter how good the meaning is. Deterministic tasks need no council and have no gates.}
  \label{tab:temporal_rubric}
\end{table}

\paragraph{Council scoring.}
The six free-form tasks are scored by a council of three heterogeneous judges rather than a single model. Each judge does not return a numeric score. Instead, it answers a fixed set of yes/no questions in JSON about whether the answer aligns with, contradicts, or entails each canonical ground-truth date, and whether it is borderline. A deterministic program then takes the majority vote of the three judges and computes the final score with Eq.~\ref{eq:base_rubric}, which keeps the scoring reproducible and removes any single judge's scale bias. Moreover, we measure inter-judge reliability with Fleiss' kappa and obtain $\kappa \in [0.8297, 0.861]$, which indicates strong agreement among the judges.

\begin{table}[t]
  \centering
  \begin{adjustbox}{width=0.47\textwidth}
    \begin{tabular}{p{0.24\textwidth} p{0.08\textwidth} p{0.60\textwidth}}
      \toprule
      \rowcolor{headshade} \textbf{Component} & \textbf{Points} & \textbf{What the SME Checks} \\
      \midrule
      Verdict accuracy &
      0--50 &
      Did the council reach the correct decision on temporal, factual, and conclusive equivalence with the ground truth? \\
      \addlinespace
      Reasoning alignment &
      0--40 &
      Is the council's analysis logical, and does it justify the verdict against the reference dates and meaning? \\
      \addlinespace
      Clarity \& justification &
      0--10 &
      Is the explanation clear, and does it properly support the decision? \\
      \midrule
      \rowcolor{gray!15} \textbf{Total} & \textbf{100} & \\
      \bottomrule
    \end{tabular}
  \end{adjustbox}
  \caption{Human validation rubric, adapted from \citet{e-sobhani-etal-2026-mathmist}, that each SME applies to council decisions on the free-form tasks. The verdict carries the most weight, and the remaining points reward sound and clearly justified reasoning.}
  \label{tab:sme_rubric}
\end{table}

\paragraph{Human validation.}
To confirm that the council tracks expert judgment, we conduct a focused human review with two independent subject-matter experts (SMEs). We adopt the human scoring rubric of \citet{e-sobhani-etal-2026-mathmist} and adapt it to our temporal setting. As Table~\ref{tab:sme_rubric} shows, the rubric awards up to 50 points for a correct verdict, up to 40 points for logical and well-grounded reasoning, and up to 10 points for a clear justification, for a total of 100. Each SME independently rescored the council's decisions on the council-judged tasks drawn from across the whole corpus. As reported in Table~\ref{tab:sme_results}, the two experts give the council total scores of 92.93 and 95.49 out of 100. 


\begin{table}[!ht]
  \centering
  \small
  \begin{adjustbox}{width=0.47\textwidth}
    \begin{tabular}{lcccc}
      \toprule
      \rowcolor{headshade} \textbf{Reviewer} & \textbf{Verdict} & \textbf{Reasoning} & \textbf{Clarity} & \textbf{Total} \\
      \rowcolor{headshade} & \textbf{(0--50)} & \textbf{(0--40)} & \textbf{(0--10)} & \textbf{(0--100)} \\
      \midrule
      SME 1 & 46.98 & 37.95 & 8.00 & 92.93 \\
      SME 2 & 48.20 & 38.49 & 8.80 & 95.49 \\
      \midrule
      \rowcolor{gray!15} Mean & 47.59 & 38.22 & 8.40 & 94.21 \\
      \bottomrule
    \end{tabular}
  \end{adjustbox}
  \caption{Mean rubric scores that each subject-matter expert assigns to the council over the council-judged tasks across the whole corpus, following the rubric in Table~\ref{tab:sme_rubric}. The verdict is scored highest and the clarity of the justification lowest, so the council agrees with the experts on the outcome and loses only minor points on explanation.}
  \label{tab:sme_results}
\end{table}

\subsection{OCR Output Schema} \label{app:ocr_schema}
Each parsed document is serialised as one JSON record. The \texttt{document\_info} block holds document-level metadata and a summary count, \texttt{parsed\_text} holds the ordered \texttt{metadata} and \texttt{clause} items, and \texttt{tables} holds every extracted table linked back to its parent item through \texttt{parent\_text\_id}. Listing~\ref{lst:ocr_schema} shows the empty template that the prompt in Appendix~\ref{app:ocr_prompt} populates.
\begin{tcolorbox}[
  colback=violet!5!white,
  colframe=violet!40!black,
  boxrule=0.5pt,
  arc=3pt,
  left=5pt, right=5pt, top=5pt, bottom=5pt,
  enhanced,
  breakable,
  label={lst:ocr_schema}
]
\footnotesize
\begin{verbatim}
{
  "document_info": {
    "file_name": "",
    "document_type": "",
    "document_date": "",
    "total_pages": 0,
    "extraction_timestamp": "",
    "summary": {
      "total_metadata": 0,
      "total_clauses": 0,
      "total_tables": 0
    }
  },
  "parsed_text": [
    {
      "type": "metadata",
      "metadata_category": "",
      "metadata_title": "",
      "text": "",
      "page_start": 0,
      "page_end": 0,
      "text_id": "",
      "table_ids": []
    },
    {
      "type": "clause",
      "clause_number": "",
      "clause_title": "",
      "text": "",
      "subclause_hierarchy": "",
      "page_start": 0,
      "page_end": 0,
      "text_id": "",
      "table_ids": []
    }
  ],
  "tables": [
    {
      "table_id": "",
      "parent_text_id": "",
      "page_start": 0,
      "page_end": 0,
      "headers": [],
      "rows": [
        []
      ]
    }
  ]
}
\end{verbatim}
\end{tcolorbox}

\subsection{Prompt Templates} \label{app:all_prompts}
\subsubsection{OCR and Structured Extraction Prompt} \label{app:ocr_prompt}
We used this prompt to guide \texttt{Gemini-3.1-Pro} when parsing the raw customs instruments into the structured schema of Section~\ref{sec:parsing}. The prompt transcribes every page in reading order, classifies each block as document metadata or a legal clause, merges clauses and tables that cross page boundaries, and emits a single JSON record in document order. It enforces zero text loss and no translation, so the mixed Bangla--English source is preserved exactly.
\begin{tcolorbox}[
  colback=green!5!white,
  colframe=green!40!black,
  boxrule=0.5pt,
  arc=3pt,
  left=5pt, right=5pt, top=5pt, bottom=5pt,
  enhanced,
  breakable
]
\footnotesize
\begin{verbatim}
You are a legal-document transcription and
structured-extraction agent for Bangladeshi
government PDFs (SROs 1969-2025, Finance Acts,
customs/tax notifications, gazette
notifications, and international treaties or
bilateral agreements between the Government of
Bangladesh and foreign governments or
entities). Documents are in Bangla, English,
or mixed Bangla-English.

This is a government gazette document for
structured indexing. Output is a field-level
transcription for research, not republication.

Extract all visible text from the attached PDF
in reading order. For scanned or image-based
pages, perform OCR. Preserve Bangla and
English scripts exactly as printed.

Use visual structure (headings, numbering,
tables) to decide metadata vs clause
boundaries.

Follow these steps in order:
  1. EXTRACT every page - capture ALL visible
     text in reading order.
  2. CLASSIFY each text block as METADATA or
     CLAUSE.
  3. MERGE clause text that continues across
     page boundaries into a single clause
     item.
  4. MERGE table rows that span multiple pages
     into a single table entry.
  5. BUILD the JSON output with all items in
     document order.

R1  OCR FIDELITY
    - Extract ALL text exactly as printed -
      every header, footer, page number,
      footnote, section number, and body text.
    - Bangla -> Bengali Unicode. English ->
      Latin script. Mixed -> keep both scripts
      exactly as they appear.
    - Do NOT translate, transliterate,
      paraphrase, or invent text.
    - ZERO TEXT LOSS: every visible character
      MUST appear in your JSON.

R2  ITEM TYPES - exactly two: "metadata" and
"clause".
    - type field is the string "metadata" or
      "clause".
    - Never use "clauses" (plural) or any
      other value.

R3  METADATA
    - DEFINITION: Text whose function is to
      IDENTIFY, LABEL, or NAVIGATE the
      document - it describes or organises the
      document but carries no legal
      obligation, right, or definition in
      itself. A metadata block could be
      removed without changing any legal rule.
    - Includes: document titles, SRO/act
      numbers, ministry names, chapter
      headings, schedule titles, preamble
      recitals, amendment titles, page
      headers/footers, page numbers,
      footnotes, running headers, and
      attestation/signature blocks.
    - metadata_category - exactly one of:
      * "document": anything that identifies
        or structures the document at a
        content level (titles, chapter
        headings, schedule headings,
        preambles).
      * "page": anything that identifies or
        structures a physical page (headers,
        footers, page numbers, footnotes).
    - metadata_title: the heading/title text
      if this block IS a heading; otherwise
      "".
    - Metadata items NEVER have tables. tables
      field = [].

R4  CLAUSES
    - DEFINITION: A numbered unit of legal
      text whose function is to CREATE,
      MODIFY, or REMOVE a legal rule - an
      obligation, right, prohibition, penalty,
      definition, or procedural requirement. A
      clause can be independently cited and
      has normative force on its own. It would
      remain meaningful if extracted from the
      document.
    - Primary markers (use to locate clause
      boundaries):
      Bengali numeral followed by the daari
      (danda) mark.
      Arabic numeral + period: 4., 5., 10.
    - When a block carries legal force AND
      bears a clause numeral, it is always a
      clause - regardless of its visual
      formatting or position on the page.
    - Sub-clauses (a)(b)(c) and
      sub-sub-clauses (i)(ii)(iii) qualify or
      elaborate the parent clause's legal
      rule. They are ALWAYS part of the parent
      clause's text. They are NEVER separate
      items in the output.

R5  CLAUSE_NUMBER - GLOBAL SEQUENTIAL
NUMBERING
    - Format: always "clause_N" where N is an
      Arabic numeral.
    - N must be the GLOBAL sequential position
      of the clause in the ENTIRE document,
      NOT the chapter-local number.
    - Documents with multiple chapters/parts
      often restart clause numbering (e.g. a
      later chapter starts again at 1). You
      MUST continue the global count from
      where the previous chapter ended.
    - Example: if Chapters 1-5 had clauses
      1-43, then Chapter 6's first clause
      becomes clause_44, its second becomes
      clause_45, and so on.
    - The clause's original printed numeral
      still appears in the text field verbatim
      - only clause_number uses the global N.
    - Never a raw script numeral, a bare
      number ("3"), or a value without the
      prefix.

R6  CLAUSE_TITLE
    - The clause_title is the heading phrase
      between the clause numeral and the
      trailing em-dash delimiter.
    - Example:
        "4. Replacement of section 25B.--"
        -> clause_title = "Replacement of
        section 25B"
    - If no heading phrase exists before the
      body text, clause_title = "".

R7  CROSS-PAGE MERGING (CRITICAL)
    - You see the ENTIRE document. If a clause
      continues across pages, merge ALL its
      text into ONE clause item.
    - page_start = first page where the clause
      appears. page_end = last page.
    - Literal strings spanning pages: all text
      stays in the parent clause's text field.

R8  SUBCLAUSE HIERARCHY
    - subclause_hierarchy: a JSON array
      capturing the FULL nested structure of
      all sub-clauses for the entire clause.
    - Flat items = top-level sub-clauses. A
      nested sub-array IMMEDIATELY after a
      parent = that parent's children.
    - Examples:
        a has children x, y; then b:
          -> ["a", ["x", "y"], "b"]
        a has child x, which has children i,
        ii;
        then y; then b:
          -> ["a", ["x", ["i", "ii"], "y"],
          "b"]
    - Use ORIGINAL numbering signs from the
      PDF.
    - EXCLUDE sub-clause markers inside quoted
      literal strings ("..."). Those are part
      of quoted law text, not structural.
    - No sub-clauses -> empty array [].

R9  TABLE EXTRACTION
    - Tables ALWAYS belong to a CLAUSE - embed
      in the clause's tables array.
    - Detect ALL table types: bordered,
      partially bordered, and borderless
      (aligned columns of data).
    - For each table:
      * headers: array of column header
        strings as printed.
      * rows: array of arrays [[cell, cell],
        ...]. NO dict rows. Empty cells -> "".
      * page_start / page_end: pages the table
        spans.
    - Extract ALL rows - never truncate, even
      for 100+ rows.
    - MULTI-PAGE TABLES: merge all rows into
      ONE table entry.
    - Schedule tables: attach to the clause
      that introduces the schedule.
    - INLINE TABLE MARKERS (CRITICAL): In the
      clause's text field, insert the marker
      [TABLE_K] at the EXACT position where
      the K-th table appears in the document
      flow. K = 1, 2, 3... per clause,
      matching the order of objects in the
      tables array. Example: if a clause has
      text before a table, then more text,
      then a second table:
        text: "... preceding text [TABLE_1]
        middle text [TABLE_2] trailing text"
    - Place [TABLE_K] at the exact spot in the
      text where the table visually appears on
      the page. If the table appears at the
      end of the clause text, place the marker
      at the end.

R10 LITERAL STRINGS
    - Verbatim legislative text between the
      outermost "...".
    - Inside a literal string: (a)(b)(i)(ii)
      are sub-sections of QUOTED LAW. Do NOT
      add them to subclause_hierarchy. Do NOT
      create new clause items from them.

R11 DATES
    - Extract the document date from page 1
      header/metadata. Report as
      document_date.

R12 ORDERING
    - Items in the output MUST follow document
      order: top-to-bottom, page-by-page.
\end{verbatim}
\end{tcolorbox}

\subsubsection{Entity--Concept Extraction Prompt} \label{app:entity_concept}
We used this prompt to guide \texttt{Gemini-3.1-Pro} in extracting entity--concept pairs from each parsed clause and metadata element (Section~\ref{subsec:ec-extraction}). The prompt fixes canonical, cross-document English names for every legal subject, so the same entity is spelled identically wherever it appears, and it frames each concept to the document level (a foundational rule for an Act, a legal change for an SRO, a procedure for a Rule, and an operational instruction for a GO). It also reads embedded tables as entity-rich content rather than formatting noise.
\begin{tcolorbox}[
  colback=orange!5!white,
  colframe=orange!55!black,
  boxrule=0.5pt,
  arc=3pt,
  left=5pt, right=5pt, top=5pt, bottom=5pt,
  enhanced,
  breakable
]
\footnotesize
\begin{verbatim}
You are an expert legal analyst specializing
in
Bangladeshi customs, taxation, and trade law.
Your task is to extract entity-concept pairs
from a legal clause or metadata element.

===== DEFINITIONS =====

ENTITY: An entity is a canonical,
re-identifiable SUBJECT of Bangladesh customs
law - a thing that keeps the SAME identity
wherever it appears across the legal hierarchy
(the Act that defines it, the SRO that amends
it, the Rule that operationalizes it, and the
GO that gives instructions about it). It may
be a law or instrument, a tax instrument, a
government body or role, a commodity or
schedule, a process/mechanism, or a formally
defined legal term. An entity must be a noun
or noun phrase. There can be multiple entities
in a single clause.

ENTITY NAMES ARE CROSS-DOCUMENT ANCHORS -
consistency is critical:
    - Always use the canonical, official
      English name, spelled IDENTICALLY every
      time (e.g., 'Value Added Tax', never
      'VAT'; 'National Board of Revenue',
      never 'NBR'; 'Customs Act 1969', never
      'the Act').
    - Prefer the specific, registered name
      over a generic reference (e.g.,
      'Passenger (Non-Tourist) Baggage
      (Import) Rules 2012', not 'the Rules';
      the instrument's own number, e.g.,
      'S.R.O. 169-AIN/2010', not 'this
      notification').
    - If a PREFERRED CANONICAL NAMES list is
      given above and the subject matches one,
      REUSE that exact spelling.

ENTITY CATEGORIES (use these to guide
consistent naming):
    - Laws/Acts: e.g., 'Customs Act 1969',
      'Finance Act 2022', 'Right to
      Information Act 2009'
    - Tax Instruments: e.g., 'Customs Duty',
      'Value Added Tax', 'Supplementary Duty'
    - Government Bodies: e.g., 'National Board
      of Revenue', 'Customs Authority'
    - Roles: e.g., 'Commissioner of Customs',
      'Appropriate Officer of Customs'
    - Processes/Mechanisms: e.g., 'Bonded
      Warehouse Operation', 'Customs
      Clearance', 'Risk Analysis'
    - Documents/Schedules: e.g., 'First
      Schedule', 'Bangladesh Customs Tariff'
    - Infrastructure: e.g., 'Customs
      Computerised Entry Processing System'

CONCEPT: A concept is a meaningful,
descriptive explanation of what is being said
ABOUT a specific entity in this clause. It
captures the central theme, action, rule,
condition, or policy being applied to or
associated with that entity. IMPORTANT: the
KIND of claim a concept should capture depends
on the document's level (see the DOCUMENT TYPE
& FUNCTION block above, when present) - a
foundational rule/power for an ACT, a specific
legal change for an SRO, a procedure or
threshold for a RULE, and a practical
operational instruction for a GO. Frame each
concept to match that level. A concept must be
at least one complete, meaningful sentence and
at most 3-4 sentences. Concepts should be
written in simple, plain English using generic
terms (avoid citing specific monetary amounts,
exact dates, or proper names - describe them
generically instead). A single entity can have
multiple concepts if the clause discusses
multiple distinct aspects of it.

===== ENTITY NAMING - RIGHT vs WRONG =====

WRONG entity names -> RIGHT entity names:
    - 'VAT' -> 'Value Added Tax'
    - 'duty' -> 'Customs Duty'
    - 'the Act' -> 'Customs Act 1969'
    - 'NBR' -> 'National Board of Revenue'
    - a Bangla name for the law -> 'Customs
      Act 1969' (always English)
    - 'section 156' -> 'Customs Offence
      Penalties' (entity = the subject, not
      the section number)

===== SPECIAL CLAUSE RULES =====

DEFINITION CLAUSES (Definitions):
When the clause defines legal terms (typically
using numbered or lettered sub-clauses like
(a), (b), or quoted phrases followed by
'means...'), extract EACH formally defined
term as its own entity. The concept for each
defined term should paraphrase the definition
in plain English. Do not merge all definitions
into one entity.

TABLE GRANULARITY:
When the clause contains an embedded table,
extract row-level entities only when a row
introduces a distinct legal subject (e.g., a
specific offence, a commodity class, a form or
schedule, a penalty category). For
questionnaires, checklists, or compliance
requirement tables, summarize by section or
theme rather than extracting one entity per
row.

===== INSTRUCTIONS =====

1. Read the clause text below carefully.
2. The clause may be in English, Bangla, or
   mixed Bangla-English - but ALL entity and
   concept values MUST be written in English
   regardless of the source language.
3. Be STRICT and CONSISTENT in entity naming -
   always use canonical, official English
   names. Use the same form every time.
4. Be DESCRIPTIVE in concept formation - each
   concept must be a meaningful sentence (1-4
   sentences) that clearly explains the legal
   provision or rule being applied to the
   entity.
5. TABLES ARE ENTITY-RICH: When the clause
   text contains embedded table data
   (pipe-separated rows), treat the table
   content as equally important to the
   surrounding clause text. Tables in legal
   clauses often introduce distinct entities
   not mentioned in the main text - such as
   specific offence types, penalty structures,
   commodity categories, systems, or section
   references. Extract entities from BOTH the
   clause text AND the table content. Ignore
   only the formatting noise (pipe characters,
   repeated headers, cell borders) but
   carefully read the semantic content of
   every table row and column.
6. If no meaningful entities can be extracted
   (e.g., the text is a page number, blank, or
   purely navigational), return an empty
   entities list.
\end{verbatim}
\end{tcolorbox}

\subsubsection{Question--Answer Generation Prompt} \label{app:qa_generation}
We used the prompt template in this section to guide \texttt{Gemini 3.1 Pro} in generating question--answer pairs from each evolving concept thread. One call produces one task type for one thread. Every call sends the same header, the same context, and exactly one task block, therefore the shared instructions and the shared output fields are stated only once. The task block adds only the definition and the extra fields of the requested type.
 
The header sets four quality rules, so a pair must name a concrete date, instrument, rate, or category, must stay answerable from the cited clauses alone, must not repeat another pair, and must follow the cited text exactly. Moreover, the header treats the per-type target as a guide instead of a quota, therefore a thin thread returns fewer pairs rather than padded ones. Finally, every pair must return the clause indices it used. We resolve these indices back to clause identifiers and source files, so each released pair carries an explicit grounding trail.
 
\begin{tcolorbox}[
  colback=blue!5!white,
  colframe=blue!55!black,
  boxrule=0.5pt,
  arc=3pt,
  left=5pt, right=5pt, top=5pt, bottom=5pt,
  enhanced,
  breakable
]
\footnotesize
\begin{verbatim}
You build a HIGH-QUALITY, exam-grade Q&A
dataset from how a Bangladeshi customs/tax
concept evolves over time. Use ONLY the facts
in the context below - never invent dates,
rates, instruments, categories, or entities.
 
===== QUALITY BAR (EVERY PAIR) =====
 
  - SPECIFIC & grounded: name the exact
    date/year, SRO/instrument number, rate,
    category/HS heading, or entity involved.
    Never vague or generic.
  - SELF-CONTAINED: fully answerable from the
    cited clauses, no outside knowledge.
  - NON-TRIVIAL & DISTINCT: no two pairs test
    the same fact. Favour pairs that require
    reading the evolution/timeline over a
    single surface fact.
  - FAITHFUL: the answer must be exactly
    supported by the cited clauses.
 
Generate AS MANY strong, distinct pairs as
the content genuinely supports (<TARGET> is a
guide; return fewer if the material is thin -
never pad with weak pairs).
 
===== OUTPUT FORMAT =====
 
Return ONLY this JSON object, with no
preamble and no code fences:
  {"qa_pairs":[{...},{...}]}
 
EVERY pair object carries these two fields:
  "type": "<the requested task block type>"
  "clause_indices": [<int>,...]
 
CRITICAL: "clause_indices" lists the clause
[index] numbers (from NUMBERED CLAUSES below)
whose content the question/answer is grounded
in. Cite the exact clauses used; never leave
it empty.
 
The task block adds its own fields on top of
these two. Emit ONLY the requested type.
 
===== CONTEXT =====
 
<THREAD CONTEXT: thread title, concept label,
 summary, EVOLUTION block (dimension: value),
 TIMELINE block (date | instrument | change)>
 
NUMBERED CLAUSES (each has a source file;
cite the ones you use):
<one entry per sampled clause, giving its
 [index], clause date, source file, entity
 names, up to three concepts, and the
 resolved clause text>
 
===== TASK BLOCK (exactly one per call) =====
 
(1) temporal_mcq  [target 6]
TEMPORAL multiple-choice (when did X change /
what held at time T / which instrument
introduced Y). 4 options, one correct,
plausible distractors.
Fields: "question", "options":["A","B","C",
"D"], "answer_index":<0-3>, "answer":
"<correct option>", "explanation":"<why, with
date/instrument>".
 
(2) event_sorting  [target 4]
Give a SHUFFLED set of 3-6 real
events/changes and ask to order them
chronologically (earliest first). The
shuffled events MUST NOT contain any date,
year, or month - no "1969", no "in 2012", no
"June", no "FY2018-19" - so the order cannot
be recovered by surface matching. Dates
appear only in the answer key.
Fields: "question":"Arrange these events in
chronological order (earliest first).",
"events_shuffled":["e1","e2","e3"],
"answer_ordered":[{"event":"e","date":
"YYYY-MM-DD"}].
 
(3) perturbation_detection  [target 6]
A statement in which exactly ONE fact is
subtly WRONG (changed date/rate/category/
instrument/scope). Ask the reader to judge
accuracy and correct it. Most perturbed; a
few may be fully correct.
Fields: "statement", "question":"Is this
statement accurate? If not, identify and
correct the error.", "is_perturbed":
<true|false>, "perturbed_fact":"<wrong part
or null>", "correct_fact", "answer":
"<verdict + correction>".
 
(4) scenario_short_answer  [target 6]
A realistic, STORY-LIKE real-world SCENARIO
(3-6 sentences: a named persona/firm/officer
in a concrete situation and year) that
naturally leads to a question, with a SHORT
factual answer about the rule/rate/
requirement that applied. Do not reveal the
answer inside the scenario.
Fields: "scenario", "question", "answer":
"<short factual answer>",
"date_or_instrument":"<relevant date/SRO>".
 
(5) temporal_scenario_short_answer [target 5]
A scenario that DEPENDS ON MULTIPLE related
clauses at DIFFERENT times, with a question
requiring TEMPORAL reasoning (before vs
after, what now applies to someone who
started under an older rule, sequence/
consequence). Only if it genuinely needs >=2
clauses across time; else fewer/none.
Fields: "scenario":"<multi-time story>",
"question", "answer":"<short factual
answer>", "clauses_involved":["<date/
instrument>"].
 
(6) relative_time_qa  [target 4]
State an explicit SEED time, then ask about a
different point via a RELATIVE offset ("six
years later", "two years earlier") WITHOUT
naming the target year in the question. The
target must be a real timeline event, and the
answer resolves it to the actual
time/instrument.
Fields: "seed_time", "relative_offset":"<e.g.
'six years later'>", "question":"<uses seed +
relative offset>", "answer":"<resolves to
actual time + fact>".
 
(7) open_ended  [target 6]
Analytical questions on HOW/WHY the concept
evolved, cross-period comparisons, drivers,
or implications. Answers are a few grounded
sentences.
Fields: "question":"<analytical q>",
"answer":"<grounded answer>".
 
(8) context_misalignment  [target 5]
Give a short MISALIGNED context passage that
mixes in a detail from the WRONG time period
or contradicts the true evolution, then a
question whose correct answer requires
RESOLVING the conflict using the true
chronology.
Fields: "misaligned_context":"<passage with a
conflict>", "question":"<q needing conflict
resolution>", "conflict":"<what is
misaligned>", "answer":"<resolution citing
correct date/instrument>".
\end{verbatim}
\end{tcolorbox}

\subsubsection{Parametric (Closed-Book) Evaluation Prompts} \label{app:parametric}
In the parametric setting (Section~\ref{sec:closed_book}), the model answers from its own knowledge with no reference documents. Every task prompt is a shared system preamble \(P\) followed by a task-specific body, so we give \(P\) once and then each body. The deterministic tasks, namely temporal MCQ, event sorting, and the perturbation verdict, additionally constrain the output to a strict JSON schema. For relative-time QA, the temporal anchor and offset are narrated inside the question text and are supplied to the judge separately, so its prompt matches open-ended QA. Perturbation detection runs in two steps, and the explanation step is issued only when the verdict step returns \texttt{INCORRECT}.
\begin{tcolorbox}[
  colback=teal!5!white,
  colframe=teal!45!black,
  boxrule=0.5pt,
  arc=3pt,
  left=5pt, right=5pt, top=5pt, bottom=5pt,
  enhanced,
  breakable
]
\footnotesize
\begin{verbatim}
Shared preamble P (prepended to every task
body below):

You are being tested on your knowledge of
Bangladesh customs regulations: SROs
(Statutory Regulatory Orders), customs and
regulatory duties, baggage rules, and trade
agreements. Answer from your own knowledge; no
reference documents are provided.

Each task prompt is P followed by the body
shown.

--- open_ended / relative_time_qa ---
Question: {question}
Give a direct, factual answer in 1-4
sentences.

--- scenario_short_answer /
temporal_scenario_short_answer ---
Scenario:
{scenario}
Question: {question}
Give a direct, factual answer in 1-4
sentences.

--- context_misalignment ---
({context} is a deliberately misaligned
passage, yet the answer is still scored
against the true ground truth.)
Context:
{context}
Question: {question}
Give a direct, factual answer in 1-4
sentences.

--- temporal_mcq (strict JSON: reasoning,
choice_index) ---
Question: {question}
Options:
{options_block}
Choose the single best option; if you are
unsure, choose the most likely one. Think
first, then answer. Respond with strict JSON
only, exactly:
{"reasoning": "<briefly weigh the options
against your knowledge>", "choice_index":
<0-based index of the chosen option>}

--- event_sorting (strict JSON:
ordered_indices) ---
Task: {question}
Events (numbered, in shuffled order):
{events_block}
Arrange ALL of the numbered events in
chronological order (earliest first). Respond
with strict JSON only, exactly:
{"ordered_indices": [<each event number
exactly once, earliest event first>]}

--- perturbation_detection, Step 1 - Verdict
(strict JSON: reasoning, verdict = CORRECT /
INCORRECT) ---
Statement:
{statement}
Is this statement factually accurate according
to the actual regulations? First check each
specific detail of the statement (values,
percentages, dates, names, authorities,
requirements) against your knowledge, then
respond with strict JSON only, exactly:
{"reasoning": "<check the statement's specific
facts against your knowledge>", "verdict":
"CORRECT" or "INCORRECT"}
Use "CORRECT" if the statement is accurate,
"INCORRECT" if it contains an error.

--- perturbation_identification, Step 2 -
Explanation (free text; only if Step 1
returned INCORRECT) ---
Statement:
{statement}
This statement contains error. Explain
precisely WHY it is incorrect and WHERE the
error lies: identify the specific wrong detail
(value, percentage, date, name, authority, or
requirement) and state what the correct fact
should be.
\end{verbatim}
\end{tcolorbox}

\subsubsection{Grounded (Document-as-Context) Evaluation Prompts}\label{app:grounded}
The grounded, in-context setting (Section~\ref{sec:gold}) prepends the resolved source documents to the parametric prompts and swaps the preamble's closing sentence from a closed-book instruction to a grounded one. The task bodies are otherwise identical to Appendix~\ref{app:parametric}, so we show only the document-context block, the grounded preamble, and the single structural change. That change is in context misalignment, where the prompt holds both the misleading passage and the authoritative source, so the setting measures context-conflict resolution rather than parametric robustness.
\begin{tcolorbox}[
  colback=gray!10!white,
  colframe=gray!60!black,
  boxrule=0.5pt,
  arc=3pt,
  left=5pt, right=5pt, top=5pt, bottom=5pt,
  enhanced,
  breakable
]
\footnotesize
\begin{verbatim}
Every prompt here is the matching parametric
prompt (previous appendix) with two changes:
(a) the resolved source document(s) are
prepended as the context block below, and (b)
the preamble's closing sentence is swapped
from the closed-book instruction to the
grounded one below. All task bodies (question
/ scenario / context / options / events /
statement) and both JSON schemas are
byte-identical to the parametric prompts.

Document-context block (prepended to every
prompt):

You are given the following official source
document(s). Use them to answer.

--- DOCUMENT 1: {document_stem} ---
{document_text}

--- DOCUMENT 2: {document_stem} ---
{document_text}
...


Grounded preamble (replaces the closed-book
closing sentence of P):

You are being tested on Bangladesh customs 
regulations: SROs (Statutory Regulatory Orders),
customs and regulatory duties, baggage rules, 
and trade agreements. Answer using the official 
source document(s) provided above; they are
authoritative.

context_misalignment (the only structural
change): the prompt contains BOTH the
misaligned {context} and the authoritative
document(s). This measures whether the model
resolves the conflict in favor of the official
document, reported as
context_conflict_resolution, and is distinct
from the closed-book parametric misalignment
robustness.
\end{verbatim}
\end{tcolorbox}

\subsubsection{Retrieval-Augmented (RAG) Evaluation Prompts}\label{app:rag_prompt}
The retrieval-augmented setting (Section~\ref{sec:rag}) repeats the grounded setting of Appendix~\ref{app:grounded} with one change. The source documents are not resolved from the gold clause identifiers of a question, but retrieved from the corpus by semantic search. Everything the model reads after retrieval, namely the document-context block, the preamble, every task body, and both strict-JSON schemas, is byte-identical to Appendix~\ref{app:grounded}. Moreover, the judge (Appendix~\ref{app:judge_prompt}) is unchanged and never sees the retrieved documents, so grading stays identical across all three settings. We therefore give only the retrieval procedure and the blocks that change.

\paragraph{Retrieval procedure.}
For each instance we embed a query with the same \texttt{gemini-embedding-2} model that built the corpus and pull the nearest clauses from the pre-built \texttt{ChromaDB} \texttt{customs\_corpus} collection. These clauses replace the gold documents and fill the context block below. The query concatenates only the task-appropriate record fields of Table~\ref{tab:rag_query}, and not the full record, so answers, options, and other leakage-prone fields stay out of retrieval. In addition, list-valued fields such as the shuffled events are flattened to text.

\paragraph{Ranking and expansion.}
We over-fetch four times the target depth and keep at most one hit per parent clause, so a chunked long clause occupies a single slot and does not crowd out ordinary single-vector clauses. The first $k = 10$ clauses after deduplication are passed to the model, ranked by cosine similarity. When a slot is won by a chunk, the rank follows the similarity of the chunk itself, whereas the prompt receives the full resolved text of the parent clause. Only the 82 oversized clauses are chunked, and all others are served whole.

\paragraph{Packing and fallback.}
Documents are packed on document boundaries, so a clause is either included whole or dropped whole and is never truncated in the middle. The character cap is disabled in the reported runs, hence all $k$ retrieved clauses are kept. Finally, if the query is empty or retrieval returns nothing, the instance is marked \texttt{none\_resolved} and the model answers with no context block, which is equivalent to the parametric body of Appendix~\ref{app:parametric}.

\begin{table}[!ht]
\centering
\footnotesize
\setlength{\tabcolsep}{4pt}
\renewcommand{\arraystretch}{1.15}
\begin{tabular}{@{}ll@{}}
\toprule
\rowcolor{headshade}
\textbf{Task type} & \textbf{Query fields used} \\
\midrule
Temporal MCQ                   & question \\
Event Sorting                  & shuffled events \\
Perturbation Detection         & statement \\
Scenario Short Answer          & scenario + question \\
Temporal Scenario Short Answer & scenario + question \\
Relative-Time QA               & question \\
Open-Ended                     & question \\
Context Misalignment           & question \\
\bottomrule
\end{tabular}
\caption{Fields concatenated into the retrieval query for each task
type. Leakage-prone fields are excluded, so the options are dropped
for temporal MCQ, the question template is dropped for event sorting,
and the misaligned passage is dropped for context misalignment.}
\label{tab:rag_query}
\end{table}

\begin{tcolorbox}[
  colback=brown!8!white,
  colframe=brown!55!black,
  boxrule=0.5pt,
  arc=3pt,
  left=5pt, right=5pt, top=5pt, bottom=5pt,
  enhanced,
  breakable
]
\footnotesize
\begin{verbatim}
Every prompt here is the matching grounded
prompt (previous appendix) with one change:
the source document(s) are retrieved from the
corpus by semantic search instead of being
resolved from the gold clause ids. All task
bodies (question / scenario / context /
options / events / statement) and both JSON
schemas are byte-identical to the parametric
prompts.

Retrieved-document context block (prepended
to every prompt):

You are given the following official source
document(s). Use them to answer.

--- DOCUMENT 1: {clause_stem} ---
{clause_text}

--- DOCUMENT 2: {clause_stem} ---
{clause_text}
...

(When a retrieved clause has no OCR text,
{clause_text} is replaced by the literal
string [This document's OCR text is empty.].
When retrieval returns nothing, the instance
is marked none_resolved and the whole block
is omitted, so the prompt equals the
parametric one.)


Grounded preamble (identical to the grounded
setting; replaces the closed-book closing
sentence of P):

You are being tested on your knowledge of
Bangladesh customs regulations: SROs
(Statutory Regulatory Orders), customs and
regulatory duties, baggage rules, and trade
agreements. Answer using the official source
document(s) provided above; they are
authoritative.

context_misalignment (the only structural
change): the prompt contains BOTH the
misaligned {context} and the retrieved
document(s). This measures whether the model
resolves the conflict in favor of the
retrieved official source, reported as
context_conflict_resolution, and is distinct
from the closed-book parametric misalignment
robustness. The misaligned passage is used
only inside the prompt body and is kept out
of the retrieval query, so it cannot steer
which clauses are retrieved.
\end{verbatim}
\end{tcolorbox}

\subsubsection{LLM-as-Judge Prompt (single council member)}\label{app:judge_prompt}
Free-form answers are graded by an LLM council of three judges (Section~\ref{eval:llm_council}). The template below is the single-judge prompt each council member receives. A judge never emits a score or verdict. Instead, it answers a fixed set of atomic yes/no questions, and a deterministic program computes the score from those booleans (Eq.~\ref{eq:base_rubric}). The same template serves both the parametric and the grounded conditions, because the judge is shown the scenario, context, and anchor but never the source documents, so grading stays identical across settings. Temporal MCQ and event sorting are scored deterministically and do not use the judge. The template has a constant header and footer, a per-item case block, and a task-specific note that is inserted only for context misalignment, perturbation explanation, and relative-time QA.
\begin{tcolorbox}[
  colback=purple!5!white,
  colframe=purple!45!black,
  boxrule=0.5pt,
  arc=3pt,
  left=5pt, right=5pt, top=5pt, bottom=5pt,
  enhanced,
  breakable
]
\footnotesize
\begin{verbatim}
===== HEADER (constant, every judge call)
=====

You are a grading assistant for a
question-answering benchmark on Bangladesh
customs regulations (SROs - Statutory
Regulatory Orders, duty rules, baggage rules,
and trade agreements).

You do NOT produce a grade, score, or verdict.
You answer a fixed set of atomic yes/no
questions about the MODEL ANSWER; a program
computes the score from your booleans.

Rules:
1. The GROUND-TRUTH ANSWER is authoritative
   and correct BY DEFINITION. Do not use your
   own knowledge or opinion to second-guess
   it.
2. The ground-truth dates listed below were
   extracted programmatically from the
   ground-truth answer. Decide ONLY whether
   the model answer commits to each of them;
   never introduce ground-truth dates of your
   own.
3. For each listed ground-truth date: matched
   = true iff the MODEL ANSWER commits to that
   date - in any language, script, or format
   (Bengali numerals, Bangla month names, "9
   June 2011", "09/06/2011", spelled-out
   ordinals, or a relative expression that
   resolves to it) - supplying at least the
   required components with matching values.
   Copy the exact span of the model answer
   that expresses the date into
   model_expression (empty string if not
   matched).
4. contradicting_dates_in_answer = true iff
   the model answer asserts a DIFFERENT date,
   at the required granularity, for a fact the
   ground truth dates - or gives multiple
   conflicting dates for the same fact without
   committing to one. Extra dates about other
   facts that do not conflict with the ground
   truth are NOT contradictions.
5. meaning_correct is TRUE only if ALL of the
   following hold, judged against the
   ground-truth answer; otherwise FALSE.
   Binary - no partial credit.
   - The model answer entails the ground
     truth's key facts and contradicts none of
     them.
   - Correct legal instrument / entity /
     section (the right SRO, Act, or clause).
   - Correct action AND direction of change -
     increased vs decreased, inserted vs
     omitted, effective vs repealed. Direction
     errors are automatically FALSE: in
     amendment QA, "raised to 15%" vs "lowered
     to 15%" is the difference between correct
     and catastrophically wrong.
   - Correct values (rates, amounts,
     thresholds) when the ground truth states
     them.
   - Paraphrase and language switching (Bangla
     <-> English) are acceptable; extra
     correct detail is acceptable.
   Judge meaning_correct on the NON-DATE
   substance of the answer; date agreement is
   captured separately by date_matches and
   must not double-count here.
6. If a judgment is genuinely borderline under
   the checklist, still commit to a boolean
   but set uncertain = true and say why in
   meaning_rationale (such items are routed to
   human review; they are never soft-scored).

===== TASK-SPECIFIC NOTES =====
(inserted after the header; only for the task
types shown)

context_misalignment:
Task note (context_misalignment): the context
shown to the model deliberately contains
misaligned or incorrect information; the
ground-truth answer remains the authority.
Additionally report these booleans about the
model's behavior (they are logged, not
scored):
- followed_context: the answer adopts the
  misaligned context's wrong value/claim.
- followed_parametric: the answer gives the
  authoritative (ground-truth) value.
- conflict_acknowledged: the answer explicitly
  notes a discrepancy between the given
  context and the actual regulation.
- blended_sources: the answer mixes the
  misaligned and authoritative values.
(If a documented conflict exists, the line
"Documented conflict for this item:
{conflict}" is appended.)

perturbation_explanation (grades Step 2; only
for items the ground truth marks as
perturbed):
Task note (perturbation explanation): the
statement below contains a deliberately
perturbed element, and the model was asked to
explain WHY and WHERE the statement is wrong.
The specific perturbed element is:
- perturbed (wrong) value in the statement:
  {perturbed_fact}
- correct value per the regulation:
  {correct_fact}
Additionally report:
- perturbation_flagged: the explanation
  asserts the statement is wrong.
- perturbation_localized: the explanation
  identifies THAT SPECIFIC element above as
  the error. Saying something is off while
  naming a DIFFERENT element is false.
- localized_element: the element the model
  identified (empty string if none).
- correction_given: the model states what the
  correct fact should be.
- correction_correct: the model's stated
  correction matches the correct value above.

relative_time_qa (anchor resolvable):
Task note (relative-time QA): the question
situates the target relative to an anchor.
Anchor: {anchor}. The model may express the
answer date absolutely or relative to the
anchor; a relative expression counts as
committing to a date only if it resolves to
the listed ground-truth date at its required
granularity.

relative_time_qa (no anchor available):
Task note (relative-time QA): no anchor date
is available for this item. Decide each date
match by semantic equivalence between the
model answer and the listed ground-truth date
(booleans only).

open_ended, scenario_short_answer, and
temporal_scenario_short_answer receive no
task-specific note.

===== CASE (per item, after the header/note)
=====

Now analyze the following case.

Context/scenario that was shown to the model
(for reference only - the ground truth below
remains the authority):
{context}

Question: {question}

Ground-truth answer: {ground_truth}

Ground-truth dates (canonical, extracted
programmatically from the ground-truth
answer):
  [0] {canonical_date} - as written in the
  ground truth: "{source_span}" - required
  match: {granularity_description}
  ...
(or, if none: "Ground-truth dates: none were
extracted from the ground-truth answer. Return
an empty date_matches list.")

Model answer: {model_answer}

(The Context/scenario block is omitted for QA
types with no context/scenario. For
perturbation_explanation the label instead
reads: "Statement under evaluation (the model
judged it wrong and explained why; analyze
that explanation against the ground truth):"
followed by {statement}.)
\end{verbatim}
\end{tcolorbox}

\subsection{Qualitative Analysis} \label{app:failure_mode_examples}

\subsubsection{Error Analysis of Perturbation Identification Failures}
\label{app:perturbation}

Each box below shows one perturbed provision, the reference answer, and the
model's own explanation. We highlight in \colorbox{red!30}{red} the fact that we
silently altered together with the incorrect or missing corrections in the
model's answer, and in \colorbox{green!30}{green} the correct value from the
source provision as well as any part of the model's answer that is right.
Examples 1 to 3 show \texttt{Claude} in the parametric setting
flagging the provision and then blaming an element that was never altered.
Example 4 shows \texttt{Gemini} under RAG naming the altered element
correctly and then supplying a replacement that the provision does not support.
Example 5 shows \texttt{GPT-5} under in-context grounding naming the
element correctly and supplying no replacement at all. Every case is detected
and every case scores zero on identification, which shows that the two abilities
fail independently.
\begin{tcolorbox}[
  colback=blue!5!white,
  colframe=blue!30!black,
  boxrule=0.5pt,
  arc=3pt,
  left=5pt, right=5pt, top=5pt, bottom=5pt,
  enhanced,
  breakable,
  title={Example 1: Detected but Misidentified (\texttt{Claude})},
]
\footnotesize

\textbf{Statement}: As of the 2011-06-09 Baggage
Rules, customs officers are permitted to perform random scanning and examination
of up to \colorbox{red!30}{10\%} of passengers passing through the Green Channel.

\smallskip
\textbf{Ground Truth Answer}: The statement is inaccurate. According to the
2011-06-09 Baggage Rules, customs officers are authorized to scan and examine up
to \colorbox{green!30}{5\%} (five percent) of passengers passing through the
Green Channel on a random basis.

\smallskip
\textbf{Model Explanation}: The statement ``\ldots up to 10\% of passengers
passing through the Green Channel'' is incorrect.
\colorbox{red!30}{The error lies in the date} \colorbox{red!30}{mentioned in the
statement.} \colorbox{red!30}{The Baggage Rules in Bangladesh are not dated}
\colorbox{red!30}{2011-06-09.} The Baggage Rules in Bangladesh are updated
periodically, and \colorbox{red!30}{the latest version is not from 2011.} The
correct and current Baggage Rules in Bangladesh \colorbox{red!30}{do not specify
a} \colorbox{red!30}{fixed percentage} of passengers to be randomly scanned and
examined. The percentage \colorbox{red!30}{is at the discretion of the customs
authorities} and may vary based on risk assessment and intelligence inputs.

\smallskip
\textbf{Model Reasoning}: The 2011-06-09 Baggage Rules do allow random scanning,
but \colorbox{red!30}{the actual percentage is not known to me.}


\smallskip
\textbf{Justification}: The model incorrectly identifies the date as the error instead
of the percentage, and it provides false information regarding the regulation's
content.
\end{tcolorbox}

\begin{tcolorbox}[
  colback=blue!5!white,
  colframe=blue!30!black,
  boxrule=0.5pt,
  arc=3pt,
  left=5pt, right=5pt, top=5pt, bottom=5pt,
  enhanced,
  breakable,
  title={Example 2: Detected but Misidentified (\texttt{Claude})},
]
\footnotesize

\textbf{Statement}: According to the June 6, 2024
Customs Order, the total value of a comprehensive guarantee must be
\colorbox{red!30}{less than} the cumulative duties and taxes payable on the
goods.

\smallskip
\textbf{Ground Truth Answer}: The statement is inaccurate. According to the
June 6, 2024 Customs Order, the total value of the comprehensive guarantee must
\colorbox{green!30}{not be less than} the cumulative duties and taxes payable on
the goods.

\smallskip
\textbf{Model Explanation}: The statement \ldots{} is incorrect.
The error lies in
\colorbox{red!30}{the date} mentioned in the statement.
\colorbox{red!30}{There is no ``June 6, 2024}
\colorbox{red!30}{Customs Order''} that I am aware of in the context of
Bangladesh customs regulations. The correct fact should be:
according to the Bangladesh Customs Act and the relevant SROs issued by the
National Board of Revenue, the total value of a comprehensive guarantee must be
\colorbox{green!30}{equal to or greater than} the cumulative duties and taxes
payable on the goods.

\smallskip
\textbf{Model Reasoning}: \colorbox{red!30}{I do not have information}
\colorbox{red!30}{on the June 6, 2024 Customs Order} \ldots{} Without access to
the actual regulations, \colorbox{red!30}{I cannot confidently verify} the
accuracy of this statement.


\smallskip
\textbf{Justification}: The model incorrectly identifies the date as the error instead
of the phrase ``less than'', and it fails to provide the correct regulation
requirement as specified in the ground truth.
\end{tcolorbox}

\begin{tcolorbox}[
  colback=blue!5!white,
  colframe=blue!30!black,
  boxrule=0.5pt,
  arc=3pt,
  left=5pt, right=5pt, top=5pt, bottom=5pt,
  enhanced,
  breakable,
  title={Example 3: Altered Fact Endorsed as Correct (\texttt{Claude})},
]
\footnotesize

\textbf{Statement}: Under the 2025-05-18
Transit/Transshipment Rules, customs authorities must ensure the release of
bonds or guarantees within \colorbox{red!30}{10 working days} after the
completion of the transit process.

\smallskip
\textbf{Ground Truth Answer}: The statement is inaccurate. According to the
Transit/Transshipment Rules 2025 (2025-05-18), customs authorities must ensure
the release of bonds or guarantees within \colorbox{green!30}{7 working days},
not 10.

\smallskip
\textbf{Model Explanation}: The statement \ldots{} contains an error.
\colorbox{red!30}{The error lies in the date} mentioned in the statement.
\colorbox{red!30}{The correct date should be 2022-05-18,}
\colorbox{red!30}{not 2025-05-18.} The reason is that Bangladesh's current
Transit/Transshipment Rules \colorbox{red!30}{were issued on May 18, 2022,} not
in the future year of 2025. The customs authorities in Bangladesh are required
to ensure the release of bonds or guarantees
\colorbox{red!30}{within 10 working days} after the completion of the transit
process, as per the 2022-05-18 Rules.

\smallskip
\textbf{Model Reasoning}: \colorbox{red!30}{I do not have any knowledge}
of these Rules \ldots{} \colorbox{red!30}{I cannot verify} the accuracy of this
statement.


\smallskip
\textbf{Justification}: The model incorrectly identifies the date as the error and fails
to identify the actual error regarding the 10-day period, providing an incorrect
correction.
\end{tcolorbox}

\begin{tcolorbox}[
  colback=blue!5!white,
  colframe=blue!30!black,
  boxrule=0.5pt,
  arc=3pt,
  left=5pt, right=5pt, top=5pt, bottom=5pt,
  enhanced,
  breakable,
  title={Example 4: Correct Element, Incorrect Replacement (\texttt{Gemini})},
]
\footnotesize

\textbf{Statement}: As of the 2011-06-09 Baggage
Rules, customs officers are permitted to perform random scanning and examination
of up to \colorbox{red!30}{10\%} of passengers passing through the Green Channel.

\smallskip
\textbf{Ground Truth Answer}: The statement is inaccurate. According to the
2011-06-09 Baggage Rules, customs officers are authorized to scan and examine up
to \colorbox{green!30}{5\%} (five percent) of passengers passing through the
Green Channel on a random basis.

\smallskip
\textbf{Model Explanation}: The statement is incorrect regarding the percentage
of passengers subject to random examination. The error lies in
\colorbox{green!30}{the specific figure ``10\%''}. Under the
\colorbox{red!30}{Baggage Rules, 2016} (which
\colorbox{red!30}{superseded the 2011 rules}) \ldots{} the threshold for random
physical examination is not 10\%. \colorbox{red!30}{There is no fixed ``10\%
quota''} or limitation that restricts officers to only 10\% of passengers
\ldots{} customs officers have the legal right to stop and examine
\colorbox{red!30}{100\% of passengers} passing through the Green Channel
\ldots{} the authority is discretionary and
\colorbox{red!30}{based on risk assessment} rather than a static percentage cap.

\smallskip
\textbf{Model Reasoning}: The Baggage Rules of 2011
\colorbox{red!30}{do not contain a provision} limiting customs officers to a
10\% examination rate \ldots{} \colorbox{red!30}{there is no such fixed}
\colorbox{red!30}{percentage restriction} in the regulatory framework.


\smallskip
\textbf{Justification}: The model fails to provide the correct 5\% threshold specified in
the ground truth and instead provides an incorrect interpretation of the
regulatory framework regarding examination limits.
\end{tcolorbox}

\begin{tcolorbox}[
  colback=blue!5!white,
  colframe=blue!30!black,
  boxrule=0.5pt,
  arc=3pt,
  left=5pt, right=5pt, top=5pt, bottom=5pt,
  enhanced,
  breakable,
  title={Example 5: Correct Element, No Replacement Offered (\texttt{GPT-5})},
]
\footnotesize

\textbf{Statement}: According to the Customs Act
1969, a customs officer may enter private residential premises to conduct an
audit at any time \colorbox{red!30}{without a warrant or the}
\colorbox{red!30}{consent of the occupant}.

\smallskip
\textbf{Ground Truth Answer}: The statement is inaccurate. The Customs Act 1969
prohibits customs officers from entering private residential premises
\colorbox{green!30}{without the consent of the occupant}
\colorbox{green!30}{or owner, or without a valid warrant} issued under the Act.

\smallskip
\textbf{Model Explanation}: The error is the claim that a customs officer may
enter private residential premises \colorbox{green!30}{at any time without a
warrant} \colorbox{green!30}{or the consent of the occupant}. That is
\colorbox{red!30}{not found in the provided} \colorbox{red!30}{official
document}. The document supplied is an S.R.O.\ amending the Customs Agent
Licensing Rules, 2016 \ldots{} Nothing in this S.R.O.\ grants a power for
customs officers to enter private residential premises. Any actual powers of
entry \colorbox{red!30}{would need to be found in the}
\colorbox{red!30}{relevant provisions of the Customs Act itself}.


\smallskip
\textbf{Justification}: The model correctly identifies the statement as wrong but fails
to provide the required correction regarding the necessity of a warrant or
consent for entry into private premises from the provided documents.
\end{tcolorbox}

\subsubsection{Examples of Cross-Setting Behavior}
\label{app:decomp_examples}

The boxes below show one question each answered across settings: an item unrecoverable even with the gold document (Example~1); a
correct memory answer the gold document overturns (Example~2); an item answerable from the gold
document but missed once retrieval must find it (Example~3); and a planted false premise the model
accepts closed-book but rejects once the authoritative document is retrieved (Example~4).

\begin{tcolorbox}[
  colback=blue!5!white,
  colframe=blue!30!black,
  boxrule=0.5pt,
  arc=3pt,
  left=5pt, right=5pt, top=5pt, bottom=5pt,
  enhanced,
  breakable,
  title={Example 1: Unrecoverable Even With the Gold Document (\texttt{DeepSeek})},
]
\footnotesize
\textbf{Question}: What is the primary legal basis for rule-making as of May 29, 2024?

\smallskip
\textbf{Ground Truth Answer}: The \colorbox{green!30}{Customs Act, 2023} (Act No.\ 57 of 2023).

\smallskip
\textbf{Parametric (closed-book)}: ``\ldots\ the primary legal basis \ldots\ is the
\colorbox{red!30}{Customs Act, 1969} (Act No.\ IV of 1969), which grants \ldots\ the authority to
issue Statutory Regulatory Orders.''

\smallskip
\textbf{ICL (gold document)}: ``The primary legal basis \ldots\ is the
\colorbox{red!30}{Customs Act, 1969}, specifically the powers conferred by Section 219B \ldots\ under
which the Post Clearance Audit Manual, 2023 was formulated.''

\smallskip
\textit{Reading}: A 2023-era provision is in the prompt, yet the model still resolves rule-making to
the superseded 1969 Act. The document is already present, so retrieval cannot help; only stronger
version resolution can.
\end{tcolorbox}

\begin{tcolorbox}[
  colback=blue!5!white,
  colframe=blue!30!black,
  boxrule=0.5pt,
  arc=3pt,
  left=5pt, right=5pt, top=5pt, bottom=5pt,
  enhanced,
  breakable,
  title={Example 2: A Correct Memory Answer Overturned by the Gold Document (\texttt{Gemini})},
]
\footnotesize
\textbf{Question}: When did the Advance Rulings Rules 2016 come into force?

\smallskip
\textbf{Ground Truth Answer}: Immediately upon issuance on \colorbox{green!30}{June 5, 2016}.

\smallskip
\textbf{Parametric (closed-book)}: ``\ldots\ came into force on \colorbox{green!30}{June 5, 2016}, the
same date they were officially notified by the National Board of Revenue.''

\smallskip
\textbf{ICL (gold document)}: ``\ldots\ came into force immediately upon the issuance of the
notification, which was dated \colorbox{red!30}{June 2, 2016} (19 Jaishtha, 1423 BS).''

\smallskip
\textit{Reading}: Closed-book the model recalls the correct date; handed the document, it misreads
the notification's Bangla-calendar date and overwrites a right answer with a wrong one.
\end{tcolorbox}

\begin{tcolorbox}[
  colback=blue!5!white,
  colframe=blue!30!black,
  boxrule=0.5pt,
  arc=3pt,
  left=5pt, right=5pt, top=5pt, bottom=5pt,
  enhanced,
  breakable,
  title={Example 3: Answerable From the Gold Document, Missed When Retrieval Must Find It (\texttt{Kimi})},
]
\footnotesize
\textbf{Question}: What new HS Code replaced 3204.13.00 in the 2015 update?

\smallskip
\textbf{Ground Truth Answer}: 3204.13.00 was replaced by \colorbox{green!30}{3204.13.90}.

\smallskip
\textbf{ICL (gold document)}: ``According to SRO No.\ 156-Law/2015/36-Customs dated 4 June 2015,
3204.13.00 was replaced by \colorbox{green!30}{3204.13.90}.''

\smallskip
\textbf{RAG (retrieved)}: ``\ldots\ reviewing SRO 140-AIN/2015/20 \ldots\ I do not see any entry for
3204.13.00 \ldots\ \colorbox{red!30}{there is no explicit information about what new HS} \colorbox{red!30}{Code replaced
3204.13.00 in the 2015 update}.''

\smallskip
\textit{Reading}: The corpus contains the answer and ICL proves it, but the retriever returns the
neighboring orders (SRO 140-AIN/2015/20, SRO 151-AIN/2016/13) instead of the amending SRO
156-Law/2015/36, so the model declares the change absent.
\end{tcolorbox}

\begin{tcolorbox}[
  colback=blue!5!white,
  colframe=blue!30!black,
  boxrule=0.5pt,
  arc=3pt,
  left=5pt, right=5pt, top=5pt, bottom=5pt,
  enhanced,
  breakable,
  title={Example 4: Following a False Premise vs.\ Rejecting It With the Document (\texttt{GPT-5})},
]
\footnotesize
\textbf{Question}: Was the transition of HS code 3402.13.00 to 3402.13.90 part of the 2015 regulatory
changes?

\smallskip
\textbf{Injected (misleading) context}: ``The amendment \ldots\ was finalized in the
\colorbox{red!30}{2015 notification SRO 158-AIN/2015}.''

\smallskip
\textbf{Ground Truth Answer}: No; the amendment was implemented by
\colorbox{green!30}{SRO 222-AIN/2016 on July 1, 2016}, not by SRO 158-AIN/2015.

\smallskip
\textbf{Parametric (closed-book)}: ``\colorbox{red!30}{Yes.} The change \ldots\ was enacted as part
of the 2015 regulatory amendments under \colorbox{red!30}{SRO 158-AIN/2015}.''

\smallskip
\textbf{RAG (retrieved)}: ``No. The change \ldots\ was made by
\colorbox{green!30}{S.R.O.\ No.\ 222-AIN/2016 (01 July 2016)}, not by SRO 158-AIN/2015.''

\smallskip
\textit{Reading}: With no source the model adopts the confident but false premise; once the
authoritative order is retrieved and stacked above the misleading context, it rejects the planted
claim.
\end{tcolorbox}

\subsubsection{Examples of Relative-Time and Event-Sorting Failures}
\label{app:task_examples}

The boxes below show worked examples for two further task types. Examples 1 and 2 are Relative-Time QA, which asks the model to resolve a relative offset over
an evolving provision. On a question about a rare \emph{downward} revision, two models both fail the
temporal reasoning: they miscompute the nine-year interval and assume the allowance rose, missing the
actual cut. Examples 3 and 4 are Event Sorting: closed-book the model orders amendments by magnitude
rather than by date, a failure shared by all nine models (Example 3), and under retrieval it still
misorders a longer timeline whose effective dates are spread across separate instruments (Example 4).
Examples 1--3 share one bias: the models assume regulatory values only increase over time.

\begin{tcolorbox}[
  colback=blue!5!white,
  colframe=blue!30!black,
  boxrule=0.5pt,
  arc=3pt,
  left=5pt, right=5pt, top=5pt, bottom=5pt,
  enhanced,
  breakable,
  title={Example 1: Miscomputed Interval and Wrong Direction (\texttt{Gemini})},
]
\footnotesize
\textbf{Question}: Starting from the 2016-06-02 policy which set the airline crew baggage allowance at
300 USD, what was the adjustment made nine years later?

\smallskip
\textbf{Ground Truth Answer}: In \colorbox{green!30}{2025-06-03}, the allowance for airline crew was
adjusted \colorbox{green!30}{downward to 150 USD}.

\smallskip
\textbf{Model Answer}: ``The 2016 policy was actually adjusted in
\colorbox{red!30}{2024 (eight years later, rather than nine)} \ldots\ the duty-free baggage allowance
for airline crew members was \colorbox{red!30}{increased from 300 USD to 400 USD}.''

\smallskip
\textit{Reading}: The model miscounts the nine-year interval, placing the change in 2024, and assumes
the allowance rose, when the actual amendment cut it to 150 USD. The temporal reasoning fails on both
the interval and the direction of change.
\end{tcolorbox}

\begin{tcolorbox}[
  colback=blue!5!white,
  colframe=blue!30!black,
  boxrule=0.5pt,
  arc=3pt,
  left=5pt, right=5pt, top=5pt, bottom=5pt,
  enhanced,
  breakable,
  title={Example 2: A Downward Change Denied as an Increase (\texttt{Qwen})},
]
\footnotesize
\textbf{Question}: Starting from the 2016-06-02 policy which set the airline crew baggage allowance at
300 USD, what was the adjustment made nine years later? (the same question as Example 1)

\smallskip
\textbf{Ground Truth Answer}: In \colorbox{green!30}{2025-06-03}, the allowance was cut
\colorbox{green!30}{to 150 USD}.

\smallskip
\textbf{Model Answer}: ``\ldots\ an adjustment occurred in \colorbox{red!30}{2023, seven years after}
the 2016 policy, increasing the limit \colorbox{red!30}{from \$300 to \$500} \ldots\ there is
\colorbox{red!30}{no official record of a specific adjustment made nine} \colorbox{red!30}{years later in 2025}.''

\smallskip
\textit{Reading}: On the same question, a second model also assumes the allowance rose (to \$500) and
then denies that any change occurred at the nine-year mark, missing the downward revision entirely.
Both models impose a monotonic-increase prior rather than tracking the actual evolution.
\end{tcolorbox}

\begin{tcolorbox}[
  colback=blue!5!white,
  colframe=blue!30!black,
  boxrule=0.5pt,
  arc=3pt,
  left=5pt, right=5pt, top=5pt, bottom=5pt,
  enhanced,
  breakable,
  title={Example 3: Ordered by Magnitude, Not by Date (\texttt{Gemini})},
]
\footnotesize
\textbf{Question}: Arrange these events in chronological order (earliest first).

\smallskip
\textbf{Events}: (0) The baggage allowance for airline crew members is set at 100 USD. \quad
(1) \ldots\ is adjusted to 150 USD. \quad (2) \ldots\ is increased to 300 USD.

\smallskip
\textbf{Correct order}: \colorbox{green!30}{100 USD $\rightarrow$ 300 USD $\rightarrow$ 150 USD}
(the allowance was raised to 300, then later cut to 150).

\smallskip
\textbf{Model's order}: \colorbox{red!30}{100 USD $\rightarrow$ 150 USD $\rightarrow$ 300 USD}
(a monotonic increase).

\smallskip
\textit{Reading}: Without the dated amendments the model orders the values by magnitude, assuming the
allowance only ever rose. \textbf{All nine models} make this same error in the closed-book setting;
given the documents, eight of nine recover the true order, confirming that the ordering needs the
dated evidence, not more reasoning.
\end{tcolorbox}

\begin{tcolorbox}[
  colback=blue!5!white,
  colframe=blue!30!black,
  boxrule=0.5pt,
  arc=3pt,
  left=5pt, right=5pt, top=5pt, bottom=5pt,
  enhanced,
  breakable,
  title={Example 4: A Longer Timeline Misordered Under Retrieval (\texttt{DeepSeek})},
]
\footnotesize
\textbf{Question}: Arrange these events in chronological order (earliest first).

\smallskip
\textbf{Events}: (0) established minimum import values for assessing customs duty; \quad
(1) formalized tariff value and duty structures under the Customs Act; \quad
(2) expanded minimum-value tables to broader categories (bovine meat, milk powder); \quad
(3) directly substituted specific HS codes for industrial headings.

\smallskip
\textbf{Correct order}: \colorbox{green!30}{(3) $\rightarrow$ (0) $\rightarrow$ (1) $\rightarrow$ (2)}.

\smallskip
\textbf{Model's order}: \colorbox{red!30}{(0) $\rightarrow$ (3) $\rightarrow$ (2) $\rightarrow$ (1)}.

\smallskip
\textit{Reading}: The four amendments' effective dates sit in separate instruments; retrieval
supplies an incomplete set, so the model produces a plausible but wrong sequence. This is the
retrieval-completeness failure behind the RAG collapse on Event Sorting.
\end{tcolorbox}

\subsection{Additional Result Analysis}\label{app:more_results}

\subsubsection{Cross-Setting Model Behavior}
\label{app:decomposition}

\noindent \textbf{\textit{Parametric knowledge recognizes instruments but cannot resolve versions.}}
Table \ref{tab:main_results} shows that with no context, the best macro average is only $35.39\%$ (\texttt{GPT-5}), and open models such as \texttt{Qwen} and \texttt{GLM} sit near $26\%$. \texttt{Claude} also scores near this level, but for a different reason: it often declines to answer rather than guess. Moreover, the difficulty is uneven across tasks. Recognition-style questions remain tractable, as Temporal MCQ reaches $48.97\%$ to $70.17\%$. In contrast, tasks that require resolving the correct version fall to the floor, with Context Misalignment between $1.72\%$ and $7.87\%$ and strict Relative-Time QA between $3.77\%$ and $12.37\%$. This split shows that a model can recall that an instrument exists while still failing to recover which version applies on a given date.

\noindent \textbf{\textit{Reading skill and memorization are decoupled across models.}}
The ranking of models changes with the setting, which shows that recalling this corpus and reading it are different abilities. \texttt{Claude} illustrates this most clearly. Its parametric macro average is only $25.91\%$, near the bottom, yet it leads all models under RAG at $58.05\%$ and ranks second under ICL at $64.39\%$, and it shows the largest jump from the parametric to the ICL setting, at $38.48$ points. The low parametric score is not only due to missing knowledge. In the parametric setting \texttt{Claude} frequently declined to answer and stated that it did not have the specific information, rather than producing a guess. A refusal is a safer failure mode than a confident wrong answer, and it contrasts with the over-trust of a false premise that we observe under Context Misalignment. \texttt{GPT-5}, by contrast, is strong in every setting, while \texttt{GLM} is consistently weakest. \texttt{Gemma} shows the opposite profile to \texttt{Claude}, with the best parametric Temporal MCQ at $70.17\%$ but a weaker use of retrieved context.


\noindent \textbf{\textit{A third of the benchmark is unrecoverable even with the gold document.}}
By parametric and ICL outcome (Table~\ref{tab:decomposition}), $33.9\%$ of items are answered from
memory, $37.0\%$ are a gap the gold document closes, and $29.2\%$ stay wrong even when it is
supplied: a reasoning ceiling, since the correct provision is already in the prompt. Two smaller
effects qualify this: the gold document overturns a correct memory answer for $5.5\%$ of items, and
$17.3\%$ are answered under ICL but missed under RAG, so retrieval recall alone explains most of
RAG's deficit (restricted to items ICL answers, RAG succeeds on only $73.5\%$).

\begin{table}[!ht]
\centering
\small
\setlength{\tabcolsep}{6pt}
\renewcommand{\arraystretch}{1.15}
\begin{tabular}{@{}lr@{}}
\toprule
\rowcolor{headshade} \textbf{Outcome regime} & \textbf{Items} \\
\midrule
Known without context                   & 33.9\% \\
Recovered by the gold document          & 37.0\% \\
Unrecoverable with the gold document    & 29.2\% \\
\midrule
Gold-document regression                & 5.5\%  \\
Retrieval gap (ICL right, RAG wrong)    & 17.3\% \\
\bottomrule
\end{tabular}
\caption{Per-item decomposition by outcome across settings, pooled over the nine models on the
matched set. The first three rows partition every item ($100\%$); the last two are cross-cutting
effects. The $29\%$ ``unrecoverable with the gold document'' is a reasoning ceiling retrieval cannot
address.}
\label{tab:decomposition}
\end{table}

\noindent \textbf{\textit{Grounding changes how models fail, not only how often.}}
Among wrong answers, flat factual errors fall from $58\%$ of parametric failures to $50\%$ (ICL) and
$53\%$ (RAG), while failures that require a document to exist (asserting a date that conflicts with
the source, or overriding a supplied value with memory) rise from under $2\%$ to $4$--$5\%$.
Grounding thus trades hallucination for \emph{misgrounding}. The event-sorting drop under RAG is a
strict, exact-match effect: under graded pairwise-order accuracy the ICL--RAG gap narrows from
$0.57$ vs.\ $0.28$ to $0.81$ vs.\ $0.67$, as most wrong orderings misplace only a few events.

\noindent \textbf{\textit{Once grounded, the residual temporal error flips from meaning to date.}}
On judged items carrying gold dates (Table~\ref{tab:temporal_split}), the main parametric residual is
a plausible date on the wrong substance ($35\%$), while the main grounded residual is the reverse,
correct substance with the wrong date ($10\%$ under ICL). Grounding settles \emph{what} the provision
says and leaves \emph{when} as the last mile, which the date gate (Section~\ref{sec:evaluation_metrics})
scores as zero.

\begin{table}[!ht]
\centering
\small
\setlength{\tabcolsep}{6pt}
\renewcommand{\arraystretch}{1.25}
\begin{tabular}{@{}lcc@{}}
\toprule
\rowcolor{headshade} \textbf{Setting} & \makecell{\textbf{Meaning right,}\\\textbf{date wrong}} &
\makecell{\textbf{Date right,}\\\textbf{meaning wrong}} \\
\midrule
Parametric & 2\%           & \textbf{35\%} \\
RAG        & 6\%           & 22\%          \\
ICL        & \textbf{10\%} & 20\%          \\
\bottomrule
\end{tabular}
\caption{Meaning versus date on judged items carrying gold dates (matched set). The dominant residual
flips from \emph{date right, meaning wrong} without a document to \emph{meaning right, date wrong}
once one is supplied.}
\label{tab:temporal_split}
\end{table}

\noindent \textbf{\textit{A correct document lets models reject a false premise they otherwise
accept.}}
In context misalignment (Table~\ref{tab:conflict_behavior}), closed-book the model adopts the planted
value $65\%$ of the time and gives the authoritative value only $14\%$; with a correct document
present the authoritative rate rises to $66\%$ (ICL) and $50\%$ (RAG), and conflict acknowledgment
from $7\%$ to about $26\%$. Two cautions: the parametric column measures robustness to an unsupported
premise while the grounded columns measure conflict resolution with the correct text present, so they
are different conditions; and ``gave the authoritative value'' is not memory reliance under grounding,
where that value sits in the document. Appendix~\ref{app:decomp_examples} gives one case per regime.

\begin{table}[!ht]
\centering
\small
\setlength{\tabcolsep}{6pt}
\renewcommand{\arraystretch}{1.25}
\begin{tabular}{@{}lcc@{}}
\toprule
\rowcolor{headshade} \textbf{Setting} & \makecell{\textbf{Adopted the}\\\textbf{misaligned value}} &
\makecell{\textbf{Gave the}\\\textbf{authoritative value}} \\
\midrule
Parametric & \textbf{65\%} & 14\%          \\
RAG        & 15\%          & 50\%          \\
ICL        & 16\%          & \textbf{66\%} \\
\bottomrule
\end{tabular}
\caption{Context-misalignment behavior on the matched set: how often the model adopts the planted
claim versus states the authoritative value. A correct document present shifts the model from
adopting the false premise to rejecting it.}
\label{tab:conflict_behavior}
\end{table}

%

\subsubsection{Task-Level Comparison of Knowledge Injection Strategies}
\label{app:task_breakdown}

\begin{figure*}[t]
    \centering
    \includegraphics[width=1\linewidth]{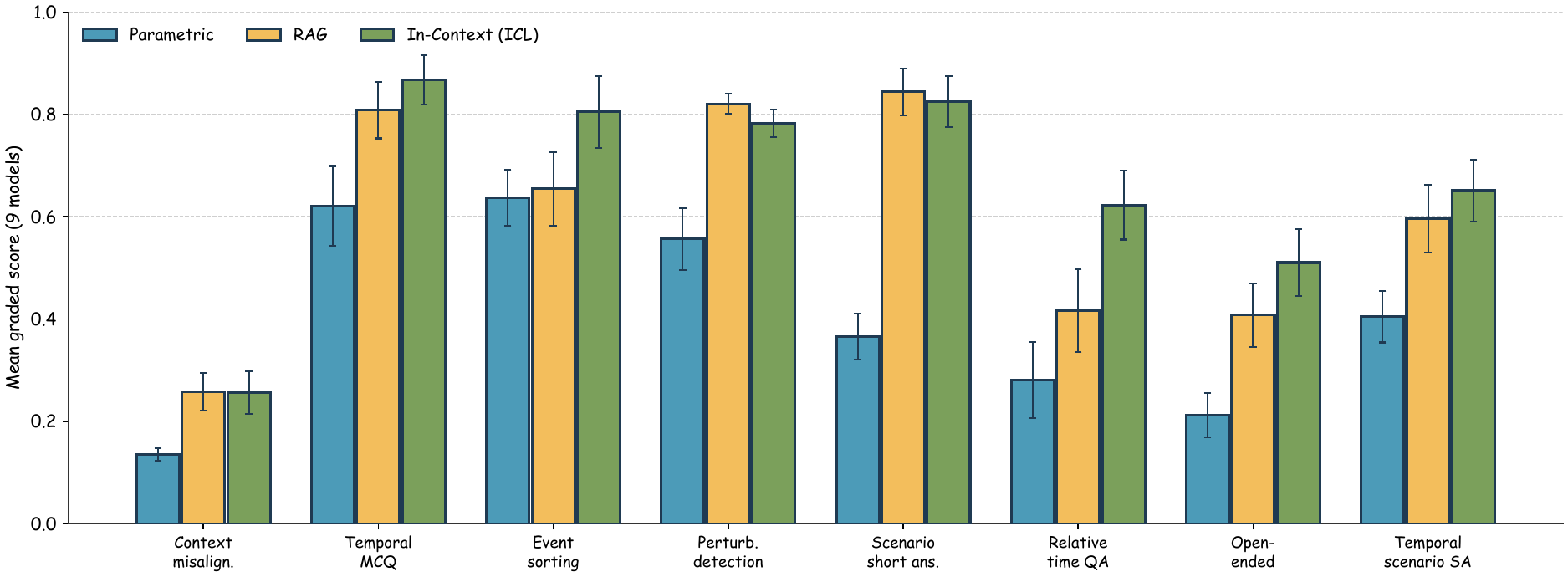}
    \caption{Partial accuracy per task type under the three settings. Each
    bar is the mean over the nine models, and each error bar is one standard
    deviation across models. Differences between the bars are given in
    Table~\ref{tab:task_setting_deltas}.}
    \label{fig:category_bars}
\end{figure*}

Table~\ref{tab:main_results} reports every model on every task under every
setting, and Figure~\ref{fig:category_bars} averages the same scores over the
nine models. This appendix adds the pairwise differences between settings,
which Table~\ref{tab:task_setting_deltas} lists. These differences show that
the macro gap between settings comes from a small group of tasks, whereas the
remaining tasks respond in a similar way to retrieval and to full context.

All values are partial accuracy, averaged over the nine models. The
dispersion quoted below is the standard deviation across models, therefore it
measures disagreement inside the model pool and not item-level sampling
error. It ranges from $0.013$ to $0.081$ over all tasks and settings. We
discuss a difference only when it is larger than this spread for both
settings involved. Three entries fail that test, and we call them
indistinguishable rather than zero. Furthermore, a paired test over the nine
per-model differences would be more sensitive, so these entries are
inconclusive rather than negative.

\begin{table}[!ht]
\centering
\small
\setlength{\tabcolsep}{2pt}
\renewcommand{\arraystretch}{1.15}
\begin{adjustbox}{max width=\columnwidth}
\begin{tabular}{@{}lccc>{\columncolor{cutshade}}c@{}}
\toprule
\rowcolor{headshade}
\textbf{Task type} & \textbf{$n$} & \textbf{RAG$-$P} & \textbf{ICL$-$P} & \textbf{ICL$-$RAG} \\
\midrule
\rowcolor{groupshade}
\multicolumn{5}{@{}l}{\textsc{Full context beyond retrieval}} \\
Relative-Time QA       & 463 & $+0.136$ & $+0.342$ & $+0.206$ \\
Event Sorting$^{\ast}$ & 207 & $+0.018$ & $+0.168$ & $+0.150$ \\
Open-Ended             & 487 & $+0.196$ & $+0.299$ & $+0.103$ \\
Temporal MCQ           & 354 & $+0.187$ & $+0.247$ & $+0.060$ \\
Temporal Scenario SA   & 363 & $+0.192$ & $+0.247$ & $+0.055$ \\
\addlinespace[2pt]
\rowcolor{groupshade}
\multicolumn{5}{@{}l}{\textsc{Retrieval already sufficient}} \\
Scenario SA            & 562 & $+0.479$ & $+0.460$ & $-0.019$ \\
Perturbation Det.      & 498 & $+0.264$ & $+0.226$ & $-0.038$ \\
\addlinespace[2pt]
\rowcolor{groupshade}
\multicolumn{5}{@{}l}{\textsc{Neither setting suffices}} \\
Context Misalign.      & 116 & $+0.123$ & $+0.121$ & $-0.002$ \\
\midrule
Macro average          & 3{,}050 & $+0.199$ & $+0.264$ & $+0.064$ \\
\bottomrule
\end{tabular}
\end{adjustbox}
\caption{Differences in partial accuracy between settings, averaged over the
nine models. RAG$-$P is the RAG score minus the parametric score (P), and a
positive value means the second setting is better. The first and third
columns sum to the second. Rows are sorted by the shaded column, whose sign
defines the three groups. Absolute scores appear in
Figure~\ref{fig:category_bars}. $^{\ast}$Event sorting uses a graded ordering
metric under ICL but exact match elsewhere, so this row is not on a common
scale.}
\label{tab:task_setting_deltas}
\end{table}

\paragraph{Full Context Beyond Retrieval.}
\label{app:task_breakdown_rag_short}

Five tasks improve clearly when the governing clauses are supplied in full
rather than retrieved. Relative-time QA improves the most, by $0.206$, which
is more than twice the across-model spread of either setting. Such a question
asks for the rule in force at an offset from an anchor date. The model must
therefore see both the anchored version and the version that replaced it, and
a top-10 clause retriever often returns one of the two but not the other.

Event sorting shows the second largest gain, at $0.150$. Retrieval alone adds
only $0.018$ here, which is smaller than the across-model spread, so it does
not help at all. However, the ICL score for this task comes from a graded
ordering metric, while the parametric and RAG scores come from exact match.
The three values are therefore not on a common scale, and part of the gain is
a metric effect rather than a capability gain.

\paragraph{Retrieval Is Already Sufficient.}
\label{app:task_breakdown_rag_enough}

In contrast, two tasks gain nothing from full context. On perturbation
detection RAG exceeds ICL by $0.038$, which is larger than the across-model
spread of both settings. On scenario short answer the two settings differ by
only $0.019$, so they are indistinguishable. Both tasks are answered from a
single local span of evidence. A plausible reason is that retrieval acts as a
filter, because ten candidate clauses remove most of the distracting text,
whereas the full document forces the model to find the same clause inside a
much longer input.

\begin{figure*}[t]
    \centering
    \includegraphics[width=1\linewidth]{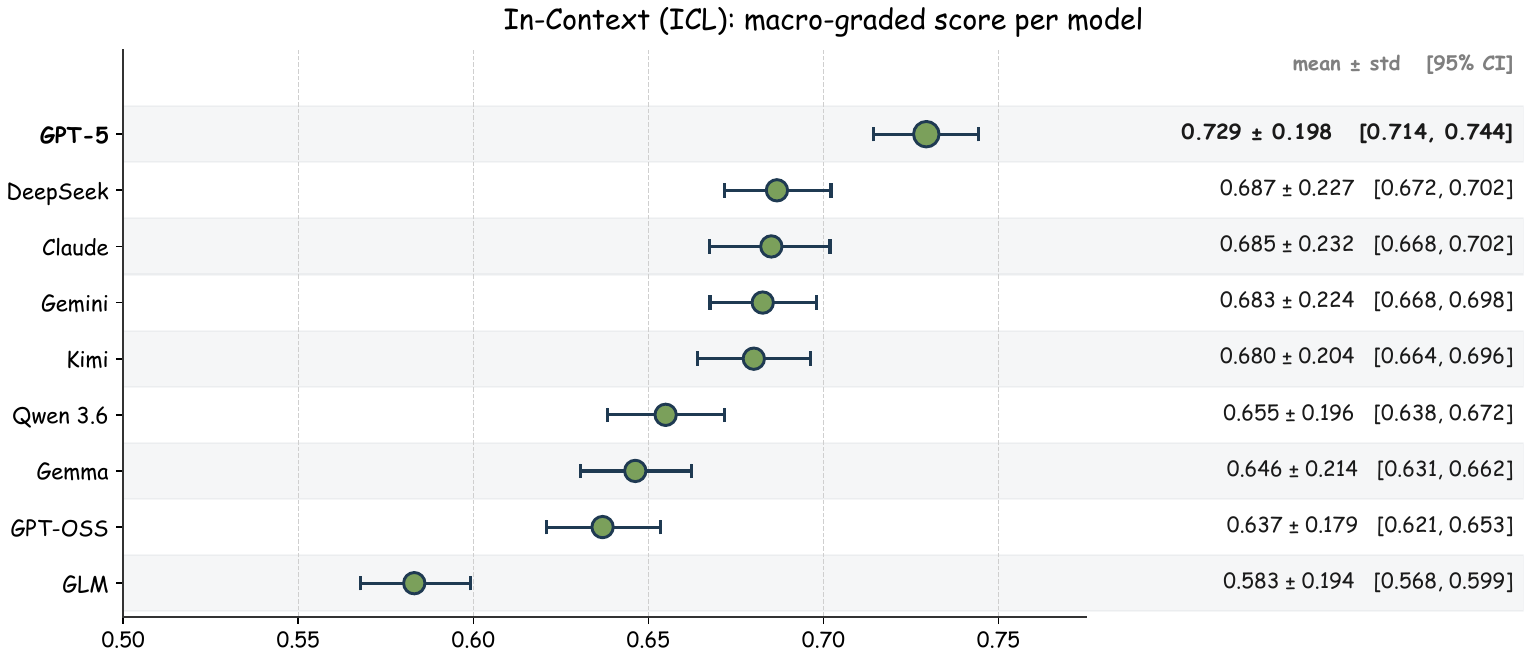}
    \caption{Macro partial accuracy per model under ICL, computed as the
    unweighted mean over the eight task types. The value after $\pm$ is the
    standard deviation across the eight task scores, and the bracketed
    interval is a 95\% confidence interval over the 3{,}050 graded items.
    These two quantities are computed over different units.}
    \label{fig:icl_macro}
\end{figure*}

\paragraph{Context Misalignment Is Not an Evidence Problem.}
\label{app:task_breakdown_misalignment}

Context misalignment is the only task where the two evidence settings behave
identically. RAG gains $0.123$ over parametric knowledge and ICL gains
$0.121$, so they differ by $0.002$, which is far below their across-model
spread of about $0.04$. Moreover, both settings still stop near $0.26$.
Supplying the governing document therefore does not help, because the task
does not ask the model to use the context. It asks the model to notice that
the context cannot answer the question.

\begin{table}[!ht]
\centering
\small
\setlength{\tabcolsep}{15pt}
\renewcommand{\arraystretch}{1.15}
\begin{adjustbox}{max width=\columnwidth}
\begin{tabular}{@{}lc>{\columncolor{cutshade}[\tabcolsep][3.5pt]}c@{}}
\toprule
\rowcolor{headshade}
\multicolumn{1}{@{}>{\columncolor{headshade}[3.5pt][\tabcolsep]}l}{\textbf{Model}} &
\textbf{Strict} &
\multicolumn{1}{>{\columncolor{headshade}[\tabcolsep][3.5pt]}c@{}}{\textbf{Gap}} \\
\midrule
\multicolumn{3}{@{}>{\columncolor{groupshade}[2pt][2pt]}l@{}}{\textsc{Proprietary}} \\
\texttt{GPT-5}           & 0.684 & 0.045 \\
\texttt{Claude}     & 0.643 & 0.042 \\
\texttt{Gemini} & 0.624 & 0.059 \\
\addlinespace[2pt]
\multicolumn{3}{@{}>{\columncolor{groupshade}[2pt][2pt]}l@{}}{\textsc{Open-weight}} \\
\texttt{Kimi}             & 0.632 & 0.048 \\
\texttt{DeepSeek}     & 0.631 & 0.056 \\
\texttt{GPT-OSS}               & 0.571 & 0.066 \\
\texttt{Qwen}         & 0.611 & 0.044 \\
\texttt{Gemma}               & 0.600 & 0.046 \\
\texttt{GLM}               & 0.503 & 0.080 \\
\midrule
Mean                  & 0.611 & 0.054 \\
\bottomrule
\end{tabular}
\end{adjustbox}
\caption{Macro strict accuracy per model under ICL, and its gap from macro
partial accuracy. The gap is the share of answers that are correct in meaning
but wrong on at least one date. Partial accuracy itself is shown in
Figure~\ref{fig:icl_macro}, and the row order follows
Table~\ref{tab:main_results}.}
\label{tab:icl_strict}
\end{table}

\paragraph{Task Difficulty Dominates Model Identity.}
\label{app:task_breakdown_spread}

ICL macro scores range from $0.583$ for \texttt{GLM} to $0.729$ for \texttt{GPT-5}, which is a spread of $0.146$ across nine models. In contrast, the ICL bars in
Figure~\ref{fig:category_bars} range from $0.256$ to $0.868$, which is a
spread of $0.612$. Task difficulty therefore varies about four times more
than model identity. Moreover, the macro gain from parametric knowledge to
ICL is $0.264$, which alone exceeds the whole range between the best and the
worst model.

Table~\ref{tab:icl_strict} reports macro strict accuracy per model. The two
weakest models show the largest gaps from partial accuracy, at $0.080$ and
$0.066$, whereas \texttt{GPT-5} and \texttt{Claude} show $0.045$ and $0.042$.
However, the relation is not monotone in the middle of the ranking, so we
report it as a tendency rather than as a rule.

Finally, half of the eight tasks stay below $0.66$ even under ICL. The
benchmark is therefore not saturated by a longer context window, and the
remaining difficulty lies in version resolution rather than in retrieval.

%

\subsubsection{Why Perturbation Detection Behaves Differently?}
\label{app:perturbation_confusion}

\begin{figure}[t]
    \centering
    \includegraphics[width=\columnwidth]{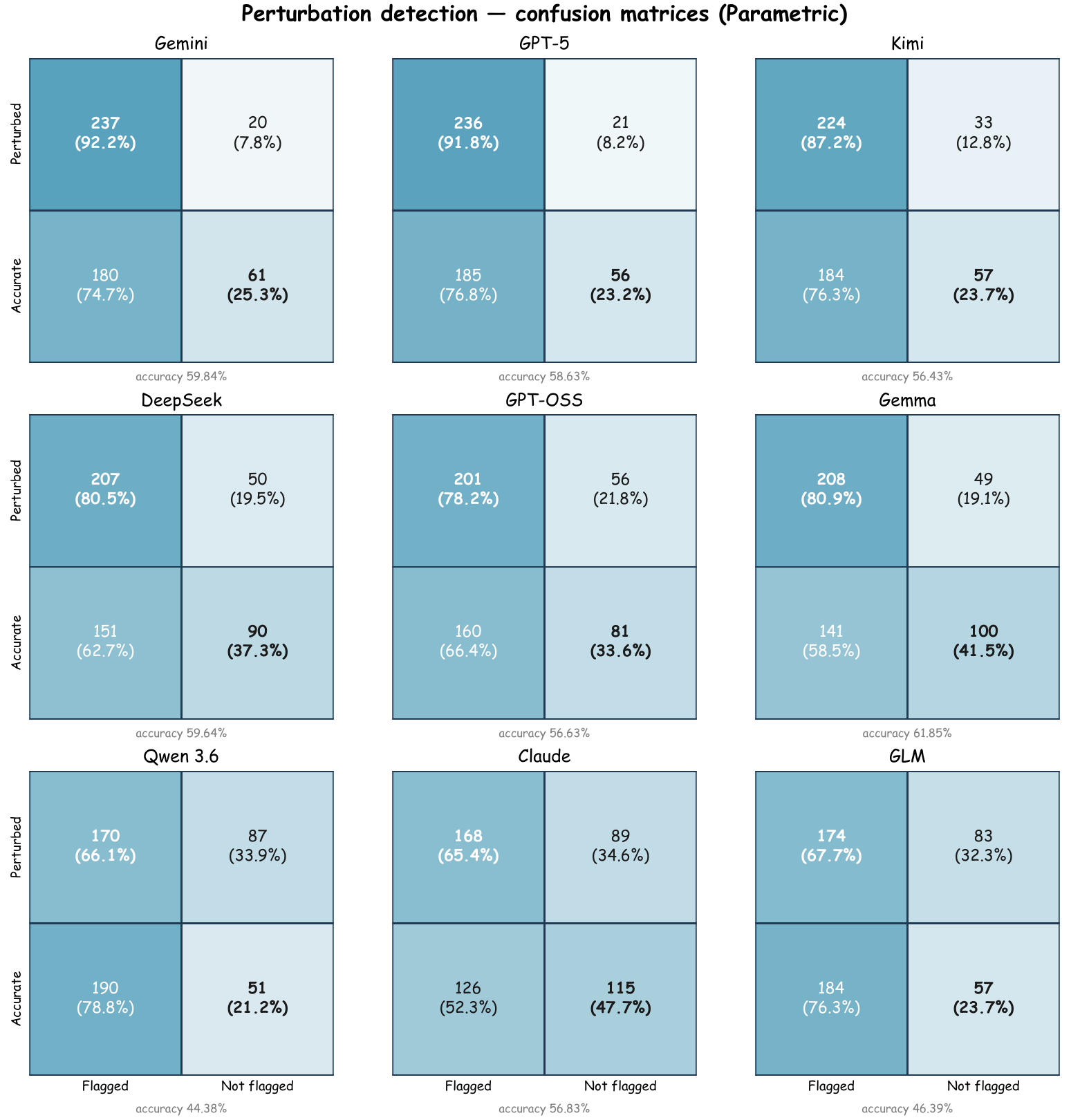}
    \caption{Perturbation detection under parametric knowledge. }
    \label{fig:confusion_parametric}
\end{figure}

\begin{figure}[t]
    \centering
    \includegraphics[width=\columnwidth]{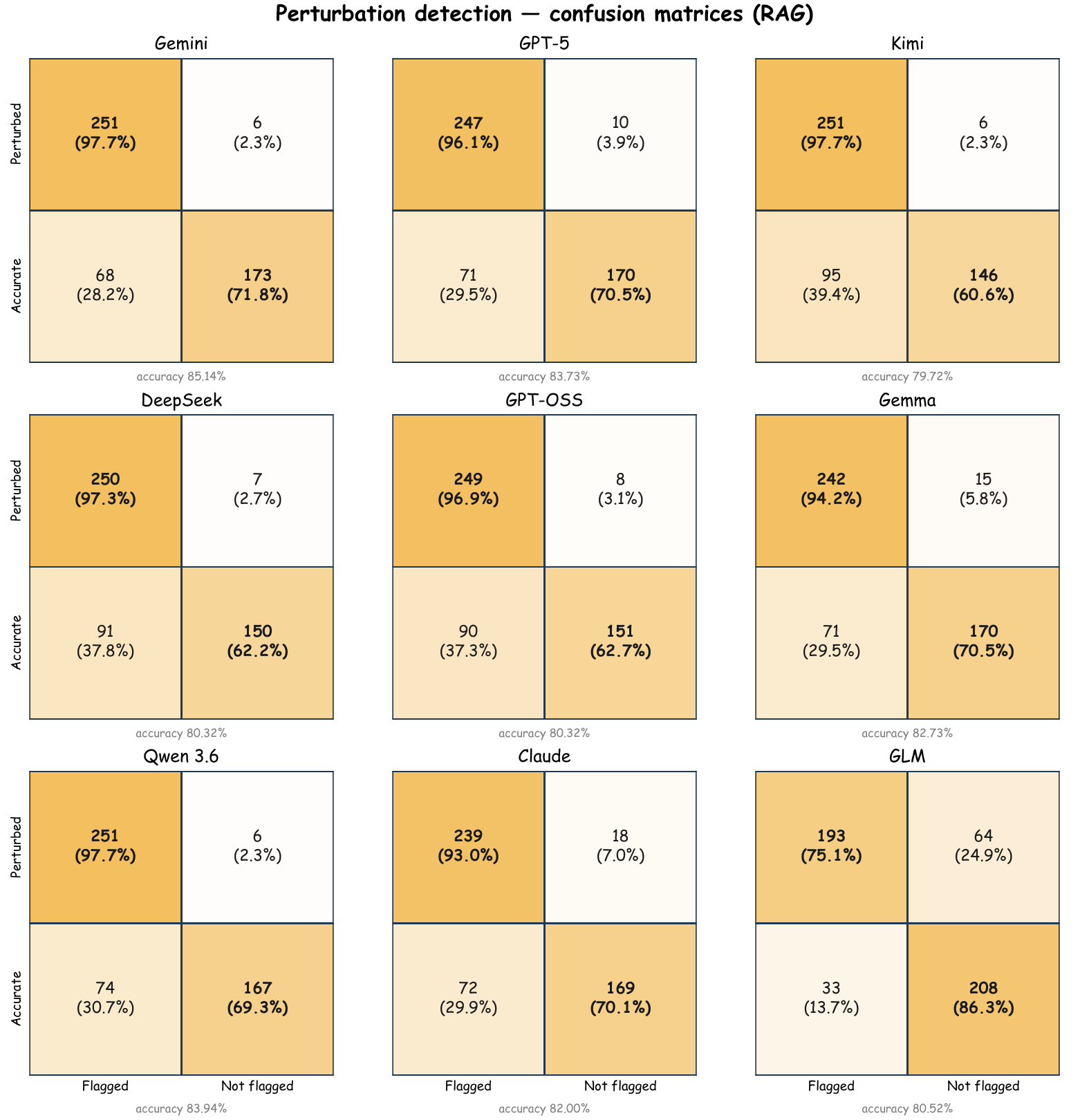}
    \caption{Perturbation detection under RAG.}
    \label{fig:confusion_rag}
\end{figure}

\begin{figure}[!ht]
    \centering
    \includegraphics[width=\columnwidth]{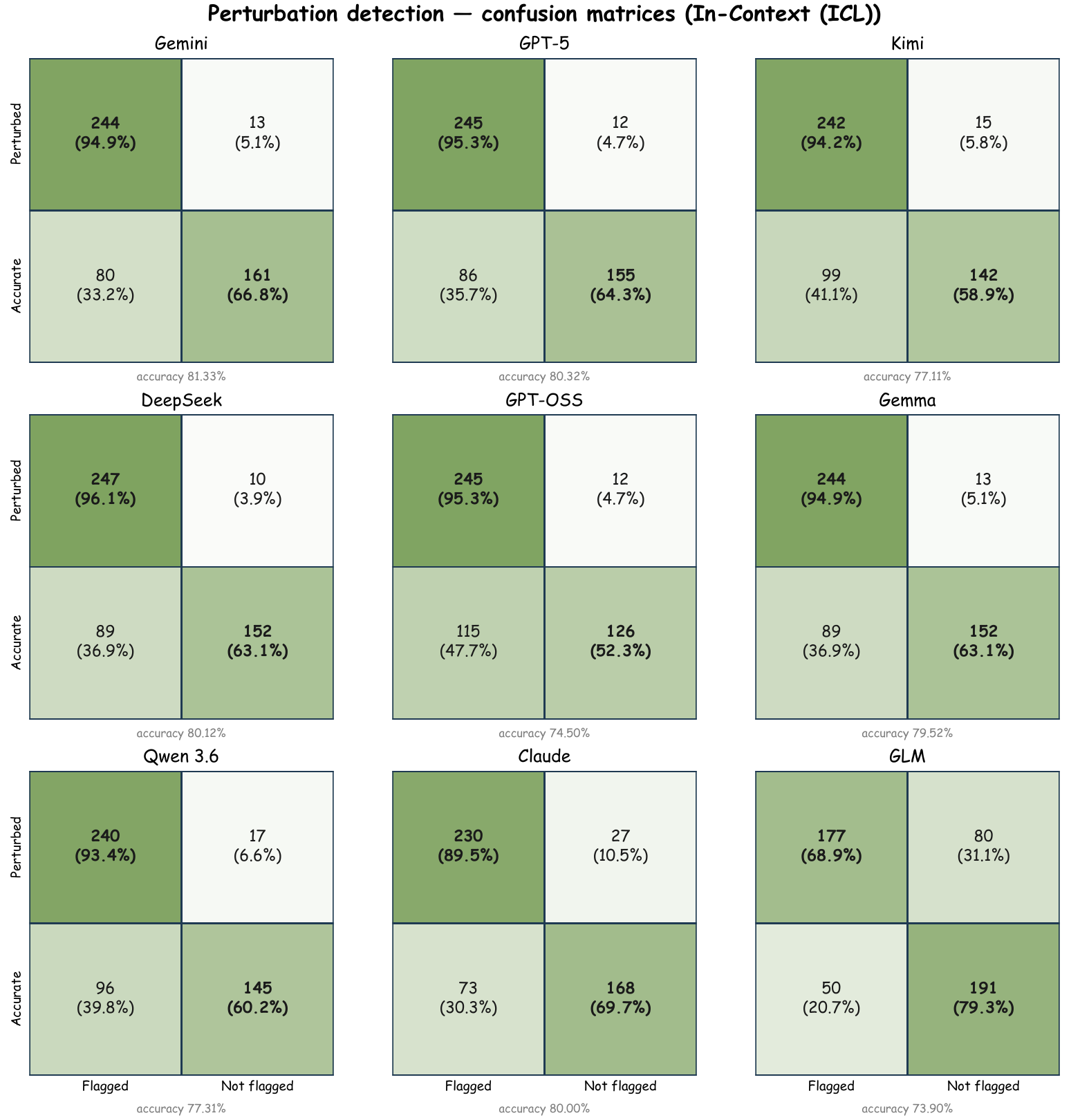}
    \caption{Perturbation detection under ICL.}
    \label{fig:confusion_icl}
\end{figure}

Perturbation detection is the only task where RAG clearly exceeds ICL.
Figures~\ref{fig:confusion_parametric} to~\ref{fig:confusion_icl} explain
why. In each matrix, the rows give the true label and the columns give the
decision of the model. The top row therefore shows how often a perturbed
clause is caught, and the bottom row shows how often an accurate \texttt{clause} is correctly left alone.

The limiting factor is false alarms rather than missed perturbations.
Averaged over the nine models, the share of perturbed clauses that are
flagged rises from $78.9\%$ under parametric knowledge to $93.9\%$ under RAG
and $91.4\%$ under ICL. In contrast, the share of accurate clauses that are
correctly left alone rises from $30.8\%$ to $69.3\%$ and $64.2\%$. The second
gain is about two and a half times the first, therefore evidence mainly
teaches the model when not to flag.
Without evidence the models flag almost everything. \texttt{Gemini} flags $74.7\%$ of
accurate clauses under parametric knowledge, yet it still catches $92.2\%$ of
perturbed ones, so a high catch rate alone is not evidence of detection
ability. Moreover, a model that always flags would score $51.6\%$ on this
task, while the mean parametric accuracy is only $55.6\%$ and two models fall
below that trivial baseline. Closed-book perturbation detection is therefore
close to a constant response.
RAG leads ICL on both rows, at $93.9\%$ against $91.4\%$ on perturbed clauses
and $69.3\%$ against $64.2\%$ on accurate ones. A retrieved clause set gives
the model a short comparison target. In contrast, the full document offers
much more text against which a correct clause can look inconsistent.
Additional context therefore adds false alarms, which is why this task
reverses the usual ordering between the two settings.

\begin{table*}[!ht]
\centering
\small
\setlength{\tabcolsep}{3pt}
\renewcommand{\arraystretch}{1.15}
\begin{tabular}{@{}lc>{\columncolor{cutshade}}ccc>{\columncolor{cutshade}}ccc>{\columncolor{cutshade}}cc@{}}
\toprule
\rowcolor{headshade}
 & \multicolumn{3}{c}{\textbf{\texttt{Gemini Embedding 2}}} & \multicolumn{3}{c}{\textbf{\texttt{GLM-Embedding-3}}} & \multicolumn{3}{c}{\textbf{\texttt{Qwen3-Embedding-8B}}} \\
\rowcolor{headshade}
 & \textbf{Top-5} & \textbf{Top-10} & \textbf{Top-15} & \textbf{Top-5} & \textbf{Top-10} & \textbf{Top-15} & \textbf{Top-5} & \textbf{Top-10} & \textbf{Top-15} \\
\midrule
Temporal MCQ & 79.79 & \textbf{80.85} & 80.66 & 77.28 & 78.15 & \textbf{78.15} & 75.63 & 76.48 & \textbf{76.49} \\
Perturbation Det. & 81.20 & \textbf{82.06} & 81.84 & 78.88 & \textbf{79.70} & 79.69 & 77.13 & \textbf{77.93} & 77.88 \\
Scenario SA & 83.28 & \textbf{84.37} & 84.33 & 80.87 & 81.71 & \textbf{81.84} & 78.88 & 79.85 & \textbf{80.00} \\
Temporal Scenario SA & 54.27 & \textbf{57.12} & 57.11 & 49.34 & \textbf{52.46} & 52.40 & 46.52 & \textbf{49.63} & 49.62 \\
Relative-Time QA & 25.01 & \textbf{28.85} & 28.59 & 20.77 & \textbf{24.20} & 24.10 & 18.14 & 20.92 & \textbf{21.04} \\
Open-Ended & 36.78 & \textbf{39.72} & 39.57 & 31.73 & 34.80 & \textbf{34.81} & 28.65 & 31.20 & \textbf{31.50} \\
Context Misalign. & 17.40 & \textbf{19.45} & 19.30 & 14.79 & 16.20 & \textbf{16.49} & 12.28 & 13.87 & \textbf{14.28} \\
\midrule
Macro average & 53.96 & \textbf{56.06} & 55.91 & 50.52 & 52.46 & \textbf{52.50} & 48.18 & 49.98 & \textbf{50.12} \\
\bottomrule
\end{tabular}
\caption{Strict accuracy of the RAG setting by embedding model and retrieval depth, averaged over the nine answer models. Shaded columns mark the depth used in the main results. Event sorting is under re-scoring and excluded, so the macro row averages the remaining seven tasks. Best depth per embedding model in bold.}
\label{tab:rag_ablation_combined}
\end{table*}

\texttt{GLM} fails in the opposite direction. Its catch rate falls to $75.1\%$ under
RAG and $68.9\%$ under ICL, whereas its rate on accurate clauses is the
highest in the pool at $86.3\%$ and $79.3\%$. Nevertheless its overall
accuracy stays close to the group mean. A single score can therefore hide two
opposite behaviours, and the confusion matrices make that visible.
\begin{table*}[!ht]
\centering
\small
\setlength{\tabcolsep}{3pt}
\renewcommand{\arraystretch}{1.15}
\begin{tabular}{@{}lccccccccc@{}}
\toprule
\rowcolor{headshade}
&
\multicolumn{3}{c}{\textbf{\texttt{GPT-5}}} &
\multicolumn{3}{c}{\textbf{\texttt{Claude}}} &
\multicolumn{3}{c}{\textbf{\texttt{Gemini}}}\\
\rowcolor{headshade}
&
\textbf{Top-5} &
\textbf{Top-10} &
\textbf{Top-15} &
\textbf{Top-5} &
\textbf{Top-10} &
\textbf{Top-15} &
\textbf{Top-5} &
\textbf{Top-10} &
\textbf{Top-15} \\
\midrule
\texttt{Gemini Embedding 2} &
54.57 & 57.06 & \cellcolor{cutshade}\textbf{57.60} &
55.90 & 58.05 & \cellcolor{cutshade}\textbf{58.53} &
49.78 & 52.87 & \cellcolor{cutshade}\textbf{53.21} \\
\texttt{GLM-Embedding-3} &
49.95 & 53.35 & \cellcolor{cutshade}\textbf{53.39} &
51.29 & 54.71 & \cellcolor{cutshade}\textbf{55.07} &
46.04 & 49.31 & \cellcolor{cutshade}\textbf{49.87} \\
\texttt{Qwen3-Embedding-8B} &
47.96 & \cellcolor{cutshade}\textbf{51.47} & 50.79 &
49.45 & \cellcolor{cutshade}\textbf{52.19} & 52.09 &
43.94 & 46.93 & \cellcolor{cutshade}\textbf{47.09} \\
\midrule
\rowcolor{headshade}
&
\multicolumn{3}{c}{\textbf{\texttt{Kimi}}} &
\multicolumn{3}{c}{\textbf{\texttt{DeepSeek}}} &
\multicolumn{3}{c}{\textbf{\texttt{GPT-OSS}}}\\
\rowcolor{headshade}
&
\textbf{Top-5} &
\textbf{Top-10} &
\textbf{Top-15} &
\textbf{Top-5} &
\textbf{Top-10} &
\textbf{Top-15} &
\textbf{Top-5} &
\textbf{Top-10} &
\textbf{Top-15} \\
\midrule
\texttt{Gemini Embedding 2} &
51.59 & 54.46 & \cellcolor{cutshade}\textbf{54.94} &
49.50 & \cellcolor{cutshade}\textbf{52.08} & 51.44 &
50.14 & \cellcolor{cutshade}\textbf{52.69} & 51.39 \\
\texttt{GLM-Embedding-3} &
47.81 & 50.65 & \cellcolor{cutshade}\textbf{51.39} &
45.30 & \cellcolor{cutshade}\textbf{48.76} & 48.05 &
46.22 & \cellcolor{cutshade}\textbf{49.36} & 49.21 \\
\texttt{Qwen3-Embedding-8B} &
45.59 & 48.27 & \cellcolor{cutshade}\textbf{49.39} &
43.12 & \cellcolor{cutshade}\textbf{46.45} & 45.27 &
44.23 & \cellcolor{cutshade}\textbf{47.13} & 46.41 \\
\midrule
\rowcolor{headshade}
&
\multicolumn{3}{c}{\textbf{\texttt{Qwen}}} &
\multicolumn{3}{c}{\textbf{\texttt{Gemma}}} &
\multicolumn{3}{c}{\textbf{\texttt{GLM}}}\\
\rowcolor{headshade}
&
\textbf{Top-5} &
\textbf{Top-10} &
\textbf{Top-15} &
\textbf{Top-5} &
\textbf{Top-10} &
\textbf{Top-15} &
\textbf{Top-5} &
\textbf{Top-10} &
\textbf{Top-15} \\
\midrule
\texttt{Gemini Embedding 2} &
49.30 & \cellcolor{cutshade}\textbf{52.13} & 51.37 &
44.73 & \cellcolor{cutshade}\textbf{47.69} & 45.32 &
42.72 & \cellcolor{cutshade}\textbf{45.43} & 43.99 \\
\texttt{GLM-Embedding-3} &
45.36 & \cellcolor{cutshade}\textbf{48.59} & 48.50 &
40.91 & \cellcolor{cutshade}\textbf{44.17} & 43.10 &
39.31 & \cellcolor{cutshade}\textbf{42.11} & 40.92 \\
\texttt{Qwen3-Embedding-8B} &
42.59 & \cellcolor{cutshade}\textbf{46.14} & 46.07 &
37.99 & \cellcolor{cutshade}\textbf{41.25} & 39.95 &
36.29 & \cellcolor{cutshade}\textbf{39.87} & 37.54 \\
\bottomrule
\end{tabular}
\caption{Macro strict accuracy (\%) of the RAG setting grouped by answer model. Rows correspond to embedding models, while columns show retrieval depth (Top-5, Top-10, and Top-15). The best retrieval depth for each embedding model is highlighted. Event sorting is excluded because it is under re-scoring.}
\label{tab:rag_modelwise}
\end{table*}
\subsection{RAG Ablation} \label{app:rag_ablation}

Table~\ref{tab:rag_ablation_combined} analyzes the effect of retrieval depth for each embedding model by averaging the performance of all nine answer models. We report the main results in Table~\ref{tab:main_results} using \texttt{gemini-embedding-2} with Top-10 retrieval because this setting provides the most consistent behavior across different task types. Although Top-15 achieves a slightly higher macro average for \texttt{gemini-embedding-2}, the improvement is marginal (56.06 vs. 55.91). In contrast, Top-10 either achieves the best performance or remains very close to the best performance on nearly every task. This indicates that Top-10 provides a better balance across heterogeneous reasoning tasks.

Furthermore, the other embedding models show less stable behavior. For both \texttt{GLM-Embedding-3} and \texttt{Qwen3-Embedding-8B}, the best retrieval depth varies across task types, and the macro average also changes more noticeably. Therefore, selecting a single retrieval depth for these embeddings requires a stronger trade-off between tasks. In comparison, \texttt{gemini-embedding-2} maintains a more consistent performance profile, making it a more reliable choice for the experiments reported in the main paper.

Table~\ref{tab:rag_modelwise} provides a complementary view by grouping results according to the answer model. A similar trend can be observed for the open-weight models, including \texttt{GPT-OSS}, \texttt{Qwen}, \texttt{Gemma}, \texttt{GLM}, and \texttt{DeepSeek}. These models generally achieve their best macro average with Top-10 retrieval, while increasing the retrieval depth to Top-15 often provides little improvement or slightly reduces performance. The main exception is \texttt{Kimi}, which continues to improve with Top-15 retrieval.

In contrast, the closed-source models, namely \texttt{GPT-5}, \texttt{Claude}, and \texttt{Gemini}, consistently obtain their best results with Top-15 retrieval. This trend suggests that these models can better utilize the additional retrieved context without suffering from the extra retrieved passages. On the other hand, most open-weight models appear to be more sensitive to the additional context introduced at Top-15. Based on these observations, we use \texttt{gemini-embedding-2} with Top-10 retrieval throughout the main experiments because it provides the most stable and broadly applicable configuration across both task types and answer models.

\end{document}